%% file: mfstride.tex
\input{header.tex}

\usepackage{footnotebackref}
\usepackage{bibunits}
\usepackage[toc,page,header]{appendix}
\usepackage{minitoc}
\usepackage{url}

\ShortHeadings{\ShortTitle}{Lee, Liang, Tang, and Toh}
\title{{\TheTitle}}

\def\blue#1{#1}

\begin{document}
\doparttoc 
\faketableofcontents 

\begin{bibunit}[plainnat]
\part{} 

\editor{Prateek Jain}
\maketitle

\begin{abstract}%
This work proposes a rapid algorithm, \ours, for
nuclear-norm-regularized convex and low-rank matrix optimization
problems.
\ours efficiently decreases the objective value via low-cost steps
leveraging the nonconvex but smooth Burer-Monteiro (BM) decomposition,
while effectively escapes saddle points and spurious local minima
ubiquitous in the BM form to obtain guarantees of fast convergence
rates to the global optima of the original nuclear-norm-regularized
problem through aperiodic inexact proximal gradient steps on it.
The proposed approach adaptively adjusts the rank for the BM
decomposition and can provably identify an optimal rank for the BM
decomposition problem automatically in the course of optimization
through tools of manifold identification.
\ours hence also spends significantly less time on parameter tuning than
existing matrix-factorization methods, which require an exhaustive
search for finding this optimal rank.
Extensive experiments on real-world large-scale problems of recommendation systems,
regularized kernel estimation, and molecular conformation confirm that
\ours can indeed effectively escapes spurious local minima at which
existing BM approaches are stuck, and is a magnitude faster than
state-of-the-art algorithms for low-rank matrix optimization problems
involving a nuclear-norm regularizer.
Based on this research, we have released an open-source package
of the proposed \ours at
\url{https://www.github.com/leepei/BM-Global/.}
\end{abstract}
\begin{keywords}
	escaping spurious local minima, low-rank models, Burer-Monteiro
	decomposition, nuclear-norm regularization, nonconvex optimization
\end{keywords}

\section{Introduction}
\label{sec:intro}


Consider the following regularized convex matrix optimization problem
\begin{equation}
\min_{X \in \R^{m\times n}}\quad F(X) \coloneqq f(X) + \Psi(X),
\tag{CVX}
\label{eq:matrixform}
\end{equation}
where the loss term \blue{$f:\R^{m\times n} \to \R$} is lower-bounded, convex, and differentiable
with Lipschitz-continuous gradient, and
the regularizer \blue{$\Psi:\R^{m\times n} \to \R \cup\{+\infty\}$}
is convex \blue{in the whole space $\R^{m \times n}$} and has the form
\begin{equation}
	\Psi(X) \coloneqq \lambda \norm{X}_*  + \delta_{\cX}(X), \quad
	X\in \R^{m\times n},
	\label{eq:Psi}
\end{equation}
\blue{
where $\lambda \in \R \setminus \{0\}$, $\|\cdot\|_*$ is the nuclear norm,
$\cX$ is a closed and convex subset of $\mathbb{R}^{m\times n}$,
and $\delta_{\cX}$ is the indicator function such that
\begin{equation}
	\label{eq:indicator}
	\delta_{\cX}(X) = \begin{cases}
		0 & \text{if } X \in \cX,\\
		\infty & \text{ otherwise}.
	\end{cases}
\end{equation}
%
Clearly, in this case, $\Psi$ is nonsmooth, proper, and closed. When
$\lambda \geq 0$, we see directly that $\Psi$ is indeed
convex.
In scenarios with $\lambda < 0$, the problems of interest are also equipped
with a suitable $\cX$ that makes $\Psi$ convex.\footnote{\blue{When
	$m = n$ and $\cX$ is a subset of $\SS^n_+$, the set of $n$ by $n$
	symmetric positive semidefinite matrices, $\Psi$ is convex (in the
	whole space) even if $\lambda < 0$, as in that case the nuclear
	norm becomes the trace of $X$ within $\cX$, which is an affine
	function of $X$, and the feasible region $\cX$ is a convex set.
	Otherwise, we will still need $\lambda \geq 0$ to make $\Psi$
convex.}}}
Without loss of generality, we assume that $m \leq n$ throughout,
which can be achieved easily by conducting a matrix transpose if
necessary.
For our case of \cref{eq:Psi}, as long as $\lambda$ is properly
selected, a low-rank optimal solution to \cref{eq:matrixform} exists,
since the nuclear norm is exactly applying the $\ell_1$-norm to the
singular values of the given matrix.
In practice, singular value decomposition (SVD) for a non-symmetric
matrix $X$ is calculated through the eigendecomposition of the
symmetric matrix $X X^{\top}$ (as we assume $m \leq n$), and thus
computation of SVDs and of eigendecompositions are nearly identical.
We will therefore summarize these two situations simply as requiring
eigendecompositions.

We focus on large-scale problems such that $mn$ is extremely large, so
forming a (possibly dense) matrix $X\in \R^{m\times n}$ explicitly is
spatially and computationally expensive, if not infeasible, and thus a
low-rank solution is necessary for practical reasons.
The nuclear-norm regularization hence serves to induce a low-rank
structure in any solution $X^*$ to \cref{eq:matrixform}.
On the other hand, we assume that $\nabla f(X)$ is either structured
or extremely sparse so that $\nabla f(X) v$ for any vector $v$ can be
computed efficiently; \blue{see \cref{sec:application} for some
examples of sparse $\nabla f$.}
This is necessary for the execution of an inexact proximal gradient
(PG) step; see details in \cref{sec:alg}.

To deal with the high problem dimensionality in \cref{eq:matrixform},
a popular approach is the Burer-Monteiro (BM)
decomposition \citep{BurM03a} that explicitly writes
$X$ as a product of two low-rank
matrices with a pre-specified rank $k \leq m$.
Explicitly, we get
\begin{equation}
\min_{W \in \R^{m \times k}, H \in \R^{n\times k}}\quad
F\left( W,H\right) \coloneqq f\left( WH^{\top} \right)
 + \Psi\left( W H^{\top} \right).
\tag{BM}
\label{eq:bm}
\end{equation}
The spatial cost of $O(mn)$ for storing $X$ in \cref{eq:matrixform} is
then reduced to $O( (m+n)k)$ in \cref{eq:bm}.
Numerous efficient algorithms for solving \cref{eq:bm} are therefore
proposed.

Among applications of \cref{eq:matrixform}, one of the most prominent
example, and also our motivating problem, is the following low-rank matrix
completion problem whose target is to recover the whole ground truth
matrix $A\in \R^{m \times n}$ from its observed entries enumerated by
an index set $\Omega$:
\begin{equation}
	\label{eq:mc}
	\min_{X \in \R^{m \times n}}\quad \frac12 \|P_{\Omega}( X
- A )\|_F^2+ \lambda \norm{X}_*,
\tag{MC}
\end{equation}
where $\norm{\cdot}_F$ is the Frobenius norm and
\[
	\left(P_{\Omega}\left( A \right)\right)_{i,j} = \begin{cases}
		A_{i,j} & \text{ if } (i,j) \in \Omega\\
		0 & \text{ otherwise}
	\end{cases}.
\]
It is well-known that the factorized form in \cref{eq:bm} of
\cref{eq:mc} for a given rank $k$ is
\begin{equation}
\label{eq:mf}
\min_{W \in \R^{m \times k}, H \in \R^{n \times k}}\quad \frac12
\|P_{\Omega}( W H^{\top} - A )\|_F^2+ \frac{\lambda}{2}
( \norm{W}_F^2 + \norm{H}_F^2 ),
\tag{MF}
\end{equation}
which is often called the matrix factorization (MF) problem in the
machine learning community.
One can easily show that the global optimal objectives of \cref{eq:mc} and
\cref{eq:mf} coincide whenever $k$ is large enough such that there is
at least one optimal solution $X^*$ of \cref{eq:mc} with $\rank(X^*)
\leq k$.
(See \cref{lemma:equivalent}.)
Apparently, even the objective evaluation for \cref{eq:mc} requires an
eigendecomposition that costs $O(m^3)$, while
for \cref{eq:mf}, objective evaluation and variable updates all
require cost of only $O((m+n)k)$.

Unfortunately, the lower computational and spatial costs of solving
\cref{eq:mf} and thus the more general \cref{eq:bm} comes with a price. A disadvantage of the BM method is that the rank $k$ needs to be
specified in advance, and a good value for $k$ can be hard to estimate
a priori.
Another even more severe issue is that the problem \cref{eq:mf} is a
nonconvex one, meaning that algorithms for solving it could get stuck
at spurious local optima (local but not global optima) or
saddle points (\blue{\textit{i.e.}, points with zero gradients that are not local extrema}).  Such points can give
terrible performance for predicting missing entries.
The simplest example would be that for any $k > 0$, letting
$W$ and $H$ be matrices of all zeros in \cref{eq:mf} will directly
generate a saddle point, but this clearly is not a solution in general.
It is also recently shown by \cite{YalZLS22a,OcaSV22a} that there are
indeed problems with a sufficiently large $k$ that still possess
spurious local minima with a terribly large objective value, and thus how to
escape from such saddle points and spurious local minima is \blue{a} critical
issue for the BM approach to produce satisfactory performance.

On the other hand, the convex problem \cref{eq:mc} or \cref{eq:matrixform} can
be solved directly through PG-type algorithms that are
able to find the global optima \citep{TohY10a}, but the cost of the
eigendecomposition in the proximal operation is extremely expensive,
even if only a subset of singular values/vectors is required.
State-of-the-art methods for \cref{eq:mc} like those by
\cite{HsiO14a,YaoKWL18a} thus resort to approximate
eigendecompositions computed through the power method to reduce the
time cost of eigendecompositions. 
The power method also has the benefit that only $XV$ for some
thin matrix $V$ is needed at each iteration, so explicit computation of
$X$ is not needed.
However, we observe that in practice, usually the convex approaches
based on PG tend to be rather slow because their
computational cost per iteration is still not comparable to those for
\cref{eq:mf} or \cref{eq:bm}.
%

In this work, we propose a highly-efficient algorithm, \ours, that
combines the advantages of both approaches with theoretical guarantees.
Our method fully utilizes the computational and spatial advantages
of \cref{eq:bm} to have a running time similar to the state of the art
for \cref{eq:bm}.
Meanwhile, \ours also possesses guarantees for convergence to the
global optima just like those approaches for solving
\cref{eq:matrixform}.
With suitable inexactness conditions for the PG steps,
we also obtain a sublinear convergence rate for the general convex
case, and further get faster rates when the objective function
satisfies the Kurdyka-{\L}ojasiewicz (KL) condition
\citep{Kur98a,Loj63a}.
Our algorithm attains these appealing guarantees through sporadically
resorting to convex lifting steps that conduct one iteration of
inexact PG on \cref{eq:matrixform} to escape from saddle points
and spurious local optima at which existing methods for \cref{eq:bm}
got stuck at.
We emphasize that for alternating between inexact PG and other
update steps, our convergence and rate guarantees are novel, as existing
analyses for convergence and rates of inexact PG rely on geometrical
properties of the PG iterates that will be destroyed when other
updates are inserted.

Through the techniques of manifold identification
\citep{HarL04a,LewZ13a} in our analysis, another major and novel
contribution of this work is that \emph{the optimal rank for
	\cref{eq:bm} will provably be found by the proposed method
automatically}, so no additional parameter tuning for the
optimization side is required for attaining satisfactory results for
practical applications of \cref{eq:matrixform}.
Numerical results also show that our method is significantly faster than
state-of-the-art solvers for \cref{eq:matrixform}, and can effectively
escape from saddle points and spurious local minima of \cref{eq:bm}.

\subsection{Related Works}
%
\noindent\textbf{Methods for \cref{eq:matrixform}.}
The convex problem \cref{eq:matrixform} falls in the category of regularized
optimization, and many efficient algorithms in the literature are
available.
However, most methods for regularized optimization concentrate on the
scenario that the proximal operation can be conducted efficiently but
obtaining information of the smooth term is the major computational
bottleneck, which is apparently not the case for \cref{eq:matrixform}.
Practical methods specifically designed for \cref{eq:matrixform} all
take into serious account the expensive SVD involved in the nuclear
norm \citep{TohY10a,HsiO14a,YaoKWL18a}, and they all focus on the most
popular setting that $f$ is a quadratic term.
In this case, high-order methods like proximal (quasi-)Newton is
useless because the subproblem has the same form as the original
problem itself.
Therefore, these works all consider first-order methods such as
the (inexact) PG or accelerated PG (APG) methods
\citep{Nes13a,BecT09a,BecT09b}.
\citet{TohY10a} used the Lanczos method to conduct approximate SVD, and
\citet{HsiO14a} proposed to apply the power method for approximate
SVD and to use the rank $k$ decided by their inexact PG method to conduct
another convex optimization step with respect to a subproblem of
dimension $k \times k$ after each PG step.
The power method in \cite{HsiO14a} effectively uses the current
iterate as warmstart and is much more efficient than the Lanczos
approach in \cite{TohY10a}, but the additional convex
optimization step turns out to be time-consuming.
To improve the efficiency of \cite{TohY10a} and \cite{HsiO14a},
\citet{YaoKWL18a} combined the two approaches to turn to inexact APG
using the power method.

More recently, motivated by PG's ability of manifold identification,
\cite{BarIM20a} proposed to alternate between an exact PG step and a
Riemannian (truncated) Newton step on the currently identified
manifold for general regularized optimization, and applied
this method to a toy problem of \cref{eq:matrixform} in their experiments.
Different from ours, their usage of manifold identification is for
showing that their method could obtain superlinear convergence,
but their algorithm is not feasible for large-scale problems
considered in this work because they considered exact PG only and
required the explicit computation of $X_t$.

\noindent\textbf{Convergence of Inexact Proximal Gradient}
The global convergence of our method is achieved through the safeguard
of inexact PG steps, but we do not require the inexact PG step to
always decrease the objective value.
This feature combined with the BM phase makes the analysis difficult.
Existing analyses for inexact PG
\cite{Com04a,SchRB11a,JiaST12a,HamWWM22a} utilize telescope sums of
inequalities in the form of $\norm{X_{t+1} - X^*}^2 \leq \text{error
term} + \norm{X_t - X^*}^2 - \alpha\left( F(X_t) - F(X^*)\right)$ for
any $X^* \in \Omega^*$ and some $\alpha > 0$ to prove convergence and
rates.
Therefore, those approaches cannot allow for alternating between
inexact PG and other steps because that will nullify the technique of
telescope sums.
On the other hand, analyses compatible with other steps like those in
\cite{SchT16a,BonLPP16a,LeeW18a} require strict decreasing of the
objective in the inexact PG step (either explicitly or implicitly),
which imposes an additional burden.

The work of \cite{YanLCT21a} that applies inexact PG to the
semidefinite programming (SDP) relaxation of polynomial optimization
problems is probably the closest to our approach in that their method
alternates between nonmonotone inexact PG and an alternative step.
However, their alternative step is only accepted when it decreases the
objective by an absolute amount $\epsilon > 0$, so eventually the
alternative steps are always rejected when the objective converges,
but we do not have such restrictions.
Our analysis also provides more comprehensive convergence guarantees
under different conditions as well as identification of the optimal
rank using techniques totally different from that of \cite{YanLCT21a}.

\noindent\textbf{Methods for escaping saddle points.}
There has recently been a thriving interest in studying smooth
optimization methods that can escape strict saddle points with at
least one negative eigenvalue in the Hessian
\citep{LeePPSJR19a,JinGNKJ17a,RoyW18a,CarDHS18a,RoyOW20a,AgaABHM17a}.
However, these methods are unable to deal with degenerate saddle points
where the smallest eigenvalue of the Hessian is exactly zero,
and neither could they deal with spurious local minima that might be
arbitrarily far away from the global ones.
On the other hand, our method can handle all such difficult cases
appearing in \cref{eq:bm} by resorting to convex lifting, and
\cref{lemma:equivalent} together with \cref{thm:conv} show that
our method indeed will converge to the global optima.
Moreover, existing methods for escaping strict saddles are mainly of
theoretical interest, and their empirical performance is usually not very
impressive, while our method is designed for practical large-scale usage
and greatly outperforms state-of-the-art methods for
\cref{eq:matrixform} and \cref{eq:bm} by a large margin on real-world
data, as we shall see in the numerical experiments later.

\noindent\textbf{Absence of spurious stationary points for \cref{eq:mf}.}
To cope with possible spurious local minima and degenerate saddle
points of matrix factorization, there is also a growing interest in
analyzing its optimization landscape.
Most such works consider the quadratic loss:
\begin{equation}
	f(X) = \norm{P_{\Omega}\left( A - X \right)}_F^2.
	\label{eq:quad}
\end{equation}
\blue{\cite{jain2013low,GeLM16a,SunL16a,GeJZ17a}} confined their
analyses to minimizing \cref{eq:quad} \blue{with deliberately selected
regularizers, in the factorized form with variables $W$ and $H$ using
the alternating minimization method that alternatively minimizes the
objective function with respect to $H$ and to $W$.}
They proved the absence of spurious local minima
for only the ideal case in which each $(i,j)$ belongs to $\Omega$ with
a fixed probability $p \in (0,1]$ and the observations are noiseless.
\blue{They showed that this approach enjoys appealing theoretical
guarantees and fast computation under suitable conditions such as
variants of the restricted isometry property and its variants.}
But these assumptions generally fail in practice, and the selected
regularizers are not widely adopted.
In particular, for applications like recommendation systems, elements
of $\Omega$ are already biased selections by an existing system and
will never obey the independent random assumption. In addition, real-world
data always contain noisy observations and measurement corruption.
\cite{CheCFM19a,YeD21a} focused on
\cref{eq:quad} with $\Omega = \{1,\dotsc,m\} \times \{1,\dotsc,n\}$ only,
and their techniques could be hard to generalize to other problems.
Different from these works, we do not have any assumption on the
underlying data, as it has been shown recently in
\cite{YalZLS22a,OcaSV22a} that some problem classes of matrix
factorization and SDP indeed contain numerous spurious local minima.
\blue{Moreover, we do not confine our algorithm or analysis to the special
case of \cref{eq:quad} but aim at general $f$ with minimal assumptions.
However,} our method can still find the global optima with ultra-high
practical efficiency in the presence of non-strict saddle points and
spurious local optima.

\subsection{Organization}
This paper is organized as follows.
Preliminaries for our algorithmic development are provided in
\cref{sec:prelim}.
In \cref{sec:alg}, we describe our main algorithmic framework.
\cref{sec:analysis} provides a comprehensive theoretical analysis of our algorithm,
including its global convergence to the global optima, convergence
rates, and identification of the right rank for \cref{eq:bm} within a
finite number of iterations.
We then give more details in \cref{sec:application} several
realizations of \cref{eq:matrixform} in applications.
Numerical experiments on real-world problems in such applications are
conducted in \cref{sec:exp}, and finally \cref{sec:conclusions}
concludes this work.
Detailed implementations for the applications used in our numerical
experiments and additional experimental results are provided in
the appendices starting from Page~\pageref{part:2}.

\section{Preliminaries}
\label{sec:prelim}
This section first lays out our notations in this work, and then
provides background knowledge that will facilitate further
descriptions in our development of algorithm and theory.
We use $\tr(\cdot)$ to denote the trace of a square matrix.
For any $x \in \R^m$, $\diag(x)$ is the $m \times m$ diagonal matrix
whose diagonal entries are those in $x$.
We use $\inprod{\cdot}{\cdot}$ to denote the standard inner product and
$\norm{\cdot}$ 
to denote its induced norm.
In particular, for vectors $x,y\in\R^n$, this is the standard inner
product such that
$\inprod{x}{y} \coloneqq \sum_{i=1}^n x_i y_i$,
with the norm being the Euclidean norm,
and for matrices $A,B \in \R^{m \times n}$, this inner product is
defined as 
$\inprod{A}{B} \coloneqq \tr\left( A^{\top} B \right)$,
where $A^{\top}$ is the transpose of $A$ and the corresponding norm is
the Frobenius norm.
We denote by $\R^m_+$ the nonnegative orthant in the $m$-dimensional
Euclidean space, by $\SS^n$ the set of $n$ by $n$ symmetric real
matrices, and by $\SS^n_+$ the cone of symmetric positive semidefinite matrices in $\SS^n$.
$I_n$ denotes the identity matrix with dimension $n \times n$, and
$e_n \in \R^n$ is the vector of ones. 
The subscript $n$ is often omitted when the dimensionality is clear.
For $x
\in \R^m$, we let $[x]_+$ be its Euclidean projection onto $\R^m_+$,
and for $X \in \SS^n$, $[X]_+$ is its Euclidean projection onto $\SS^n_+$.
In particular, if $X$ admits an eigendecomposition $X = U \diag(\Sigma) U^{\top}$,
where $U \in \R^{n \times n}$ is orthonormal such that $U U^{\top} =
U^{\top} U = I_n$ and $\Sigma \in \R^n$, we have that $[X]_+ = U \diag\left( \left[ \Sigma \right]_+ \right) U^{\top}$.
We note that SVDs and eigendecompositions are unique up to
permutations of the eigenvalues or the singular values, so we will
simply say ``the'' SVD or ``the'' eigendecomposition to refer to the
one such that the singular values or eigenvalues are sorted in
descending order. (And this can be an arbitrary one when there are
repeated singular values.)
Given any set $\mathcal{X}$, we use $\delta_{\mathcal{X}}$ to denote its indicator function
described in \cref{eq:indicator}.
For any convex function $h$, we use $\partial h$ to denote its
subdifferential.

Throughout this work, we will heavily use
the proximal operation.
Given a function $\Psi$, this operation is defined as
\begin{equation}
	\prox_{\Psi}(X) \coloneqq \argmin_{\hat X}\;
	\frac{1}{2} \|X - \hat X\|^2 + \Psi( \hat X
	).
	\label{eq:proximal}
\end{equation}
For $\Psi$ convex, proper, and closed, it is well-known that
\cref{eq:proximal} is well-defined and single-valued everywhere.
When $\Psi = \lambda \norm{\cdot}_*$ for some $\lambda > 0$,
\cref{eq:proximal} has a closed-form solution \citep{LewO96a}.
Given a matrix $X\in\R^{m\times n}$ with rank $k$ and with its SVD
written as $X = U \diag(\Sigma) V^{\top}$ for some $U \in \R^{m \times
k}, V \in \R^{n \times k}$ that are orthogonal and $\Sigma \in
\R^k_+$, we have
\begin{equation}
	\prox_{\beta \norm{\cdot}_*}\left( X \right) = U
	\diag\left( \left[\Sigma -  \beta e\right]_+\right) V^{\top}
	\label{eq:proxnuc}
\end{equation}
for any $\beta > 0$.
Similarly, if $X \in \SS^n$ admits an eigendecomposition $X = U \diag(\Sigma)
U^{\top}$, we have from \cite{LewO96a}
that for any $\beta > 0$, 
\begin{equation}
	\prox_{\beta (\lambda \norm{\cdot}_* + \delta_{\SS^n_+})}\left( X \right) = [X]_+ = U
	\diag\left( \left[\Sigma - \beta \lambda e\right]_+\right) U^{\top}.
	\label{eq:proxsym}
\end{equation}
In this paper, we focus on the two scenarios of
$\Psi = \lambda \norm{\cdot}_*$ and $\Psi = \lambda \norm{\cdot}_* +
\delta_{\SS^n_+ \cap \mathcal{S}}\left( \cdot \right)$ (for $m=n$) for some
closed and convex set $\mathcal{S}$. 
We will see in \cref{lemma:equivalent} that
the former results in the following form of \cref{eq:bm}:
\begin{equation}
\min_{U \in \R^{m \times k}, V \in \R^{n\times k}}\quad
\tilde F\left( W,H \right) \coloneqq f\left( WH^{\top} \right)
 + \frac{\lambda}{2} \left( \norm{W}_F^2 + \norm{H}_F^2
\right),
\tag{BM-nuclear}
\label{eq:bm1}
\end{equation}
while the latter leads to
\begin{equation}
\min_{W \in \R^{n \times k}}\quad
\tilde F\left( W \right) \coloneqq f\left( WW^{\top} \right)
 + \lambda \norm{W}_F^2 + \delta_{\mathcal{S}}\left( W W^{\top} \right).
\tag{BM-PSD}
\label{eq:bm2}
\end{equation}
For the latter case, when we deal with
\cref{eq:matrixform}, for easier calculation, we will sometimes
consider the smooth term as $\tilde f(X) \coloneqq f(X) + \lambda
\inprod{I}{X}$ and the regularizer as $\tilde \Psi(X) \coloneqq
\delta_{\SS^n_+ \cap \mathcal{S}}(X)$ instead, which is equivalent to the
original problem because the nuclear norm on a positive semidefinite
matrix is exactly the trace of the same matrix.

The equivalence between the nuclear norm in \cref{eq:matrixform} and
the Frobenius norm squared in \cref{eq:bm1,eq:bm2} is formally stated in the
following lemma.
\begin{lemma}
\label{lemma:equivalent}
Given any $X \in \R^{m \times n}$, we have
\begin{equation}
\norm{X}_* = \min_{W,H: WH^{\top} = X}\quad \frac12 \left( \norm{W}_F^2
+ \norm{H}_F^2
\right).
\label{eq:nuclear}
\end{equation}
Moreover, if the SVD of $X$ is
\[
	X = USV^\top = \sum_{i=1}^{k} \sigma_i u_i v_i^{\top},
\]
where $S = \diag(\sigma_1,\dots, \sigma_k)$ and $\sigma_i > 0$ are the singular values, $k = \rank(X)$,
$U = [u_1,\dotsc,u_{k}] \in \R^{m \times k}$ and $V = [v_1,\dotsc,v_{k}] \in
\R^{n \times k}$ are both orthonormal,
the minima of the right-hand side of \cref{eq:nuclear} are exactly
those
\begin{equation}
	\hat W \coloneqq \left[ \sqrt{\sigma_{\tau(1)}} u_{\tau (1)},\dotsc,
	\sqrt{\sigma_{\tau(k)}} u_{\tau(k)} \right],
	\hat H \coloneqq \left[ \sqrt{\sigma_{\tau (1)}} v_{\tau(1)},\dotsc,
	\sqrt{\sigma_{\tau (k)}} v_{\tau (k)} \right],
	\label{eq:bmsol}
\end{equation}
where $\tau$ is \blue{any given} permutation of $\{1,\dots,k\}$.
Therefore, for any global optimum $(W^*,H^*)$ of \cref{eq:bm1},
$X^* \coloneqq W^* \left( H^* \right)^{\top}$ is also a global optimum
to \cref{eq:matrixform} with $\Psi = \lambda \norm{\cdot}_*$,
provided that there is a global optimum $\hat X$ of
\cref{eq:matrixform} with $\rank(\hat X) \leq k$, and for any optimal
solution $X^*$ of \cref{eq:matrixform} with SVD $X^* = U^* \Sigma^*
(V^*)^{\top}$,
\[
	W^* = U^* (\Sigma^*)^{\frac12}, \quad 
	H^* = V^* (\Sigma^*)^{\frac12}
\]
form a global solution of \cref{eq:bm1} for any $k \geq
\rank(X^*)$.
Likewise,
for any global optimum $W^*$ of \cref{eq:bm2},
$X^* \coloneqq W^* \left( W^* \right)^{\top}$ is also a global optimum
to \cref{eq:matrixform} with $\Psi = \lambda \norm{\cdot}_* +
\delta_{\SS^n_+}(\cdot) + \delta_{\mathcal{S}}(\cdot)$,
provided that there is a global optimum $\hat X$ of
\cref{eq:matrixform} with $\rank(\hat X) \leq k$.
\end{lemma}
\cref{eq:bmsol} is directly from \cite{RenS05a}, and
the rest of \cref{lemma:equivalent} are also well-known.

\section{Algorithmic framework}
\label{sec:alg}

In this section, we present a detailed description of the  proposed
\ours that can be split into two phases: the BM phase and the
convex lifting phase.
The high-level idea of the proposed framework is to fully utilize the
efficiency in solving \cref{eq:bm} from the smoothness of the
objective function whenever possible.
When only limited progress can be further made in the BM phase with
the current iterate $(\tilde W_t, \tilde H_t)$, we turn to the convex
lifting phase, which conducts one step of inexact proximal gradient
(PG) on \cref{eq:matrixform} from the iterate $\tilde X_t = \tilde W_t
\tilde H_t^{\top}$.
In the inexact PG step, an approximate
eigendecomposition algorithm is employed to obtain the next iterate
$X_{t+1} = W_{t+1} H_{t+1}^{\top}$.
The rank of the approximate eigendecomposition is dynamically
increased (and the proximal step decreases the rank of its output) to
guarantee that the correct rank at the optimum can be found within
finite iterations of our algorithm.

\cref{eq:bmsol} provides a convenient way to transform an
iterate of \cref{eq:matrixform} to that of \cref{eq:bm1} or
\cref{eq:bm2}.
(Transforming the other way round is straightforward.)
Although this transformation requires an eigendecomposition, we will
see shortly that our convex lifting step will
generate the exact eigendecomposition of its output (which is obtained
from conducting an exact proximal operation on an approximate
eigendecomposition of the input matrix),
so the transformation can be done almost for free.
We emphasize that the matrix $X = WH^\top$ is never explicitly formed
when executing \ours.

The only requirement we put on the BM phase is a mild and
implementable nonincreasing objective condition:
\begin{equation}
	\tilde F(\tilde W_t, \tilde H_t) \leq \tilde F( W_t,  H_t)
	\quad 
	\Leftrightarrow\quad 
    F(\tilde W_t\tilde H_t^\top) \leq F(W_tH_t^\top).
	\label{eq:nonincrease}
\end{equation}
\blue{With this minimum requirement, many suitable solvers for
the BM-phase subproblem can be applied at the user's will, and so is
the stopping condition for the selected solver.}
We can even skip the BM phase from time to time without violating
\cref{eq:nonincrease} when this skipping is deemed useful.

In the convex lifting phase, we emphasize that since we never
explicitly store the dense matrix variable $X$ due to the high spatial
cost, exact decomposition for computing the eigendecomposition
becomes impractical. This is one of the major reasons to consider
approximate eigendecompositions that are computed through an iterative
process where each iteration of which only requires the computation of
matrix-matrix products of the form $XV$ for some thin matrix $V$.
This product $XV$ can be computed efficiently without explicitly
forming $X$ if $X$ can be decomposed into the sum of a low-rank matrix
and a highly sparse matrix. This is another reason to consider
low-rank problems promoted by the nuclear norm, as the proximal
operation of the nuclear norm often leads to a low-rank iterate.
See more details in \cref{eq:proxnuc,eq:proxsym}, and
\cref{thm:identify} in \cref{sec:analysis}.

Although APG is state of the art for \cref{eq:matrixform}, such as the
works of \cite{TohY10a,YaoKWL18a} that utilized the APG method
to obtain theoretical and practical convergence
faster than that of the PG method, \blue{APG is not applicable in our
framework because when we insert other update steps between two APG iterations, existing proofs for convergence guarantees of APG are invalidated.
We also note that for the error-bound condition considered in
\cref{eq:sharpness}, PG could achieve faster convergence rates (see \cref{linearT}) than APG.}
Moreover, notwithstanding using a vanilla PG method in our algorithm results
in a slower worst-case convergence rate than the APG method, our major
workhorse in reducing the objective value efficiently is actually the
BM phase, while the PG step mainly serves as a safeguard for global
convergence and the mechanism for identifying the correct rank.
Therefore, we do not expect the PG step to provide much
objective decrease empirically. In the numerical experiments, we will
also see that the added BM phase indeed effectively decreases the
objective value with a short running time, making the proposed algorithm
outperform the APG method of \cite{TohY10a,YaoKWL18a} significantly.

\subsection{Inexact Proximal Gradient Step}
Given the iterate $\tilde X_t$ and a step size $\alpha_t > 0$ at the $t$th
iteration, the exact PG step $\hat X_t^+(\alpha_t)$ is
computed by
\begin{equation}
	\begin{aligned}
		\hat X_t^+( \alpha_t )
	=&~ \prox_{\alpha_t \Psi} ( \tilde X_t - \alpha_t \nabla f(
	\tilde X_t )) \\
	=&~
	\argmin_{Y}\, \big\{ Q^{\alpha_t}_t( Y ) \coloneqq
	\inprod{\nabla f(\tilde X_t)}{Y-\tilde X_t} + \frac{\|Y - \tilde X_t\|^2_F}{2\alpha_t}
	 + \Psi( Y )  -
	\Psi( \tilde X_t )\big\}.
\end{aligned}
	\label{eq:proxgrad}
\end{equation}
After finding a suitable $\alpha_t$ ensuring a sufficient decrease
in $F$, the exact PG method then assigns $X_{t+1} = \hat
X_t^+(\alpha_t)$.
For our inexact scheme, we focus on the scheme that the computation of
$\nabla f(\tilde X_t)$ and the proximal operation in \cref{eq:proxnuc}
or \cref{eq:proxsym} are exact, and the inexactness in the PG step
comes from the approximate eigendecomposition.
In particular, the approximate eigendecomposition is the exact
eigendecomposition of a matrix approximating the original one, and
thus we can easily conduct exact eigenvalue/singular value truncation in
\cref{eq:proxnuc} or \cref{eq:proxsym} of this approximation matrix.

We can therefore view the calculation of our inexact PG step with such
an inexactness quantified by some $\epsilon_t \geq 0$ as
\begin{equation}
	X_{t+1} = X^+_{t}(\alpha_t)= \prox_{\alpha_t \Psi} \left( \tilde
	Z_t \right) = \argmin_{\hat Y} \inprod{\nabla f(\tilde X_t) +
	\Epsilon_t}{\hat Y - \tilde X_t} + \frac{1}{2\alpha_t} \norm{\hat
	Y - \tilde X_t}^2_F + \Psi(\hat Y),
	\label{eq:inexactpg}
\end{equation}
for some $\Epsilon_t$ and $\tilde Z_t$ that satisfy
\begin{equation}
	\norm{\Epsilon_t}_F \leq \epsilon_t,\quad
	\tilde Z_t = \tilde X_t - \alpha_t \nabla f(\tilde X_t)- \alpha_t\Epsilon_t.
	\label{eq:errordef}
\end{equation}
Such an inexactness can also be described by the following abstract
representation.
\begin{equation}
	\blue{\min_{G \in \partial Q^{\alpha_t}_t\left( X_{t}^+(\alpha_t) \right)}
	\norm{G}_F \leq \epsilon_t.}
	\label{eq:inexact}
\end{equation}

For the stepsize choices, the only requirement of our framework is
that the stepsize $\alpha_t$ is uniformly bounded and satisfies either
of the following criteria for some given $\delta \in (0,1)$.
\begin{align}
	\label{eq:back}
	f(X_{t+1} ) &\leq f( \tilde X_t ) + \inprod{\nabla
		f( \tilde X_t
	)}{X_{t+1} - \tilde X_t} + \frac{\delta}{ \alpha_t} \|{X_{t+1} -
\tilde X_t}\|_F^2,\quad\text{ or }\\
	F(X_{t+1} ) &\leq F( \tilde X_t ) +
	\delta Q_t^{\alpha_t}( X_{t+1} ).
	\label{eq:decrease}
\end{align}

We note that \cref{eq:back} does not necessarily imply monotonically
decreasing objective values, and it is actually in general independent
of how accurately \cref{eq:inexact} is solved.
On the other hand, although the value of $Q_t^{\alpha_t}(X_{t+1})$ is
affected by $\epsilon_t$, it is not necessarily negative
(although the minimum of $Q_t^{\alpha_t}(\cdot)$ is), and thus the
objective value might still be nonmonotone.

To ensure that $\alpha_t$ is bounded, we need to specify an upper bound
$\alpha_{\max}$ and a lower bound $\alpha_{\min}$.
It is known that if $\nabla f$ is $L$-Lipschitz continuous, then
\cref{eq:back} holds for any $\alpha_t < 2\delta / L$, and thus we
assign $\alpha_{\min} <2\delta/L$ to ensure that our algorithm is
well-defined. As for \cref{eq:decrease}, note that when
$\alpha_t\leq 1/L$, $Q_t^{\alpha_t}(Y)$ is a majorization of
$F(Y)-F(\tilde{X_t})$, so \cref{eq:decrease} holds. Thus, we can set
$\alpha_{\min} <1/L.$

Since the inexact PG step is not the major tool for
reducing the objective value,
we simply assign a fixed step size $\alpha_t
\equiv \alpha$ in our implementation.
We have also experimented with the approach of SpaRSA \citep{WriNF09a}
that combines a spectral initialization of \citet{barzilai1988two}
with backtracking line search, but its empirical performance is worse
than the fixed-step variant (see the supplementary materials), likely due to the additional
eigendecompositions in backtracking.

We summarize the version of our algorithm for \cref{eq:bm1} in
\cref{alg:mfstride}, and the version for \cref{eq:bm2} can be obtained
by considering only one matrix $W\in \mathbb{R}^{m\times k}$ and
replacing SVDs with eigendecompositions.
\blue{In either case, the approximate SVD ($U_t$,$\hat \Sigma_t$,
$V_t$) of $\tilde Z_t$ is first computed, and we can use it to compute
$\epsilon_t^2 = \|\tilde Z_t - (\tilde X_t - \alpha_t \nabla f(\tilde
X_t)\|_F^2$
easily by utilizing the fact that both $\tilde Z_t$ and $\tilde X_t$
are presented in a low-rank factorized form and $\nabla f(\tilde X_t)$
is either structured or sparse. We can therefore monitor the progress
of the approximate SVD algorithm for reaching a given $\epsilon_t$,
and even use it to adjust the rank.}

\begin{algorithm}[tbh]
\DontPrintSemicolon
\SetKwInOut{Input}{input}\SetKwInOut{Output}{output}
\SetKwComment{Comment}{//}{}
\caption{\ours}
\label{alg:mfstride}
\Input{$\lambda,\delta,\alpha_{\max} \geq \alpha_{\min} > 0$ with
$\alpha_{\min} < \delta / (2L)$ or $\alpha_{\min} < 1/L$,
 initial rank $k$, a nonnegative sequence
	$\{\epsilon_t\}$ with $\epsilon_t \rightarrow 0$}

	{\bf initialization:} $W_0 \in \R^{m \times k}, H_0
	\in \R^{n\times k}$ such that $W_0 H_0^\top \in \dom(\Psi)$.

\For{$t=0,\dotsc$}{
	{\bf (BM phase)} Compute $\tilde W_t \in \R^{m \times k}$, $ \tilde H_t \in
	\R^{n\times k}$ as an approximate solution to \cref{eq:bm}
	(starting from $(W_t, H_t)$) satisfying $\tilde W_t \tilde
	H_t^\top\in \dom(\Psi)$ and \cref{eq:nonincrease}

	{\bf (Convex lifting phase)} Decide $\alpha_t \in [\alpha_{ \min},
	\alpha_{\max}]$ and obtain the SVD $U_t, \Sigma_t, V_t$
	of $X_{t+1}$ (without forming $X_{t+1}$ explicitly) through
	\cref{eq:proxnuc}
	such that \cref{eq:inexact} holds and either
	\cref{eq:back} or \cref{eq:decrease} is satisfied

	$W_{t+1} \leftarrow U_{t} \diag(\sqrt{\Sigma_{t}})$,
	$H_{t+1} \leftarrow V_t \diag(\sqrt{\Sigma_{t}})$
	\label{l:wh}
    
    If a partial/inexact eigendecomposition is used, the number of
	eigenvalues to compute is updated to $k + k_{\rm add}$ where $k_{\rm add} \geq 0$

	{\bf (Rank update)} $k = \rank(\diag(\Sigma_t))$.
}
\end{algorithm}

\blue{
\begin{remark}
    \label{remark-bmsolver}
    From the description of Algorithm \ref{alg:mfstride}, one needs to
	choose a subproblem solver and a stopping condition for the factorized nonconvex subproblem.
	We emphasize that the choice of the solver can be arbitrary and
	highly depends on the application.
	For example, in our experiments, we applied the polyMF-SS method of
	\citep{WanLK17a} in the matrix completion problem
	and Manopt \citep{manopt} for manifold optimization in the
	nonlinear semidefinite programming problem.
	It is also possible to extend the idea of preconditioned
	gradient-type methods \citep{tong2021accelerating,xu2023power} for
	unregularized ill-conditioned problems to our scenario.
\end{remark}
}

\section{Analysis}
\label{sec:analysis}
This section provides theoretical guarantees for \cref{alg:mfstride}.
In partiular, we first give suitable conditions for $\epsilon_t$ to
guarantee global convergence, and then further obtain convergence
rates by imposing further requirements on $\epsilon_t$.
Next, rank identification of \cref{alg:mfstride} is proven under a
nondegeneracy condition, which shows that for any subsequence
$\{X_{t_i}\}_{i=0}^\infty$ of the iterates that converge to a solution
$X^*$, $\rank(X_{t_i}) = \rank(X^*)$ for all $i$ large enough, so
the rank $k_t$ in \cref{eq:bm1} will eventually be automatically
adjusted to the optimal value.

\subsection{Global Convergence and Worst-case Rates}
Our first main theoretical result is the global convergence of our
algorithm.
In our proofs, we let $\Omega^*$ denote the solution set to
\cref{eq:matrixform}, and $F^*$ the optimal objective.
For notational simplicity, we denote
\[\dist(X, \Omega^*) \coloneqq \inf_{X^* \in \Omega^*} \norm{X
-X^*}_F.\]

\begin{theorem}
	Consider \cref{eq:matrixform} with $\Psi$ defined in
	\cref{eq:Psi}.
	Then $\Omega^*$ is compact and nonempty, and $F$ is coercive.
%
	If $\nabla f$ is Lipschitz continuous and
	\begin{equation}
		\sum\epsilon_t^2<\infty
		\label{eq:sqsummable}
	\end{equation}
	in \cref{alg:mfstride} with the condition \cref{eq:back} being
	enforced,
	then for any initialization $W_0, H_0$, we always
	have $\dist(X_t, \Omega^*) \rightarrow 0$, there is at least one
	limit point of $\{X_t\}$, any such limit point is a global
	solution to \cref{eq:matrixform}, and $F(X_t) \rightarrow F^*$. Moreover, the same results also apply to $\{\tilde X_t\}$.
\label{thm:conv}
\end{theorem}
\begin{proof}
The coerciveness of $F$ follows directly from the fact that the nuclear norm is
coercive and that $f$ is lower-bounded.
This implies that the level sets of $F$ are bounded, and thus so is
$\Omega^*$.
Moreover, because $F$ is lower semicontinuous, it attains its minimum
confined to any compact set, so we see that $\Omega^*$ is nonempty.

From \cref{eq:sqsummable}, we have
		\begin{equation}
			\infty > c^2 \coloneqq \sum_{t=0}^\infty \epsilon_t^2.
			\label{eq:C}
		\end{equation}
    We first show that $\{X_t\}$ admits at least one limit point. From \cref{eq:inexact}, we have that there exists $G_t\in \R^{m\times n}$ such that
    \begin{equation}
        G_t\in \nabla f\left(\tilde X_t\right)+\frac{1}{\alpha_t}\left( X_{t+1}-\tilde
		X_t\right)+\partial \Psi(X_{t+1}),\quad  \norm{G_t}_F \leq \epsilon_t.
        \label{eq:T3_1}
    \end{equation}
From
the convexity of $\Psi(\cdot)$, we have
    \begin{align}
   \Psi(X_{t+1}) \leq \inprod{\Xi}{X_{t+1}-X} + \Psi(X),\quad \forall \Xi \in \partial \Psi(X_{t+1}),\quad \forall X.
        \label{eq:T3_2.5}
    \end{align}
	By combining \cref{eq:T3_2.5} with $X = \tilde X_t$,
	\cref{eq:back,eq:T3_1}, and defining $\gamma \coloneqq (1-\delta)/\alpha_{\max}$,
		we get the following inequality:
    \begin{align}
        F(X_{t+1})&\leq F(\tilde X_t)+\inprod{G_t}{X_{t+1}-\tilde X_t}
		-\frac{1 - \delta}{\alpha_t}\| X_{t+1}-\tilde X_t\|_F^2\notag \\
        &\leq F(\tilde X_t)+\| G_t\|_F \|X_{t+1}-\tilde X_t \|_F -\frac{1
		- \delta}{\alpha_{\max}}\| X_{t+1}-\tilde X_t\|_F^2\notag\\
        &\leq F(\tilde X_t)+\epsilon_t  \|X_{t+1}-\tilde X_t \|_F -\gamma\| X_{t+1}-\tilde X_t\|_F^2
		\label{eq:T3_3.-1}\\
        &\leq F(X_t)+\epsilon_t  \|X_{t+1}-\tilde X_t \|_F
		-\gamma\| X_{t+1}-\tilde X_t\|_F^2,
        \label{eq:T3_3}
    \end{align}
	where the last inequality is from \cref{eq:nonincrease}.
    \cref{eq:T3_3} implies that 
    \begin{equation}
        \gamma\|X_{t+1}-\tilde X_t\|_F^2\leq F(X_t)-F(X_{t+1})+\epsilon_t \| X_{t+1}-\tilde X_t\|_F.
        \label{eq:T3_4}
    \end{equation}
    By summing \cref{eq:T3_4} from $t=0$ to $t=k,$ we have that
    \begin{align}
        \gamma\sum_{t=0}^k\|X_{t+1}-\tilde X_t\|_F^2&\leq F(X_0)-F(X_{k+1})+ \sum_{t=0}^k \epsilon_t\| X_{t+1}-\tilde X_t\|_F\label{eq:T3_4.5} \\
        &\leq F(X_0)-F^*+\sum_{t=0}^k \epsilon_t\| X_{t+1}-\tilde X_t\|_F\notag\\
		&\leq F(X_0)-F^*+\sqrt{\sum_{t=0}^k \epsilon_t^2}\sqrt{\sum_{t=0}^k\|X_{t+1}-\tilde X_t\|_F^2}\label{eq:explain1}\\
		&\leq F(X_0)-F^*+c\sqrt{\sum_{t=0}^k\|X_{t+1}-\tilde X_t\|_F^2},
        \label{eq:T3_5}
    \end{align}
	where \cref{eq:explain1} is from the Cauchy-Schwarz inequality.
    By applying the quadratic formula to \cref{eq:T3_5}, we obtain that
    \begin{equation}
        \sqrt{\sum_{t=0}^k\|X_{t+1}-\tilde X_t\|_F^2}\leq \frac{c +\sqrt{c^2+4\gamma\left( F(X_0)-F^*\right)}}{2\gamma}.
        \label{eq:T3_6}
    \end{equation}
    This implies that for all $k \geq 0$, we get
    \begin{equation}
        \sum_{t=0}^k \epsilon_t \|X_{t+1}-\tilde X_t\|_F\leq
		\sqrt{\sum_{t=0}^k
	\epsilon_t^2}\sqrt{\sum_{t=0}^k\|X_{t+1}-\tilde X_t\|_F^2}\leq
	\frac{c^2+c\sqrt{c^2+4\gamma\left( F(X_0)-F^*\right)}}{2\gamma}.
        \label{eq:T3_7}
    \end{equation}
    Combining \cref{eq:T3_7} and \cref{eq:T3_4.5}, we have that
	$\{F(X_t)\}$ is upper-bounded. Then from the
	coerciveness of $F$, the sequence $\{ X_t\}$ is bounded, and thus
	it has at least one limit point.
    
    Next, we prove that $\dist(X_t,\Omega^*) \rightarrow 0$ by
	conrtadiction.
	Suppose this statement is false, then
    there exists $\sigma>0$ and a subsequence $\{X_{t_k}\}_{k}$
	such that $\dist(X_{t_k},\Omega)\geq\sigma>0$ for any $k$.
	Since $\{X_{t_k}\}$ is also a bounded sequence, we have that there
	exists a subsubsequence $\{X_{\ell_k}\} \subseteq \{X_{t_k}\}$ such that
    \begin{equation}
        X_{\ell_k}\rightarrow \widetilde{X^*}\notin \Omega^*.
        \label{eq:T3_8}
    \end{equation}
    From \cref{eq:T3_1}, we have that 
    \begin{equation}
        G_t+\nabla f(X_{t+1})-\nabla f(\tilde X_t)-\frac{1}{\alpha_t}(X_{t+1}-\tilde
		X_t)\in \partial F(X_{t+1}).
        \label{eq:T3_9}
    \end{equation}
	From \cref{eq:sqsummable}, we have $\epsilon_t\rightarrow 0$ and
	hence $G_t\rightarrow 0.$ From
	\cref{eq:T3_6},
	we have that $X_{t+1}-\tilde X_t \rightarrow 0.$ 
	This together with the Lipschitz continuity of $\nabla f$ and the
	boundedness of $\alpha_t$ implies that $\nabla f(X_{t+1})-\nabla
	f(\tilde X_t)\rightarrow 0$ and $ \alpha_t^{-1}(X_{t+1}-\tilde
	X_t) \rightarrow 0.$
	These results together imply that
    \begin{equation}
       G_t+\nabla f(X_{t+1})-\nabla f(\tilde X_t)-\frac{1}{\alpha_t}(X_{t+1}-\tilde
		X_t)\rightarrow 0.
        \label{eq:T3_10}
    \end{equation}
    From \cref{eq:T3_8,eq:T3_9,eq:T3_10} and the outer semi-continuity of
	$\partial F$ (see \cite[Proposition~8.7]{RCK09v} and
	\cite[Proposition~20.37]{BSK11c}) in \cref{eq:T3_9}, we have that $0\in \partial F\big(\widetilde{X^*}\big)$.
    From the convexity of $F$, we thus get $\widetilde{X^*}\in \Omega^*$,
	contradicting \cref{eq:T3_8}. Therefore, we conclude
	$\dist(X_t,\Omega^*)\rightarrow 0.$
	This also implies that any limit point of $\{X_t\}$ must lie in
	$\Omega^*$.

	Finally, we prove the convergence of $\{F(X_t)\}$. From the convexity of
	$F$ and \cref{eq:T3_9},
	we see that for any $X^* \in \Omega^*$,
	\begin{align}
		\notag
		0 &\leq  F\left( X_{t+1} \right) - F\left( X^*
		\right)\\
		\notag
		&\leq \inprod{G_t + \nabla f\left( X_{t+1} \right) -
		\nabla f\big( \tilde X_t \big) - \alpha_t^{-1} \big( X_{t+1} - \tilde
	X_t\big)}{X_{t+1} - X^*}\\
	&\leq \norm{ G_t + \nabla f\big( X_{t+1} \big) - \nabla f\big(
	\tilde X_t \big) - \alpha_t^{-1} \big( X_{t+1} - \tilde X_t\big)}_F\norm{X_{t+1}
- X^*}_F.
		\label{eq:upper}
\end{align}
By the boundedness of $\{X_t\}$, $\norm{X_{t+1} - X^*}_F$ is upper
bounded. Recall that the first norm term in \cref{eq:upper} approaches to
$0$ as shown in \cref{eq:T3_10}. Thus, $F(X_t) \rightarrow F^*$ by the sandwich lemma.

For the part of $\{\tilde X_t\}$, we see that the boundedness of the
iterates and the convergence of the objective value follow from
\cref{eq:nonincrease} and again the coerciveness of $F$.
From this boundedness we then conclude the existence of a limit point,
and convergence of $\dist(\tilde X_t, \Omega^*)$ follows from the same
argument above.
\end{proof}

Although our global convergence is guaranteed by the inexact PG step,
existing analyses for the inexact PG method, like those by
\cite{Com04a,SchRB11a,JiaST12a,HamWWM22a}, utilize the geometry of the
iterates, and are hence not applicable to our algorithm because
our additional BM phase could move the iterates arbitrarily within the
level sets and this may make such geometry properties no longer valid.
Therefore, another contribution of this work is in developing new proof
techniques for obtaining global convergence guarantees for alternating
between general nonmonotone inexact PG steps and some other descent
optimization steps.

We next provide convergence rate guarantees for \cref{alg:mfstride} in
the coming two theorems.
For such results, we use the definitions below.
	\[
		\triangle F_t\coloneqq F(X_t)-F^*,\quad \triangle \tilde{
		F_t}\coloneqq F(\tilde X_t)-F^*.
	\]
The following theorem and its proof are partially motivated by \cite{SchT16a}.
\begin{theorem}
	Suppose the conditions in  \cref{thm:conv} hold.
	Then there exists a constant $\beta > 0$ such that the following inequality holds:
    \begin{equation}\label{inecon}
		\triangle \tilde F_{t+1}\leq \max\left\{ \xi_t
			,\frac{\triangle \tilde{F}_t}{1+\beta^{-1}\triangle \tilde{F}_t
	}\right\},\quad
	\xi_t \coloneqq \sqrt{\beta \psi_t} + \psi_t,\quad
    \psi_t \coloneqq
				\epsilon_t(\|X_{t+1}-\tilde{X}_t\|_F+ \gamma
				\epsilon_t),
    \end{equation}
    where
				$\gamma\coloneqq (1-\delta)/\alpha_{\max}$.
	Moreover, if $\epsilon_t = O(t^{-2})$, 
		then $ \triangle F_t=O(t^{-1})$ and $ \triangle
		\tilde{F}_t = O(t^{-1})$.
\label{thm:rate}
\end{theorem}
\begin{proof}
	From the boundedness of $\{F(X_t)\}$ and $\{F(\tilde X_t)\}$
	obtained in the proof of \cref{thm:conv},
	we know that there is a value $F_0$ such that $F(X_t) \leq F_0$
	and $F(\tilde X_t) \leq F_0$ for all $t$, and thus from the
	coerciveness of $F$ from \cref{thm:conv}, there is a
	nonnegative and finite constant $R_0$ such that
		\begin{equation}
			R_0 \coloneqq \max_{X_1,X_2 \in \{X\mid F(X) \leq F_0\}}
			\norm{X_1 - X_2}_F < \infty.
			\label{eq:R0}
		\end{equation}
    Next, from the convexity of $f$, we have that for any $X\in \R^{m\times n},$
    \begin{align}
		\label{eq:T5_1}
    f(\tilde{X}_{t})& \leq f(X)+\inprod{\nabla
		f(\tilde{X}_t)}{\tilde{X}_t-X}.
    \end{align}
    Add up \cref{eq:back} and \cref{eq:T5_1}, we get
    \begin{equation}\label{eq:T5_3}
    f(X_{t+1})\leq f(X)+\inprod{\nabla f(\tilde{X}_t)}{X_{t+1}-X}+\frac{\delta}{\alpha_t}\| X_{t+1}-\tilde{X}_t\|_F^2.
    \end{equation}
    Add up \cref{eq:T3_2.5} and \cref{eq:T5_3}, we get
    \begin{align}
    F(X_{t+1})
	&\leq
	F(X)+\frac{1}{\alpha_t}\inprod{\tilde{X}_t-X_{t+1}}{X_{t+1}-X}+\frac{\delta}{\alpha_t}\|
	X_{t+1}-\tilde{X}_t\|_F^2+\inprod{G_t}{X_{t+1}-X}
	\label{eq:T5_4}
    \end{align}
	for $G_t$ defined in \cref{eq:T3_1}.
	Choose $X=X^*$ for some $X^* \in \Omega^*$ in \cref{eq:T5_4} and
	use \cref{eq:R0},
	we get
    \begin{align}
        F(X_{t+1})&\leq
		F(X^*) + \frac{R_0}{\alpha_t}
		\|\tilde X_t - X_{t+1}\|_F +
		\frac{\delta R_0}{\alpha_t}\|{\tilde X_t - X_{t+1}}\|_F +
		\epsilon_t R_0\nonumber\\
		&\leq F(X^*)+ \tilde c (\| X_{t+1}-\tilde{X}_t \|_F +\epsilon_t),
		\label{eq:T5_5}
    \end{align}
	where $\tilde c \coloneqq R_0(1 + \delta + \alpha_{\min})/\alpha_{\min}
	\in [0, \infty)$ is a constant.
	Note that $F$ is Lipschitz continuous in any bounded region, so
	we can further obtain from \cref{eq:T5_5} that
    \begin{equation}\label{eq:T5_6}
        F(\tilde{X}_{t})\leq F(X^*)+ \bar c (\| X_{t+1}-\tilde{X}_t
		\|_F+\epsilon_t),
    \end{equation}
    where $\bar c>0$ is a constant. From \cref{eq:T5_6}, we further get that
    \begin{equation}\label{eq:T5_6.5}
        (F(\tilde{X}_{t})-F(X^*))^2\leq 2\bar c^2\| X_{t+1}-\tilde{X}_t \|_F^2+2\bar c^2\epsilon_t^2,
    \end{equation}
    Substitute \cref{eq:T5_6.5} into \cref{eq:T3_3.-1} and use
	\cref{inecon}, we get
    \begin{equation}\label{eq:T5_7}
    F(\tilde{X}_{t+1})\leq F(X_{t+1})\leq F(\tilde{X_t})+\psi_t -\frac{\gamma}{2\bar c^2}(F(\tilde{X}_{t})- F(X^*))^2,
    \end{equation}
	Let $\beta=4\bar c^2/\gamma$, we then have the following two cases from
	\cref{eq:T5_7}:

    \noindent{\bf Case 1.} $
	 \psi_t >\beta^{-1} \triangle \tilde F_t^2.$\\
	We have that $\triangle \tilde{F_t}\leq \sqrt{\beta \psi_t}$.
	Combine this and \cref{eq:T5_7}, we get
    \begin{equation}\label{eq:T5_8}
    \triangle \tilde{F}_{t+1}\leq  \triangle F_{t+1}\leq \sqrt{\beta
	\psi_t }+\psi_t.
    \end{equation}

    \noindent{\bf Case 2.}
	$\psi_t \leq \beta^{-1}\triangle \tilde F_t^2.$\\
	Substitute this into
	\cref{eq:T5_7}, we get 
    \begin{equation}\label{eq:T5_9}
    \triangle \tilde{F}_{t+1}\leq \triangle F_{t+1}\leq \triangle \tilde{F}_{t}-\frac{1}{\beta}\triangle \tilde{F}_t^2\leq \triangle \tilde{F}_t.
    \end{equation}
    From \cref{eq:T5_9}, we have that $\triangle \tilde{F}_{t+1}\leq
	\triangle \tilde{F}_t-\beta^{-1} \triangle \tilde{F}_t \triangle \tilde{F}_{t+1},$ which implies that
    \begin{equation}\label{eq:T5_10}
    \triangle \tilde{F}_{t+1}\leq \frac{\triangle
		\tilde{F}_t}{1+\beta^{-1}\triangle \tilde{F}_t}.
    \end{equation}
    Combine \cref{eq:T5_8} and \cref{eq:T5_10}, we get \cref{inecon}.

	Now, assume that $\epsilon_t = O(t^{-2})$.
		From \cref{inecon,eq:R0}, we see that $\xi_t =	O(t^{-1})$.
	Namely, there exists $\kappa\geq 0$ such that $\xi_t\leq \kappa/t$ for
	all $t\geq 1$.
	For any $t\in \N$, we first consider the
	case in which there is some index
	$\tilde t_0 \in \{1,\dots, t\}$ such that the first term in \cref{inecon} is larger than the second one and let $t_0$ be the maximum of such
	indices. Thus, for any $k\in \{t_0+1,\ldots,t\},$ we have that
	$\triangle \tilde{F}_{k+1}\leq \triangle \tilde{F}_k/
	(1+\beta^{-1}\triangle \tilde{F}_k)$, which implies that
	\[
		\frac{1}{\triangle \tilde{F}_k}+\frac{1}{\beta}\leq
		\frac{1}{\triangle \tilde{F}_{k+1}}.
	\]
	Summing the inequality above from $k=t_0+1$ to $k=t$ and
	telescoping, we have that 
    \begin{equation}\label{eq:T5_11}
		\frac{1}{\triangle \tilde{F}_{t+1}}\geq \frac{1}{\triangle
			\tilde{F}_{t_0+1}}+ \frac{(t-t_0)}{\beta}\geq
			\frac{t_0}{\kappa}+\frac{t-t_0}{\beta}\geq \frac{t}{\max\{\kappa,\beta\}},
    \end{equation}
    which implies
    \begin{equation}\label{eq:T5_12}
		\triangle \tilde{F}_{t+1}\leq \frac{\max\{\beta,\kappa\}}{t} =
		O(t^{-1}).
    \end{equation}

	Now let us turn to the case in which the second term in
	\cref{inecon} is larger than the first one for all $k\in
	\{1,\ldots,t\}$.
%
Then an analysis analogous to \cref{eq:T5_11} leads to the following
inequality:
    \begin{equation}\label{eq:T5_14}
		\triangle \tilde{F}_{t+1}\leq \frac{1}{\triangle
			\tilde{F}_1^{-1}+\beta^{-1} t} = O(t^{-1}).
    \end{equation}
    Combine \cref{eq:T5_12} and \cref{eq:T5_14}, we obtain that
	$\triangle \tilde{F}_{t}=O(t^{-1}).$
	Finally, viewing from \cref{eq:T5_7} and the optimality of $F^*$,
	we get from \cref{eq:T5_12}, \cref{eq:T5_14}, and $\psi_t = O(t^{-1})$ that
		$\Delta F_{t+1} \leq \Delta \tilde F_t + \psi_t = O(t^{-1})$.
\end{proof}



In the next result, we show faster convergence rates under a
H\"olderian error-bound condition
\begin{equation}
\zeta \dist(X,\Omega^*) \leq \left(F(X) - F^*\right)^{\theta},\quad
\forall X
\label{eq:sharpness}
\end{equation}
for some $\zeta > 0$ and some $\theta \in [0,1]$.
In particular, when $\theta \geq 1/2$, we obtain linear convergence
for the objective.
Under convexity of $F$, it is shown by \cite{Bol17a} that \cref{eq:sharpness}
is equivalent to the Kurdyka-{\L}ojasiewicz (KL) condition
\citep{Kur98a,Loj63a}.

\begin{theorem}\label{linearT}
	Consider \cref{eq:matrixform} with $\Psi$ defined in
	\cref{eq:Psi}.
	Suppose the line-search criterion \cref{eq:decrease} is used with
	$\delta \in (0,1)$ in \cref{alg:mfstride}. If $F$ satisfies \cref{eq:sharpness}, then the
	following convergence results hold.
    \begin{enumerate}[label=(\roman*),leftmargin=*]
		\itemsep 0pt
		\topsep 0pt
    \item When $\theta \geq 1/2$:
		Let
		\begin{equation}
			M\coloneqq \min_{\mu\in
			[0,1]}1-\delta\mu+\frac{\delta\mu^2}{2\zeta^2\alpha_{\min}} <
			1.
			\label{eq:Mdef}
		\end{equation}
		If 
			\begin{equation}
				\sum_{t=1}^\infty \frac{\epsilon_t^2}{M^{t}}<\infty,
				\label{eq:epsbound}
			\end{equation}
		then $\triangle \tilde{F}_t=O(M^t).$
    \item When $0\leq \theta<1/2$:
		If 
		\begin{equation}
			\epsilon_t^2=O\left(t^{-\frac{2-2\theta}{1-2\theta}}\right),
			\label{eq:epsbound2}
		\end{equation}
		then
		\[\triangle \tilde{F}_t=O\left(t^{-\frac{1}{1-2\theta}}\right).
		\]
    \end{enumerate}
\end{theorem}

\begin{proof}
Let $X_t^*$ and $Q_t^*$ be the unique minimizer and minimum of
$Q_t^{\alpha_t}(Y)$ respectively. Because $Q_t^{\alpha_t}(Y)$ is a
strongly-convex function with modulus $\alpha_t^{-1}$, from \cref{eq:inexact}, there exist $G_t$ such that
\begin{equation}
	\frac{1}{\alpha_t}\|X_{t+1}-X^*_t\|_F\leq \|G_t\|_F\leq \epsilon_t.
\label{eq:C5-1}
\end{equation}
From \cref{eq:decrease}, we have that
\begin{align}
F(X_{t+1})-F(\tilde{X}_t)&\leq \delta Q^{\alpha_t}_t(X_{t+1})\notag \\
&\leq \delta(Q_t^*+\inprod{G_t}{X_{t+1}-X_t^*})\notag \\
&\leq \delta(Q_t^*+\|G_t\|_F\|X_{t+1}-X_t^*\|_F)\notag \\
&\leq \delta(Q_t^*+\alpha_t \epsilon_t^2),\label{eq:C5-2}
\end{align}
where the second
inequality comes from the convexity of $Q_t^{\alpha_t}(\cdot)$, and the
last one comes from \cref{eq:C5-1}.
From Lemma~5 in \cite{LeeW18a} \citep[also see Equation~70
of][]{Lee20a}, we have that
\begin{equation}
Q^*_t\leq \mu(F^*-F(\tilde{X}_t))+\frac{\mu^2}{2\alpha_t}\dist(\tilde{X}_t,\Omega^*)^2,
\quad \forall \mu \in [0,1].
\label{eq:C5-2.5}
\end{equation}
Substitute \cref{eq:C5-2.5} into \cref{eq:C5-2} and use \cref{eq:sharpness}, we get
\begin{align}
 F(\tilde{X}_{t+1})-F(\tilde{X}_t)&\leq F(X_{t+1})-F(\tilde{X_t}) \notag\\
 &\leq \delta\min_{\mu \in [0,1]}\left( \mu(F^*-F(\tilde{X}_t))+\frac{\mu^2}{2\alpha_t}\dist(\tilde{X}_t,\Omega^*)^2+\alpha_t\epsilon_t^2 \right)\notag \\
 &\leq \delta\min_{\mu \in [0,1]}\left( \mu(F^*-F(\tilde{X}_t))+\frac{\mu^2}{2\zeta^2\alpha_t }(F(\tilde{X}_t)-F^*)^{2\theta}+\alpha_t\epsilon_t^2 \right).
\label{eq:C5-4} 
\end{align}

\noindent{\bf Proof of (i).}
We first consider the case of $\theta = 1/2$.
Let \[
	M_t\coloneqq \min_{\mu\in
	[0,1]}1-\delta\mu+\frac{\delta\mu^2}{2\zeta^2\alpha_t},\quad \forall t
\geq 0.\]
	Because $0<\delta<1$ and $\alpha_t \leq \alpha_{\max}$, we have
	that $0<M_t<M<1$ for all $t\geq 0$.
	From \cref{eq:C5-4} and the assumption that $\theta = 1/2$, we get
\begin{equation}
\triangle \tilde{F}_{t+1}/M^{t+1}\leq \triangle \tilde{F}_t/M^t+\delta\alpha_t\epsilon_t^2/M^{t+1}.
\label{eq:C5-6}
\end{equation}
Because $\epsilon_t^2/M^{t}$ is summable from \cref{eq:epsbound} and $\alpha_t \leq
\alpha_{\max} < \infty$, \cref{eq:C5-6} clearly shows $\triangle
\tilde{F}_{t}=O(M^t)$.

Now for $\theta > 1/2$, since the level sets are bounded and so is
$\Omega^*$, by changing $\zeta$ if necessary, we know that
\cref{eq:sharpness} with $\theta = 1/2$ still holds within the region
of interest (as \cref{eq:C5-2,eq:decrease,eq:epsbound} indicate that
the iterates stay within a bounded level set).
The result therefore follows from our analysis for $\theta = 1/2$.

\noindent{\bf Proof of (ii).}
From \cref{eq:C5-4}, we have
\begin{equation}
\triangle \tilde{F}_{t+1}\leq \min_{\mu\in
[0,1]}\left\{1-\delta\mu+\frac{\delta \mu^2}{2\zeta^2\alpha_t}\triangle
\tilde{F}_t^{2\theta-1}\right\} \triangle \tilde{F}_t+\delta\alpha_t\epsilon_t^2.
\label{eq:C5-7}
\end{equation}
Clearly, the minimizer of $\mu$ in \cref{eq:C5-7} is $\min\{\zeta^2\alpha_t \triangle \tilde{F}_t^{1-2\theta},1\}.$ Substitute this into \cref{eq:C5-7}, we get
\begin{equation}
\triangle \tilde{F}_{t+1}\leq \left( 1-\frac{\delta \zeta^2\alpha_t}{2}\min\left\{ \triangle\tilde{F}_t^{1-2\theta},\frac{1}{\zeta^2\alpha_t} \right\} \right)\triangle \tilde{F}_t+\delta\alpha_t\epsilon_t^2.
\label{eq:C5-8.0}
\end{equation}

We first show that $\lim\inf_{t\rightarrow \infty} \triangle
\tilde{F}_t=0.$ Assume on the contrary that there exists $\eta>0$ such
that $\triangle \tilde{F}_t\geq \eta$ for all $t \geq 0$.
Then, from \cref{eq:C5-8.0}, we have that
\[
\triangle \tilde{F}_{t+1}\leq \left( 1-\frac{\delta \zeta^2\alpha_t}{2}\min\left\{ \eta^{1-2\theta},\frac{1}{\zeta^2\alpha_t} \right\} +\frac{\delta\alpha_t \epsilon_t^2}{\eta}\right)\triangle \tilde{F}_t,
\]
which implies that $\triangle \tilde{F}_{t}\rightarrow 0$
with a linear rate when $t$ is sufficiently large to make $\epsilon_t$
small enough. This contradicts to
 $\triangle \tilde{F}_t\geq \eta>0$.
 Now, since $\lim\inf_{t\rightarrow
\infty} \triangle \tilde{F}_t=0$ and $\sum_{t=1}^\infty \alpha_t
\epsilon_t^2<\infty,$ there exists $T_0 \geq 0$ such that
$\triangle \tilde{F}_{T_0}\leq C_1$
and
$\sum_{t=T_0}^\infty \delta
\alpha_t\epsilon_t^2<C_1$,
where
\[
	C_1 \coloneqq \frac12 (\zeta^2\alpha_{\max})^{-\frac{1}{1-2\theta}}.
\]
From \cref{eq:C5-8.0}, we thus get
\begin{equation}
	\triangle \tilde{F}_t\leq \triangle \tilde{F}_{T_0} +
	\sum_{k=T_0}^{t-1} \delta \alpha_t \epsilon_t^2 <
2 C_1\leq
(\zeta^2\alpha_{t})^{-\frac{1}{1-2\theta}}, \quad \forall t \geq T_0.
\label{eq:C5-7.5}
\end{equation}
Thus,
\[
	\min\left\{ \triangle
	\tilde{F}_t^{1-2\theta},\frac{1}{\zeta^2\alpha_t}
\right\}=\triangle \tilde{F}_t^{1-2\theta},\quad \forall t \geq T_0.
\]
Let $M_1\coloneqq\delta \zeta^2\alpha_{\min}/2$, $M_2\coloneqq\delta\alpha_{\max}$,
then from the equation above and \cref{eq:C5-8.0}, we have that
\begin{equation}
\triangle \tilde{F}_{t+1}\leq \big( 1-M_1
\triangle\tilde{F}_t^{1-2\theta}\big)\triangle
\tilde{F}_t+M_2\epsilon_t^2, \quad \forall t \geq T_0.
\label{eq:C5-9}
\end{equation}
From \cref{eq:C5-7.5} and  that $0<\delta<1,$ we have that
$M_1\triangle \tilde{F}^{1-2\theta}_t<1$ for any $t\geq T_0$.
Now we choose $D>0$ to be a sufficiently large number such that the following three conditions hold.
\begin{subequations}
	\begin{align}
		\label{eq:conda}
\triangle \tilde{F}_{T_0}&\leq DT_0^{\frac{-1}{1-2\theta}}, \\
		\label{eq:condb}
	M_2\epsilon_t^2+\left(\frac{2M_2}{M_1}\epsilon_t^2\right)^{\frac{1}{2-2\theta}}&\leq
	D(t+1)^{\frac{-1}{1-2\theta}},\quad \forall t \geq T_0,\\
		\label{eq:condc}
	D^{-(1-2\theta)}&\leq \frac{(1-2\theta) M_1}{2},
	\end{align}
\end{subequations}
where \cref{eq:condb} is guaranteed by
\cref{eq:epsbound2} and
$\theta < 1/2$, and \cref{eq:condc} can be guaranteed by $\theta<1/2$.
Now, we use
mathematical induction to prove that $\triangle \tilde{F}_t\leq
Dt^{\frac{-1}{1-2\theta}}$ for all $t\geq T_0.$ The case $t=T_0$
directly comes from \cref{eq:conda}. Suppose the inequality holds for some $t\geq T_0,$ we have the following two cases.

\noindent{\bf Case 1.} $M_2\epsilon_t^2\leq M_1\triangle
\tilde{F}^{2-2\theta}_t/2$.\\
From \cref{eq:C5-9}, we have that
\begin{equation*}
\triangle \tilde{F}_{t+1}\leq \left( 1-\frac{M_1}{2} \triangle\tilde{F}_t^{1-2\theta}\right)\triangle \tilde{F}_t,
\end{equation*}
which leads to
\begin{align}
\triangle\tilde{F}_{t+1}^{-(1-2\theta)}&\geq \left( 1-\frac{M_1}{2}\triangle \tilde{F}_t^{1-2\theta} \right)^{-(1-2\theta)}\triangle \tilde{F}_t^{-(1-2\theta)}\notag \\
&\geq \left(1+\frac{(1-2\theta)M_1}{2}\triangle
\tilde{F}_t^{1-2\theta}\right)\triangle \tilde{F}_t^{-(1-2\theta)}\notag\\
&=\triangle \tilde{F}_t^{-(1-2\theta)}+\frac{(1-2\theta)M_1}{2}\notag \\
&\geq D^{-(1-2\theta)}t+\frac{(1-2\theta)M_1}{2} \notag \\
&\geq D^{-(1-2\theta)}(t+1),
\label{eq:C5-10}
\end{align}
where the second inequality comes from the fact that $(1-x)^{-p}\geq 1+px$ for
any $x<1$ and $p>0$, and  the last inequality comes from
\cref{eq:condc}. \cref{eq:C5-10} implies that $\triangle \tilde{F}_{t+1}\leq D(t+1)^{\frac{-1}{1-2\theta}}.$

\noindent{\bf Case 2.} $M_2\epsilon_t^2 > M_1\triangle
\tilde{F}^{2-2\theta}_t/2$.\\
In this case, we have that
\begin{equation}
\triangle \tilde{F}_t\leq \left(\frac{2M_2}{M_1}\epsilon_t^2\right)^{\frac{1}{2-2\theta}}.
\label{eq:C5-11}
\end{equation}
By substituting \cref{eq:C5-11} into \cref{eq:C5-9}, we obtain
\begin{equation}
\triangle \tilde{F}_{t+1}\leq \triangle \tilde{F}_t+M_2\epsilon_t^2\leq \left(\frac{2M_2}{M_1}\epsilon_t^2\right)^{\frac{1}{2-2\theta}}+M_2\epsilon_t^2\leq D(t+1)^{\frac{-1}{1-2\theta}},
\end{equation}
where the last inequality comes from \cref{eq:condb}.

Combining Cases 1 and 2, we get $\triangle \tilde{F}_t\leq
Dt^{\frac{-1}{1-2\theta}}$ for any $t\geq T_0$, as desired.
\end{proof}
By setting $\theta = 0$ in \cref{linearT}, we recover the same
convergence rate in \cref{thm:rate}, but instead of $\epsilon_t
=O(t^{-2})$ in \cref{thm:rate}, \cref{linearT} only needs $\epsilon_t
= O(t^{-1})$.
The difference between the two theorems is that \cref{thm:rate} uses
\cref{eq:back} that allows a more aggressive step size selection but
with the price of a higher accuracy in the PG step, while
\cref{linearT} uses \cref{eq:decrease} that leads to a more
conservative step size to trade for less time spent on computing the
approximate SVD. Moreover, for \cref{linearT}, \cref{eq:sharpness} with $\theta = 0$ directly assumes that the iterates are bounded.

\subsection{Rank identification}
We proceed on to show that under a nondegeneracy condition, the
rank of $X_t$ for any convergent subsequence will eventually become
fixed and equivalent to the point of convergence.
First, we need the definition below of convex partly smooth functions.
This definition involves the usage of $\mathcal{C}^2$-manifold,
which means the system of equations defining such a manifold is
$\mathcal{C}^2$.
\begin{definition}[Partly smooth \citep{Lew02a}]
\label{def:ps}
A convex function $\Psi$ is partly smooth at a point $X^*$ relative
to a set $\M$ containing $X^*$ if $\partial \Psi(X^*) \neq \emptyset$
and:
\begin{enumerate}
		\itemsep 0pt
		\topsep 0pt
	\item Around $X^*$, $\M$ is a $\mathcal{C}^2$-manifold and
		$\Psi|_{\M}$ is $\mathcal{C}^2$.
	\item The affine span of $\partial \Psi(X)$ is a translate
		of the normal space to $\M$ at $X^*$.
	\item $\partial \Psi$ is continuous at $X^*$ relative to $\M$.
\end{enumerate}
\end{definition}
Loosely speaking, this means that $\Psi|_{\M}$ is smooth at $x^*$,
but the value of $\Psi$ changes drastically along directions leaving
$\M$ around $x^*$.

It is known \citep{DanDL14a} that at every $X \in \R^{m \times n}$,
$\norm{\cdot}_*$ is partly smooth with respect to the manifold
	\begin{equation}
		\label{eq:M1}
		\M(X) \coloneqq \left\{ Y \in \R^{m\times n}\mid \rank(Y) =
	\rank(X) \right\}.
\end{equation}
Similarly, if $X \in \SS^n$, we also have that
$\delta_{\SS^n_+}$ is partly smooth everywhere in $\SS^n_+$, with
respect to the manifold
	\begin{equation}
		\label{eq:M2}
	\M_2(X) \coloneqq \left\{ Y \in \SS^n_+ \mid \rank(Y) =
	\rank(X) \right\}.
\end{equation}
Finally, when $\mathcal{S}$ is a polyhedron, it is widely known that
$\delta_{\mathcal{S}}$ is also partly smooth everywhere, with respect to the
minimal face containing the reference point.
As intersections of manifolds are still manifolds, if
\begin{equation}
	\Psi(X) = \lambda \norm{X}_* + \lambda_2 \delta_{\SS^n_+}(X) +
	\lambda_3 \delta_{\mathcal{S}}(X)
	\label{eq:Psi2}
\end{equation}
for $\lambda \in \R$, 
$\lambda_2, \lambda_3 \in \{0,1\}$ and some polyhedral
$\mathcal{S}$, we have that $\Psi$ is partly smooth everywhere, with respect to
a submanifold $\bar \M(X) \subseteq \M(X)$.

Now we can leverage tools from partial smoothness and manifold
identification to show that our algorithm will find the correct rank
for \cref{eq:bm} that contains a global optimum.

\begin{theorem}
	Consider \cref{eq:matrixform} with $\Psi$ defined in \cref{eq:Psi}.
	Consider the two sequences of iterates $\{X_t\}$ and $\{\tilde X_t\}$ generated by
	\cref{alg:mfstride} from some starting point $X_0 = (W_0, H_0)$
	with $\epsilon_t \rightarrow 0$ in \cref{eq:inexact}.
	Then the following hold.
    \begin{enumerate}[leftmargin=*, label=(\roman*)]
	\topsep 0pt
	\partopsep 0pt
	\parsep 0pt
		\item For any subsequence $\{\tilde X_{t_i}\}_i$ such that
			$\tilde X_{t_i} \rightarrow X^*$ for some $X^* \in
			\Omega^*$, $X_{t_i+1} \rightarrow X^*$ as
			well.
		\item For the same subsequence as above,
			if $X^*$ satisfies the nondegeneracy condition
			\begin{equation}
				0 \in \relint\left( \partial F\left( X^* \right)
				\right)
				\label{eq:nod}
			\end{equation}
			and $\Psi$ is as defined in \cref{eq:Psi} with either
			$\lambda \geq 0$ or $\lambda < 0$ and $\Psi$ accords with
			\cref{eq:Psi2},
			then there is $i_0 \ge 0$ such that $\rank(X_{t_i+1}) =
			\rank(X^*)$ for all $i \geq i_0$.
	\end{enumerate}
	\label{thm:identify}
\end{theorem}
\begin{proof}

\noindent \textbf{Proof of (i).}
		Let us denote the exact solution of \cref{eq:proxgrad}
			at the $t_i$-th iteration given $\tilde X_{t_i}$ as $
			X_{t_i+1}^*$, and the real update we use from
			\cref{eq:inexact} as $X_{t_i+1}$.
			Following the proof of \cite[Proposition~1]{YueZS19a}, we have
			from the optimality of $X^*$, which implies $\prox_{\alpha
				\Psi}(X^* - \alpha \nabla f(X^*)) =
				X^*$ for any $\alpha > 0$, that
				\begin{align}
					\nonumber
					& \;\norm{X_{t_i}^* - \tilde X_{t_i}}_F
					\\
					\nonumber
					=&\;
					\norm{\prox_{\alpha_{t_i} \Psi}\left(
					\tilde X_{t_i} -  \alpha_{t_i} \nabla f(\tilde
					X_{t_i}\right) -
						\tilde X_{t_i} +
						\left(X^* - \prox_{\alpha_{t_i}
						\Psi}(X^* - \alpha_{t_i} \nabla
				f(X^*))\right)}_F\\
					\nonumber
					\leq &\; \norm{\tilde X_{t_i} - X^*}_F +
					\norm{\prox_{\alpha_{t_i} \Psi}\left( \tilde X_{t_i} -  \alpha_{t_i} \nabla
					f(\tilde X_{t_i})\right) - \prox_{\alpha_{t_i}
						\Psi} \left( X^* -
						\alpha_{t_i} \nabla f(X^*) \right)}_F \\
					\label{eq:intermediate1}
					\leq &\; \norm{\tilde X_{t_i} - X^*}_F + \norm{
						\left( \tilde X_{t_i} -  \alpha_{t_i} \nabla
						f(\tilde X_{t_i})\right) - \left( X^* -
						\alpha_{t_i} \nabla f(X^*) \right)}_F \\
					\nonumber
					\leq &\;  2\norm{\tilde X_{t_i} - X^*}_F + \alpha_{t_i} \norm{
					\nabla
					f(\tilde X_{t_i}) -
				\nabla f(X^*)}_F \\
				\label{eq:intermediate2}
					\leq &\;
					(2 + L \alpha_{\max})
					\norm{\tilde X_{t_i} - X^*}_F \rightarrow 0,
				\end{align}
				where \cref{eq:intermediate1} is from the
				nonexpansiveness of the proximal operation of any
				convex function, and \cref{eq:intermediate2} is from
				our assumption.
				\cref{eq:intermediate2} then leads to
			\begin{equation}
				\norm{X_{t_i}^* - \tilde
				X_{t_i}}_F \rightarrow 0.
				\label{eq:bound1}
			\end{equation}
			On the other hand, \cref{eq:C5-1} shows that
			\begin{align}
				0 \leq \norm{X_{t_i+1} - X_{t_i}^*}_F
					\leq \alpha_{t_i} \epsilon_{t_i} \rightarrow 0,
			\label{eq:bound2}
		\end{align}
		where the limit is obtained from that $\epsilon_t \rightarrow
		0$ and that $\alpha_t$ is upper-bounded.
		By combining \cref{eq:bound1,eq:bound2}, it is clear that
		\begin{equation}
			\norm{X_{t_i+1} - \tilde X_{t_i}} \rightarrow 0 \quad
			\Rightarrow \quad
			\norm{X_{t_i+1} - X^*} \rightarrow 0,
			\label{eq:bound3}
		\end{equation}
		proving the desired result.

\noindent \textbf{Proof of (ii).}
	From our arguments preceding the theorem, $\Psi$ is partly
		smooth at every $X$ relative to a submanifold of either
		\cref{eq:M1} or \cref{eq:M2}.
		Therefore, the result of the second item is equivalent to
		$X_{t_i+1} \in \M(X^*)$ for all $i$ large enough.
		As $\epsilon_t \rightarrow 0$, we see that
			all conditions of \cite[Theorem~1]{Lee20a} are satisfied,
			and therefore $X_{t_i+1} \in \bar \M(X^*) \subseteq
			\M(X^*)$ for all $i$ large enough.
	The conclusion therefore follows.
\end{proof}
%
%
Due to the flexibility in the BM step, we have less control over the
iterates than ordinary PG methods.
Therefore, convergence of the whole sequence of iterates cannot be
directly guaranteed and we can only get subsequential convergence.
However, in our experiments in \cref{sec:exp}, we often observe
empirically that the iterates are convergent to a point, and the rank
always becomes fixed after a few iterations of \ours.

\section{Applications}
\label{sec:application}
We provide two applications of \cref{eq:matrixform}.
One is our motivating example of matrix completion with $\cX$ being
the whole space,
and the other one is a special class of convex quadratic semidefinite
programming problems.

\subsection{Matrix Completion}
\label{sup:mc}
Our first application of \cref{eq:matrixform} is the low-rank matrix
completion problem \cref{eq:mc}.
This problem is widely seen in many machine learning tasks like
recommendation systems, localization in Internet of Things (IoTs),
and image denoising and compression.
Interested readers are referred to a recent survey \cite{NguKS19a}
for more details of these applications.
\blue{A common feature for many of these tasks is that the observed
	data are extremely sparse in comparison to the unobserved entries
	that we aim to predict.
	In other words, $|\Omega| \ll mn$, and thus the resulting gradient
	of the smooth part $\nabla f$ is also sparse, as it can
	be nonzero only at those entries in $\Omega$.
	\Cref{tbl:data} provides some examples of the sparsity level of
$\Omega$ in real-world data used in our numerical experiments.}

We observe that the loss term of \cref{eq:mc} has an $L$-Lipschitz
continuous gradient with $L=1$, and when $\alpha_t = L^{-1} = 1$, $X_t - \alpha_t \nabla
f(X_t)$ is the same as replacing the entries of $X_t$ in $\Omega$ with
$P_{\Omega}(A)$, hence standard PG with
$\alpha_t \equiv L^{-1}$ is also called soft impute for this
problem \citep{MazHT10a}.
Often in real applications described above, we can easily have $m$ and
$n$ in the scale of millions with $|\Omega|$ rather small, so indeed
we are unable to explicitly form $X_t$ and need to rely on low-rank
assumptions or to force low-rank approximations for practical reasons.
On the other hand, thanks to the extreme popularity and the simple
forms of \cref{eq:mc} and \cref{eq:mf}, there are many well-developed
algorithms for them.

Theoretical analyses for \cref{eq:mc} and \cref{eq:mf} often consider
the noiseless case such that the ground truth $A$ is indeed of low
rank and we observe entries without any noise, and show that under
such cases, one can recover the whole $A$ by solving \cref{eq:mf} with
a sufficient rank.
However, in practice, the observed entries are often noisy, either due
to measurement errors (like in the IoTs case) or randomness in nature
(rating in recommendation systems could be affected by factors beyond
the users' preference for certain items).
We will see in the numerical experiments in \cref{sec:exp} that it is
often the case that solving \cref{eq:mf} alone does not guarantee convergence
to a global optimum even if the correct rank is specified, and
therefore the convex lifting step in \cref{alg:mfstride} is necessary.

For this problem, in the convex lifting step, we adopt a long-step
variant of PG by setting $\alpha_t$ close to $2
L^{-1}$ to obtain a slightly better empirical performance.
For the BM stage, we adopt the state-of-the-art solver polyMF-SS for
\cref{eq:mf} developed by \cite{WanLK17a}
that conducts block coordinate descent with an exact line
search, where each block is one column of $W$ and one column of $H$.

More implementation details of our algorithm tailored for this
application are described in the supplementary materials.

We note that this problem satisfies \cref{eq:sharpness} with $\theta =
1/2$ according to \cite{HouZSL13a}, and thus linear convergence is
expected according to \cref{linearT}.

\subsection{A class of convex quadratic semidefinite programming problems}
\label{sec:app-qsdp}

Our second application is the following convex quadratic semidefinite programming (QSDP) problem:
\begin{equation}
    \label{eq-qsdp}
	\tag{QSDP}
    \min_{X\in \SS^n_+}\; \left(f(X)\coloneqq
	\frac{1}{2}\norm{\mathcal{A}(X) - b}^2 +  \inprod{C}{X}\right)\quad
	\mathrm{s.t.}\quad \inprod{E}{X} = 0,
\end{equation}
where $\mathcal{A}:\SS^n\to \mathbb{R}^p$ is a linear mapping whose adjoint mapping is denoted by $\mathcal{A}^*$, $b\in \mathbb{R}^p$, $C \in \mathbb{S}^n$, and $E\in \SS^n$ denotes the matrix of all ones. The gradient of $f$ is given by
\begin{equation}
	\nabla f(X) = \mathcal{A}^*\left(\mathcal{A}(X) - b\right) + C. 
	\label{eq:grad}
\end{equation}
Thus $\nabla f (\cdot)$ is Lipschitz continuous with modulus $L= \norm{\mathcal{A}^*\mathcal{A}}_2$.

The QSDP problem \cref{eq-qsdp} arises in many important applications
when one needs to find a low-rank approximation of a given matrix
while preserving certain useful structures (via linear constraints).
In this part, we introduce the following two data analysis problems.
\begin{itemize}
    \item The regularized kernel estimation (RKE) problem
		\citep{lu2005framework}: Given a set of $n$ objects and
		dissimilarity measures $d_{ij}^2$ for certain object pairs
		$(i, j)\in \Omega$, the goal of RKE is to find a positive semidefinite matrix $X$ such that the fitted squared distances between the objects induced by $X$ satisfy
    \[
        X_{ii}+X_{jj}-2X_{ij} \;\approx\; d_{ij}^2, \quad \forall
		(i,j) \in \Omega.
    \]
    To obtain a low-rank solution for $X$, the following regularized     semidefinite least squares problem is often considered:
		\begin{equation}
			\min_{X\in \SS^n_+}\, \frac{1}{2}\sum_{(i,j)\in\Omega}
			w_{ij} (X_{ii}+X_{jj}-2X_{ij}-d_{ij}^2)^2 + \lambda
			\inprod{I}{X} \quad \mathrm{s.t.}\quad  \inprod{E}{X} = 0,
			\label{eq:QSDP}
		\end{equation}
    where $\lambda>0$ is a positive regularization parameter and     $w_{ij} > 0$ for any $(i,j)\in\Omega$.  In the above, the constraint $\inprod{E}{X}=0$ is a normalization to put the center of mass of the realized Euclidean embedding at the origin. 
	As argued in \cref{sec:prelim}, $\inprod{I}{X}$ is equivalent to
	the nuclear norm for $X \in \SS^n_+$, so \cref{eq:QSDP}
	induces low-rank solutions.
    \item The molecular conformation problem \citep{fang2013using}:
		Given a molecule with $n$ atoms and the estimated inter-atomic distances $d_{ij}$ between some pairs $(i,j)\in \Omega$ of atoms, the goal is to determine the positions $x_1,\dots, x_n\in \mathbb{R}^d$ of all the atoms. Mathematically, the molecular conformation problem can be stated as follows:
    \[
        \min_{x_i\in \mathbb{R}^d, 1\leq i\leq n}\;  \frac{1}{2}\sum_{(i,j)\in\Omega} w_{ij} \left(\norm{x_i - x_j}^2-d_{ij}^2\right)^2 - \frac{\rho}{2n}\sum_{i, j = 1}^n \norm{x_i - x_j}^2\quad \mathrm{s.t.}\quad \sum_{i = 1}^nx_i = 0,
    \]
    where $w_{ij} > 0$ for all $ (i,j)\in \Omega $ and the second term involving $\rho > 0$ is used to maximize the pairwise separations between atoms. Define the matrix 
    \[
        X\coloneqq 
        \begin{bmatrix}
            x_1 & \dots & x_n 
        \end{bmatrix}^\top  
        \begin{bmatrix}
            x_1 & \dots & x_n 
        \end{bmatrix} \in \SS^n,
    \]
    then, it is easy to check that
	\[
		\norm{x_i-x_j}^2 = X_{ii} + X_{jj}
	- 2X_{ij},\quad \sum_{i, j = 1}^n \norm{x_i - x_j}^2 =
	2n\inprod{I}{X},
\]
	and the constraint $\sum_{i = 1}^nx_i = 0$ can
	be replaced by $\inprod{E}{X} = 0$. We therefore get the same QSDP
	relaxation \cref{eq:QSDP} for the molecular conformation problem
	with $\lambda\coloneqq -\rho < 0$.

	Although $\lambda < 0$ in this application, the problem is still
	in the form \cref{eq:matrixform} with a regularizer in
	\cref{eq:Psi}, so the objective function is still partly smooth
	everywhere with respect to \cref{eq:M2}.
	Hence, our algorithm will eventually generate iterates
	that have the same rank as the global optimum to which the iterates
	converge.
	If this optimum is low-rank, then so will the generated iterates
	be.
\end{itemize}

\blue{In the applications above, we see that for $\nabla f$ in
\cref{eq:grad}, $C$ is a sparse and structured matrix (actually the
identity matrix) and $\mathcal{A}$ and $\mathcal{A}^*$ are sparse
mappings such that only $\nabla_{i,j} f$ with either $(i,j) \in \Omega$ or $i
= j$ could be nonzero.}
We usually have $p \ll n^2$ in applications, and thus the resulting
gradient has a sparse part.
Driven by the fruitful and important applications of QSDPs in diverse fields, many
efficient algorithms for solving them have been developed.
We refer the readers to \cite{li2018qsdpnal} for a comprehensive
literature review and a powerful  state-of-the-art solver, QSDPNAL,
for the problem \cref{eq-qsdp}.

To apply \cref{alg:mfstride}, PG, or APG to \cref{eq:QSDP}, we need to
perform the projection onto the feasible set
\begin{equation}
\mathcal{X} \coloneqq \left\{X\in \SS^n_+\mid \inprod{E}{X} =
0\right\}.
\label{eq:X}
\end{equation}
The following
lemma provides an effective way for performing such a projection.
\begin{lemma}
\label{lemma-PiX}
Define $J\coloneqq I_n - n^{-1}ee^\top \in \SS^n $, then for the set
$\mathcal{X}$ defined in \cref{eq:X}, it holds that
\[
    P_{\mathcal{X}}\left(G\right) = P_{\SS_+^n}(JGJ),\quad \forall G\in \SS^n.
\]
\end{lemma}
\begin{proof}
    First, for any $X\in \mathcal{X}$, clearly $Xe = 0$ and $e^\top X
	= 0$. Therefore, we observe that
        \begin{align*}
            &\; \norm{X - JGJ}_F^2 \\
            = &\; \norm{X - \left(I_n - \frac{1}{n}ee^\top\right)G\left(I_n - \frac{1}{n}ee^\top\right) }_F^2 \\
            = &\; \norm{X - G + \frac{1}{n}Gee^\top + \frac{1}{n} ee^\top G + \frac{1}{n}ee^\top}_F^2 \\
			= &\; \norm{X- G}_F^2 + \frac{2}{n}\inprod{X - G}{Gee^\top
			+ ee^\top G + ee^\top} + \frac1n \norm{Gee^\top + ee^\top G + ee^\top}_F^2 \\
			= &\; \norm{X- G}_F^2 - \frac{2}{n} \inprod{G}{Gee^\top +
			ee^\top G + ee^\top} + \frac1n \norm{Gee^\top + ee^\top G + ee^\top}_F^2.
        \end{align*}
    As a consequence, it holds that $P_{\mathcal{X}}(G) =
	P_{\mathcal{X}}(JGJ)$.
	Moreover, as $JGJe = JG(Je) = JG(0) = 0$, namely, $JGJ$ has an eigenvalue of $0$
	associated with the eigenvector $e$, we get that
	$P_{\SS_+^n}(JGJ) \in \mathcal{X}$ because the projection onto
	$\SS_+^n$ only truncates negative eigenvalues to zero in the
	eigendecomposition.
	Since $\mathcal{X}\subseteq \SS_+^n$, it follows that $P_{\mathcal{X}}(JGJ) = P_{\SS_+^n}(JGJ)$.
\end{proof}
From \cref{lemma-PiX}, we see that the computational bottleneck
lies in the eigendecomposition of matrices in $\SS^n$, which could be
highly expensive or even computationally prohibited in our high
dimensional setting. Thus, we need to rely on low-rank approximate
eigendecomposition to perform inexact projections.

%
%
%

Similar to \cref{eq:bm}, we can also use the BM approach
to  solve \cref{eq-qsdp}. In particular, the factorized problem takes the following form
\begin{equation}
	\min_{W\in \R^{n \times k}}
	g\left( W \right) \coloneqq \frac{1}{2}
	\norm{\mathcal{A}\left(WW^\top\right) - b}^2 +
	\inprod{C}{WW^\top},\quad \mathrm{s.t.}\quad
	W^\top e = 0.
	\label{eq:bmprob}
\end{equation}
The gradient of the function $g:\R^{n\times k}\rightarrow \R$ is then given as
\[
    \nabla g(W) = 2\mathcal{A}^*\left(\mathcal{A}\left(WW^\top\right) - b\right)W + 2CW,
\]
and for any $D\in \mathbb{R}^{n\times k}$, the Hessian operator of $g$ performed on $D$ is given by 
\[
    \nabla^2 g(W)[D] = 2\mathcal{A}^*\left(\mathcal{A}\left(WW^\top \right) -b\right)D + 2\mathcal{A}^*\left(\mathcal{A}\left(WD^\top + DW^\top\right)\right)W + 2CD.
\]
Since $\left\{W\in \mathbb{R}^{n\times k}\mid W^\top e = 0 \right\}$
defines a Riemannian manifold, by using the above information related
to $g(\cdot)$, we can apply many efficient solvers for Riemannian
optimization to solve \cref{eq:bmprob}. In our experiments in
\cref{sec:qsdpexp} for this QSDP problem, we use the state-of-the-art
solver Manopt \citep{manopt} in our BM phase.

\section{Numerical experiments}
\label{sec:exp}
We conduct numerical experiments to exemplify the practical efficiency
of the proposed algorithmic framework. In particular, we consider the
two tasks discussed in \cref{sec:application} with large-scale
real-world data sets in multicore environments. All algorithms are
implemented in MATLAB and C/C++.

\subsection{Matrix completion}
The first task we consider is the matrix completion problem
in the forms of \cref{eq:mc,eq:mf}.
We use one toy example included in the package LIBPMF
(\url{https://www.cs.utexas.edu/~rofuyu/libpmf/}) and four publicly
available large-scale recommendation system data
sets for this set of experiments.\footnote{movielens100k:
	\url{https://www.kaggle.com/prajitdatta/movielens-100k-dataset}.
	(We used the split from ua).
	movielens10m:
	\url{https://www.kaggle.com/smritisingh1997/movielens-10m-dataset}.
	(We used the split from ra).
	Netflix:
	\url{https://www.kaggle.com/netflix-inc/netflix-prize-data}.
Yahoo-musc: the R2 one at
\url{https://webscope.sandbox.yahoo.com/catalog.php?datatype=r}.}
The only preprocessing we did was to tranpose the data matrices when
necessary to conform to our blanket assumption of $m \leq n$.
For all data sets, We use their original training/test split.
These data sets are summarized in \cref{tbl:data}.
The column $|\Omega_{\text{test}}|$ indicates the number of entries in
the test set.
For the value of $\lambda$ on the real-world data, we follow the
values provided by \cite{HsiO14a} that were obtained through
cross-validation, while the final $k$ is the rank of the final output
of our algorithm, obtained by running our algorithm with the given
$\lambda$ till the objective cannot be further improved.
The value of $\lambda$ for the toy example is from some simple tuning
to make the final rank not too far away from that of other data sets.

\begin{table}[tbh]
\centering
\begin{tabular}{lrrrrrr}
\toprule
Data set & $m$ & $n$ & $|\Omega|$ & $|\Omega_{\text{test}}|$ &
$\lambda$ & final $k$\\
\midrule
toy-example & 3952 & 6040
& 900189 & 100020 & 36 & 62\\
movielens100k & 943 & 1682 & 90570 & 9430 & 15 & 68 \\
movielens10m & 65133 & 71567 & 9301274 & 698780 & 100 & 50 \\
netflix & 17770 & 2649429 & 99072112 & 1408395 & 300 & 68\\
yahoo-music & 624961 & 1000990 & 252800275 & 4003960 & 10000 & 52 \\
\bottomrule
\end{tabular}
\caption{Data statistics for matrix completion.}
\label{tbl:data}
\end{table}

Experiments on the first four data sets are conducted on an Amazon AWS
EC2 c6i.4xlarge instance with an 8-core Intel Xeon Ice Lake processor
and 32GB memory.
For the larger yahoo-music data set, an m6i.4xlarge instance that has
the same processor but with 64GB memory is used.
Our experiments in this subsection utilize all cores available for all
algorithms through parallelization by MATLAB and openMP.

For this task, we conduct four sets of experiments.
First, we use the first two smaller data sets to see how different
numbers of consecutive inexact proximal gradient iterations and consecutive
epochs in the BM phase affect the behavior of our algorithm.
Next, we empirically examine the result of \cref{thm:identify} by
checking how fast \ours identifies the active manifold, namely the correct rank.
We then compare our whole method with its BM solver subroutine alone to see that our
method is as efficient as the BM solver and can escape from stationary
points of \cref{eq:mf} that are not global optima.
In the last set of experments, we compare our method with the state of
the art for \cref{eq:mc}.
We note that for this problem, $\nabla f$ is $1$-Lipschitz continuous,
and thus a fixed step size of $\alpha = 1.99$ can be used to satisfy
\cref{eq:back} without any data-dependent computation.
We have also tested a version that follows SpaRSA \citep{WriNF09a} to
use the spectral step size initialization strategy of
\cite{barzilai1988two} together with backtracking linesearch, but it
did not result in better performance, and therefore we will use this
fixed-step variant throughout.
For completeness, we include the experiments with the SpaRSA variant
in the supplementary materials.

To compare different methods, we consider two criteria, one from the
optimization point of view and the other from the task-oriented angle.
In particular, we first run our algorithm till the objective cannot be
further improved, and take the obtained output $X^*$ as the numerical
global optimal solution. With the knowledge of this $X^*$, our first criterion is the relative objective
\begin{equation}
	\frac{F(X) - F(X^*)}{F(X^*)}.
	\label{eq:obj}
\end{equation}
The second measure we use is the relative root mean squared error (RMSE), which is computed as
\begin{equation}
	\frac{\text{RMSE}(X) - \text{RMSE}(X^*)}{\text{RMSE}(0) -
	\text{RMSE}(X^*)},\quad
	\text{RMSE}(X) \coloneqq
	\sqrt{\frac{\norm{P_{\Omega_{\text{test}}}\left( X - A
\right)}^2_F}{|\Omega_{\text{test}}|}}.
	\label{eq:rmse}
\end{equation}

Although in general the norm of the exact proximal gradient step would
also be a better optimization progress measure especially because it does not require the
knowledge of $F(X^*)$, its computation is impractical in this set of experiments because $mn$ is usually too huge for us to form $X_t$ explicitly and compute its exact SVD that is needed for calculating the exact proximal gradient step.

\subsubsection{Parameter tuning for our method}
\label{subsec:mc-parameter}
We first use the toy example and movielens100k to finalize details in
the parameters setting of our algorithm.
In particular, we test the setting of alternating between
$x$ consecutive inexact proximal gradient steps and $y$ consecutive
iterations of the BM phase solver, with $x \in \{1,\dotsc,5\}$ and $y
\in \{1,\dotsc,8\}$, for our fixed step variant with $\alpha \equiv
1.99$.
More Details and the results are shown in the supplementary materials.
Our result indicates that there is no definite best performer in all
cases.
But in general, $x=1$ and $y=3$ seems to be a rather robust choice.
This observation accords with our argument that eigendecompositions
are rather expensive and the BM steps should be utilized more often than the
proximal gradient steps.
We therefore will stick to this setting in all the remaining
experiments in this subsection.

\subsubsection{Stabilization of the rank}
\begin{figure}[tbh]
	\centering
	\begin{tabular}{@{}c@{\hspace{5pt}}c@{\hspace{5pt}}c@{\hspace{5pt}}c@{}}
		\begin{subfigure}[b]{.24\textwidth}
			\includegraphics[width=\textwidth]{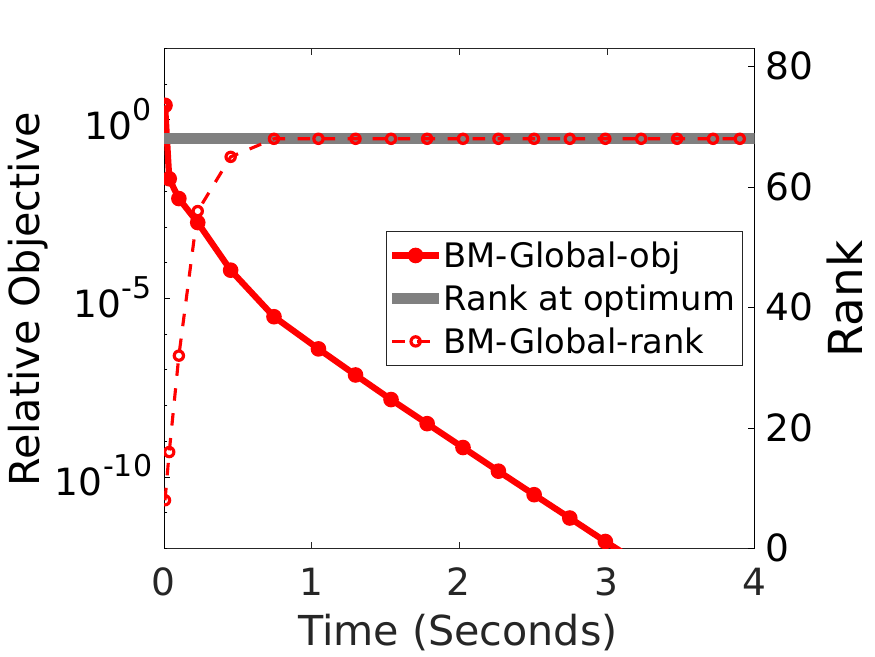}
			\caption{movielens100k}
		\end{subfigure} &
		\begin{subfigure}[b]{.24\textwidth}
			\includegraphics[width=\textwidth]{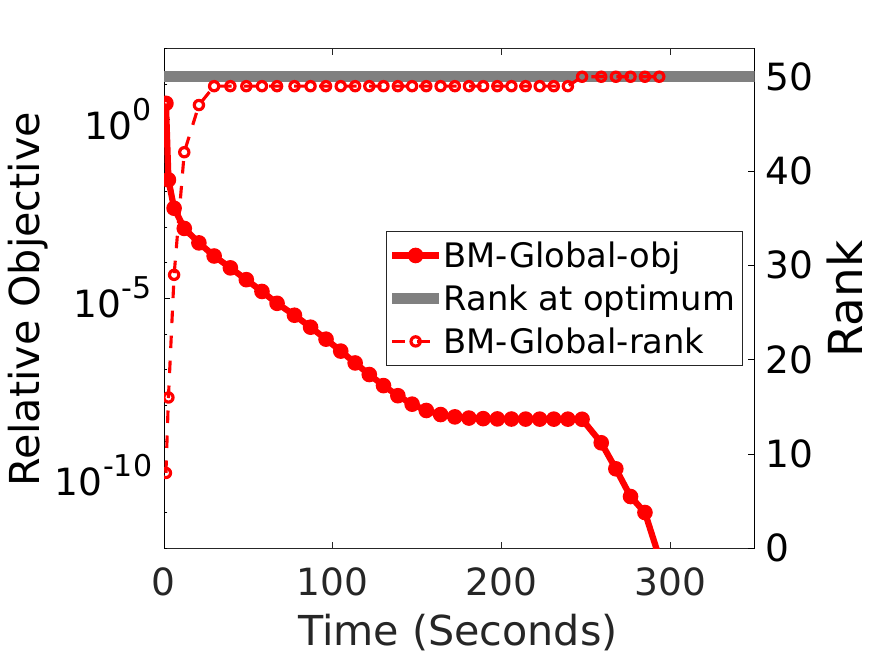}
			\caption{movielens10m}
		\end{subfigure} &
		\begin{subfigure}[b]{.24\textwidth}
			\includegraphics[width=\textwidth]{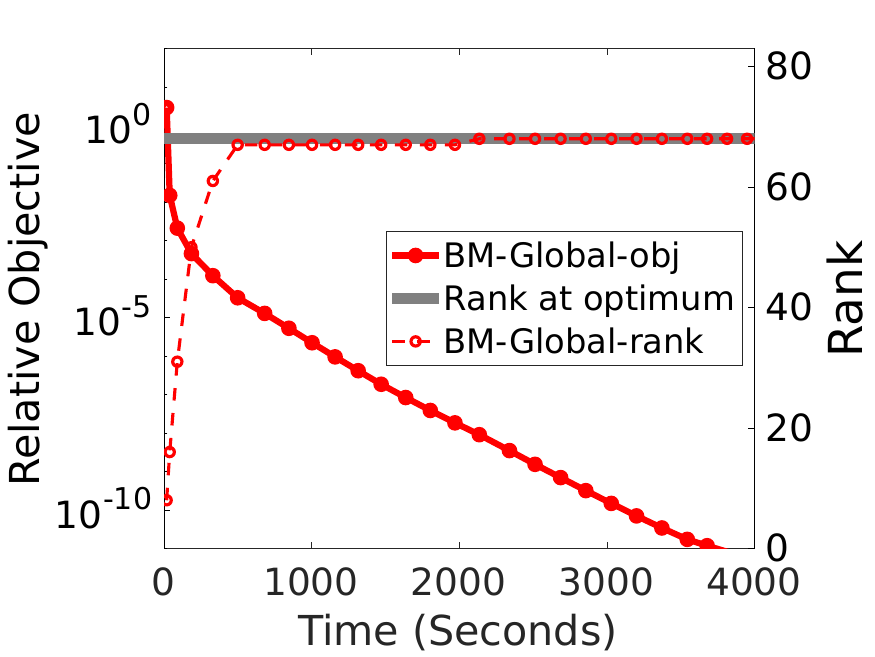}
			\caption{netflix}
		\end{subfigure} &
		\begin{subfigure}[b]{.24\textwidth}
			\includegraphics[width=\textwidth]{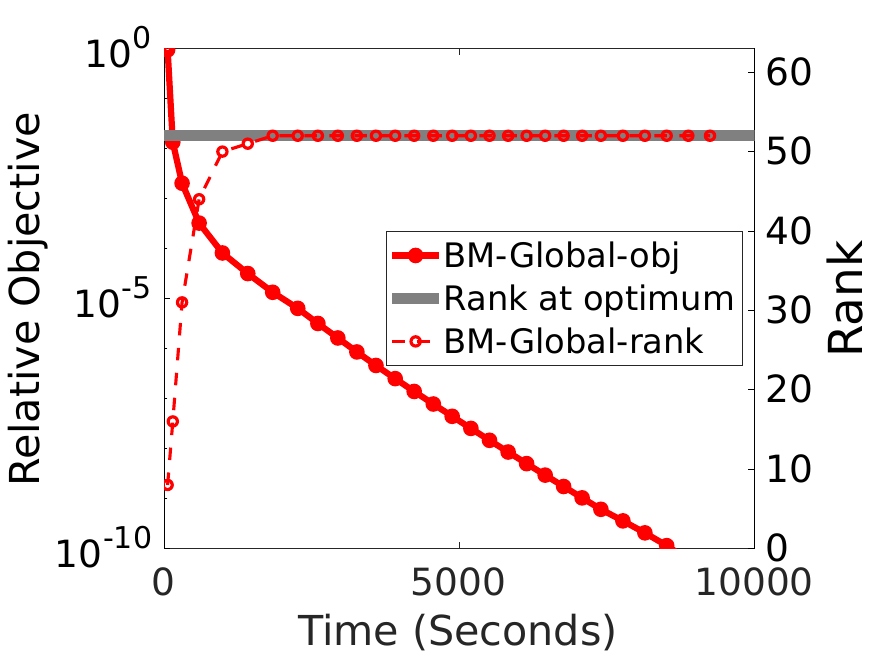}
			\caption{yahoo-music}
		\end{subfigure}
\end{tabular}
	\caption{Rank and relative objective of the iterates of
		\ours over running time.}
	\label{fig:rank}
\end{figure}

We then show the rank of $X_t$ over iterations of \ours
In \cref{fig:rank}, we use solid lines and dash lines to respectively
show the relative objective value and the rank of the iterates of our
method.
The gray line represents the rank at the optimum $X^*$.
We can see that the rank of $X_t$ increases quickly at first, and
eventually stabilizes at the rank of the point of convergence in all cases.
Sometimes, the rank remains fixed for a while, then is
increased by a small number, and finally stays at the new rank.
This is the situation that a safeguard (see the supplementary
materials) kicks in to resolve the insufficient rank problem and
ensures that the iterates indeed converge to a global optimum.
We can also see that when the rank reaches $\rank(X^*)$,
the relative objective also drops significantly,
indicating that finding the right rank is essential in solving
\cref{eq:matrixform} to a high precision.

\subsubsection{Comparison between \ours and the BM
solver alone}
We next compare \ours with running a BM solver only for \cref{eq:mf}.
We directly use the solver in our BM phase, namely, the polyMF-SS
method of \citet{WanLK17a}, with their original random start scheme to
avoid starting from the origin, which is a known saddle point of
\cref{eq:mf}.
Given any value $k$, their method starts from a randomized $W \in \R^{m
\times k}$ and $H \in \R^{n \times k}$.
We favor their method to directly assign $k$ as the final rank shown
in \cref{tbl:data}, but we emphasize that in real-world applications,
finding this $k$ will require additional effort in parameter search.

\begin{figure}[tbh]
	\centering
	\begin{tabular}{@{}c@{\hspace{5pt}}c@{\hspace{5pt}}c@{\hspace{5pt}}c@{}}
		\multicolumn{4}{c}{Iterations}\\
		\begin{subfigure}[b]{.24\textwidth}
			\includegraphics[width=\textwidth]{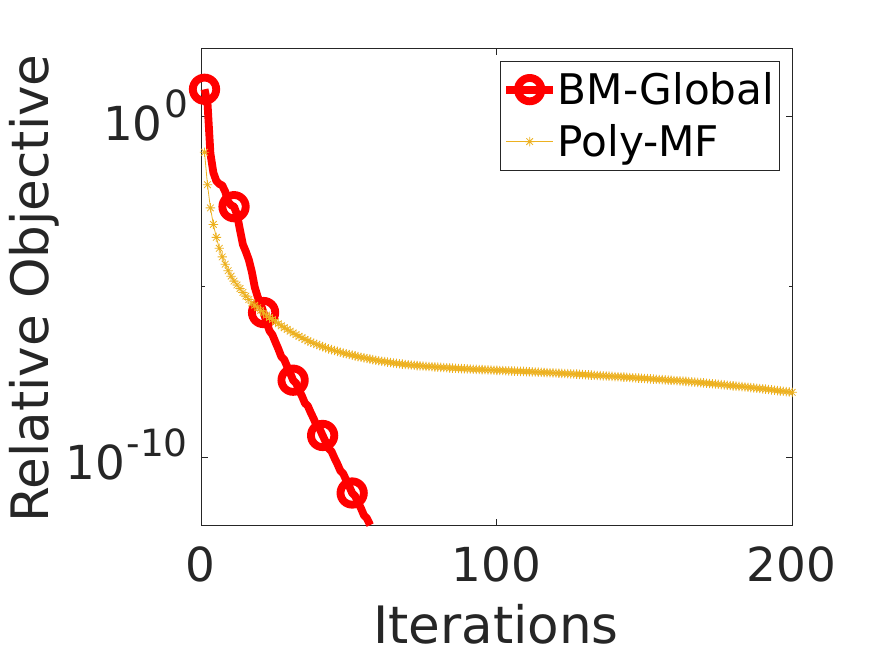}
			\caption{movielens100k}
		\end{subfigure} &
		\begin{subfigure}[b]{.24\textwidth}
			\includegraphics[width=\textwidth]{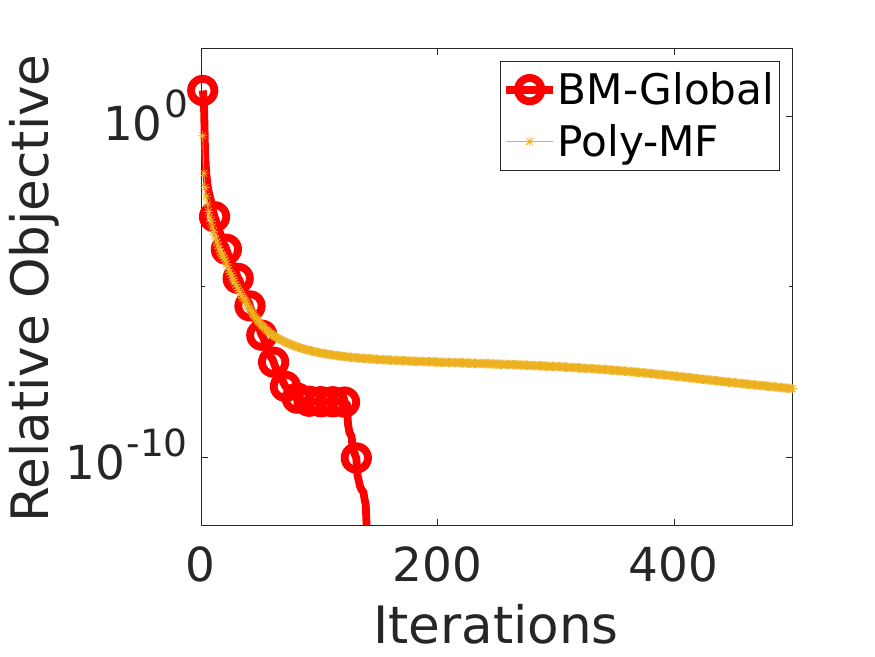}
			\caption{movielens10m}
		\end{subfigure} &
		\begin{subfigure}[b]{.24\textwidth}
			\includegraphics[width=\textwidth]{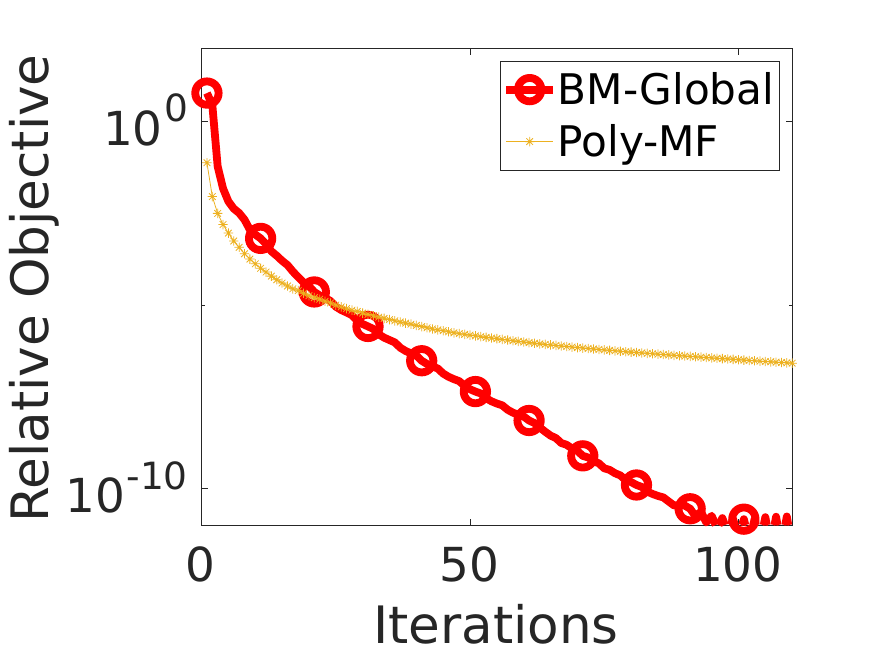}
			\caption{netflix}
		\end{subfigure} &
		\begin{subfigure}[b]{.24\textwidth}
			\includegraphics[width=\textwidth]{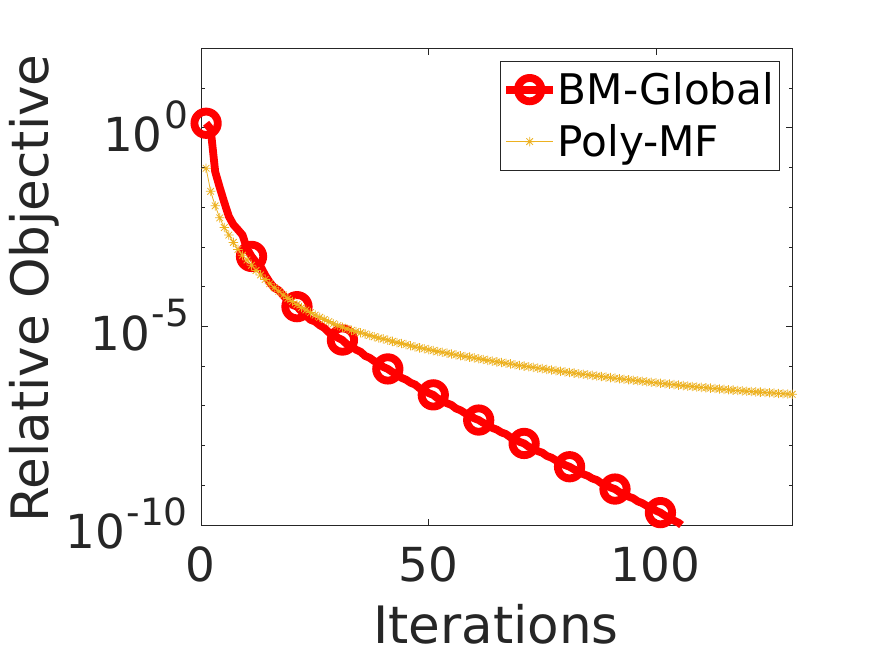}
			\caption{yahoo-music}
		\end{subfigure}\\
		\multicolumn{4}{c}{Running time}\\
		\begin{subfigure}[b]{.24\textwidth}
			\includegraphics[width=\textwidth]{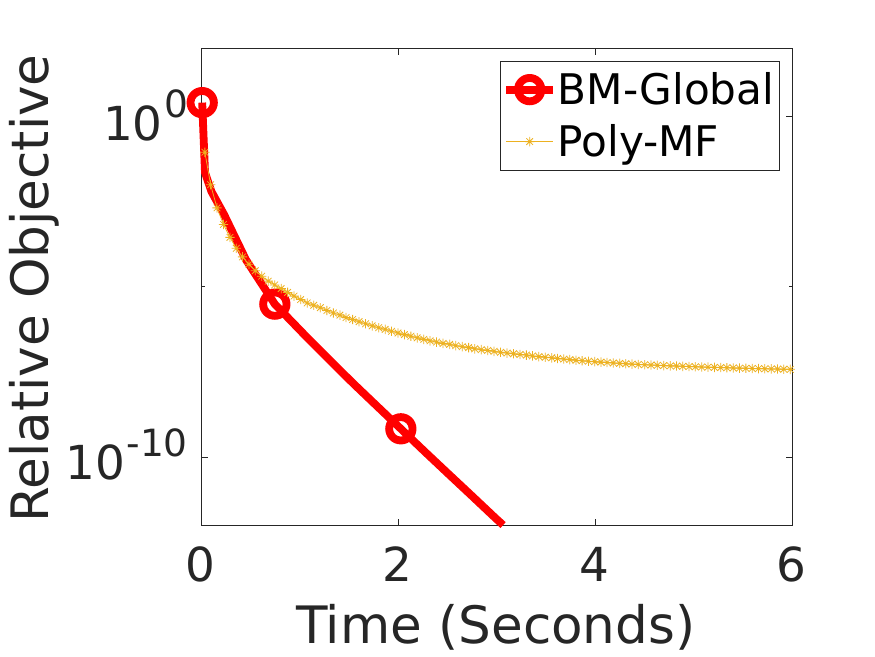}
			\caption{movielens100k}
		\end{subfigure} &
		\begin{subfigure}[b]{.24\textwidth}
			\includegraphics[width=\textwidth]{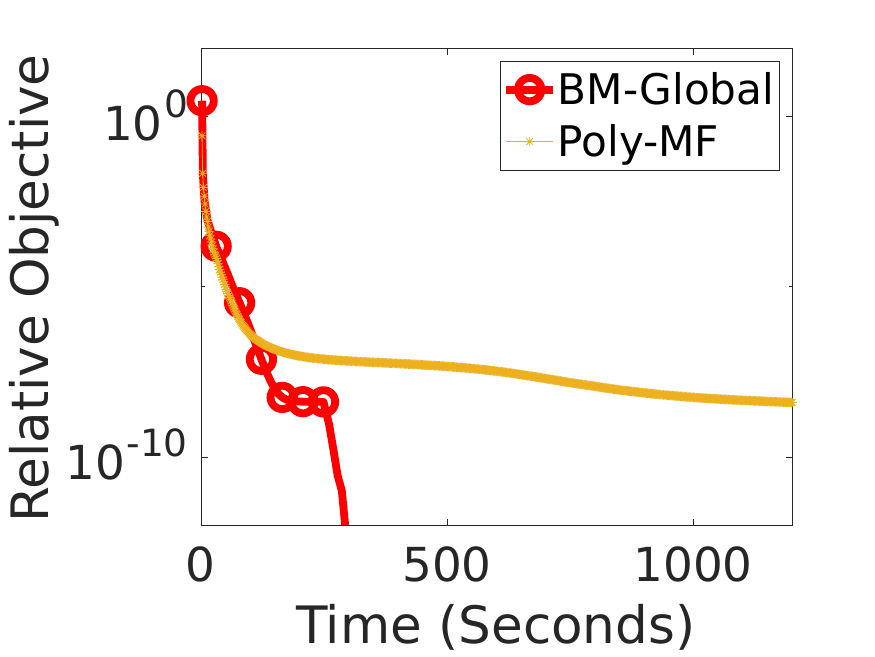}
			\caption{movielens10m}
		\end{subfigure} &
		\begin{subfigure}[b]{.24\textwidth}
			\includegraphics[width=\textwidth]{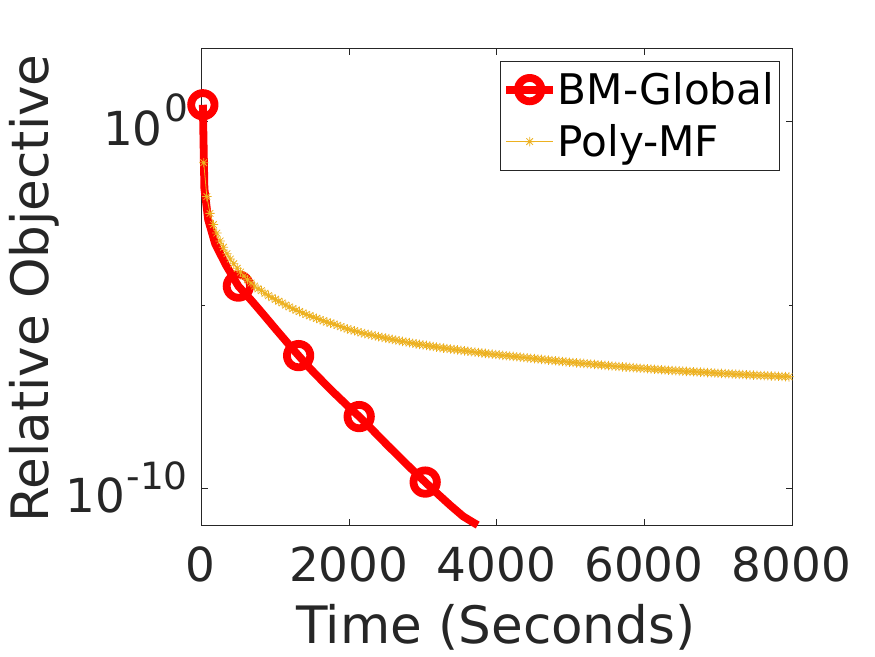}
			\caption{netflix}
		\end{subfigure} &
		\begin{subfigure}[b]{.24\textwidth}
			\includegraphics[width=\textwidth]{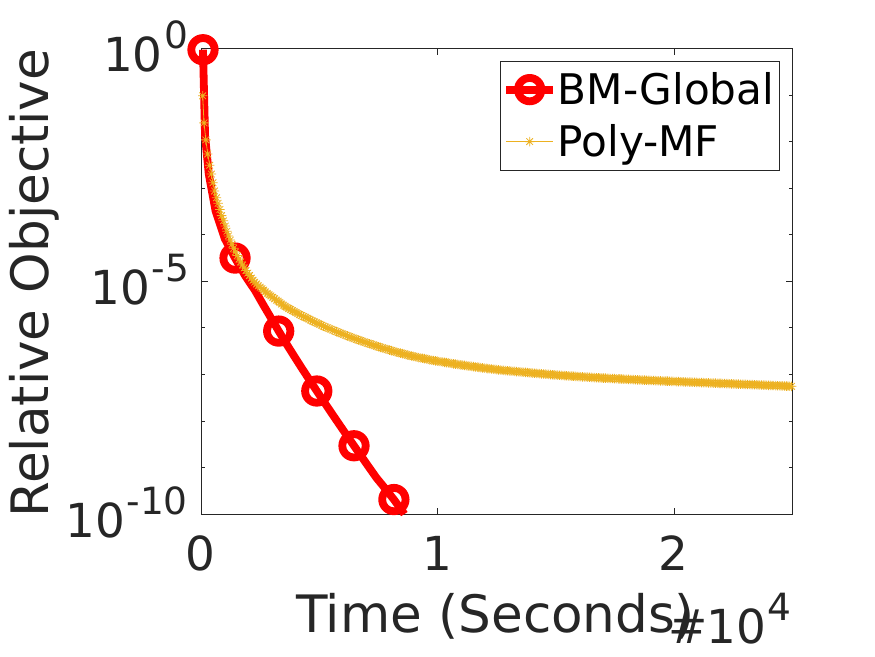}
			\caption{yahoo-music}
		\end{subfigure}
\end{tabular}
	\caption{Comparison between PolyMF-SS and \ours.
		Top row: iterations v.s. relative objective.
		Bottom row:
		running time (seconds) v.s. relative objective.}
	\label{fig:mf}
\end{figure}

The purpose of this experiment is to show that solving \cref{eq:mf}
only can get the iterates stuck at saddle points or spurious local
minima, while \ours can effectively and efficiently
escape from such points.
Therefore, we consider the relative objective as the only comparison
criterion in this experiment.
The results of running time and number of iterations are shown in
\cref{fig:mf}.
For the number of iterations, we count either one inexact proximal
gradient step or one epoch of polyMF-SS (one sweep through the whole
data) as one iteration.

We observe that in terms of iterations, PolyMF-SS has a small early
advantage due to the larger starting rank in \cref{eq:bm}.
But its convergence quickly slows down, suggesting that likely
the iterates are attracted to a saddle point or a spurious local
minimum that is strictly worse than the global optima.
On the other hand, the story in the running time comparison is very
different.
We see that the higher rank in PolyMF-SS from the beginning on
actually increases the time cost per epoch, and thus the early
advantage of PolyMF-SS over \ours we observed in terms of iterations
is not present in the time comparison.
Another observation is that in the numerical experiments, the
empirical convergence speed of \ours is indeed $Q$-linear as predicted
by \cref{linearT}.

Overall speaking, \ours is as efficient as running a
solver for \cref{eq:mf} alone, but it provides multiple advantages
including the guarantee of convergence to the global optima.
Although in this experiment, the stationary points to which the iterates
of PolyMF-SS converge seem to be of good enough quality, we have no
guarantee that on other data sets, or even on these data sets but with a
different $\lambda$, their points of convergence will still be of
satisfactory quality.

\subsubsection{Comparison with existing methods}
Now that it is clear our method is advantageous over running a
solver for \cref{eq:mf} alone,
we proceed to compare \ours with the state of the art for
\cref{eq:mc}.
In particular, we compare \ours with the following:
\begin{itemize}
	\item Active-ALT \citep{HsiO14a}:
		This method alternates between conducting an inexact PG step
		and solving a lower-dimensional convex subproblem.
		In the approximate SVD part for inexact PG, \citet{HsiO14a}
		use the power method with warmstart from the output of
		the previous iteration plus some random columns as a safeguard.
	\item AIS-Impute \citep{YaoKWL18a}:
		An inexact APG method that also uses the power method for
		approximate SVDs.
		They use the combination of the outputs of the previous
		iteration and the iteration preceding it to form the warmstart
		matrix.
\end{itemize}
The inexact APG method in \cite{TohY10a} is
not included because the underlying APG part is the same as that of
AIS-Impute, but their approximate SVD using Lanczos is shown by
\cite{YaoKWL18a} to be less efficient.

\begin{figure}[tbh]
	\centering
	\begin{tabular}{@{}c@{\hspace{5pt}}c@{\hspace{5pt}}c@{\hspace{5pt}}c@{}}
		\multicolumn{4}{c}{Relative objective}\\
		\begin{subfigure}[b]{.24\textwidth}
			\includegraphics[width=\textwidth]{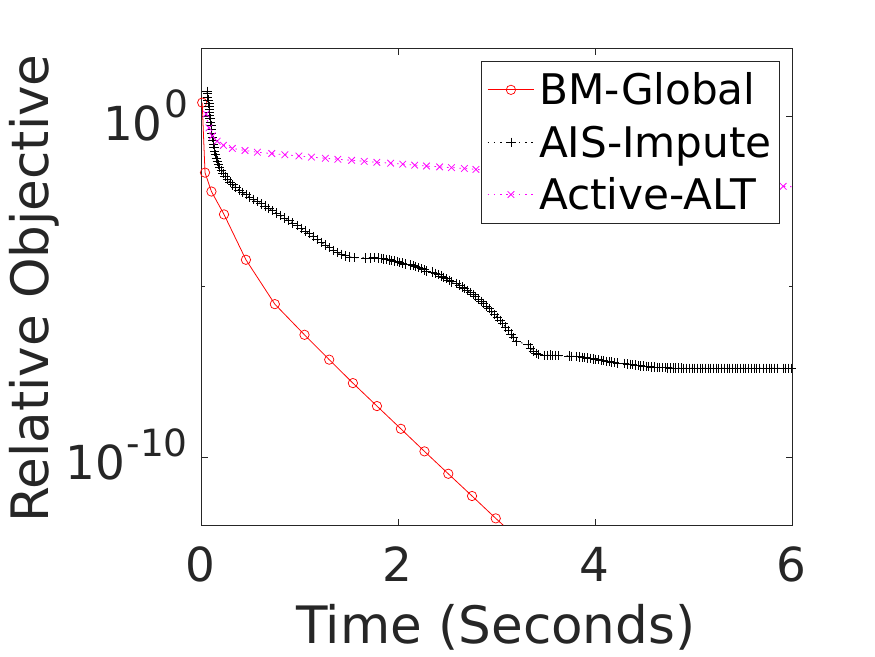}
			\caption{movielens100k}
		\end{subfigure} &
		\begin{subfigure}[b]{.24\textwidth}
			\includegraphics[width=\textwidth]{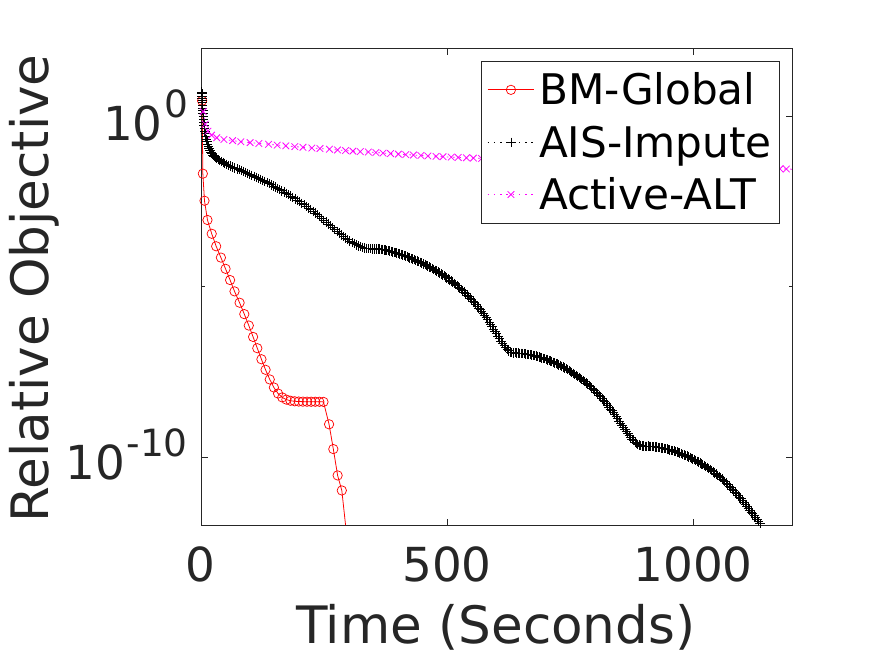}
			\caption{movielens10m}
		\end{subfigure} &
		\begin{subfigure}[b]{.24\textwidth}
			\includegraphics[width=\textwidth]{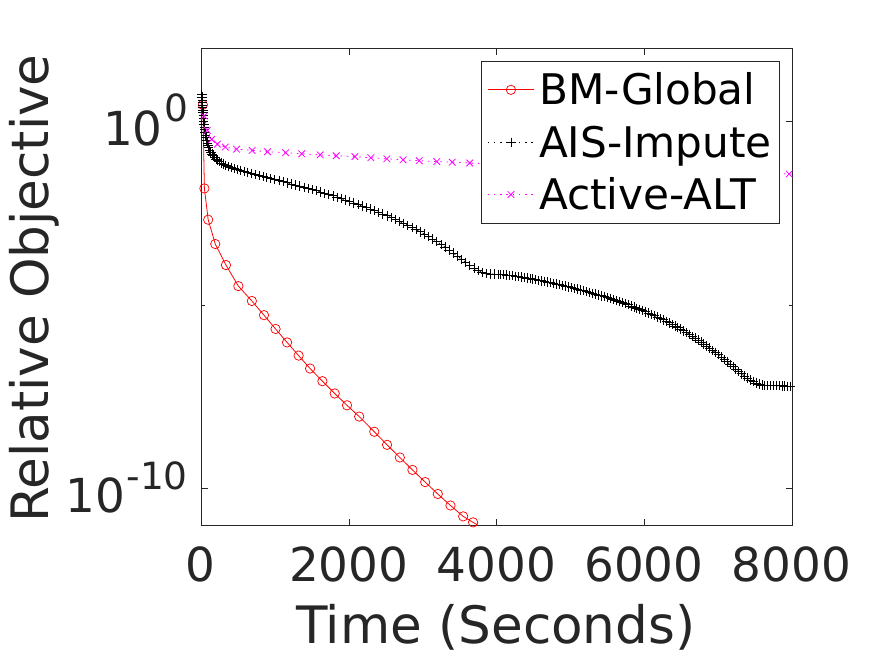}
			\caption{netflix}
		\end{subfigure} &
		\begin{subfigure}[b]{.24\textwidth}
			\includegraphics[width=\textwidth]{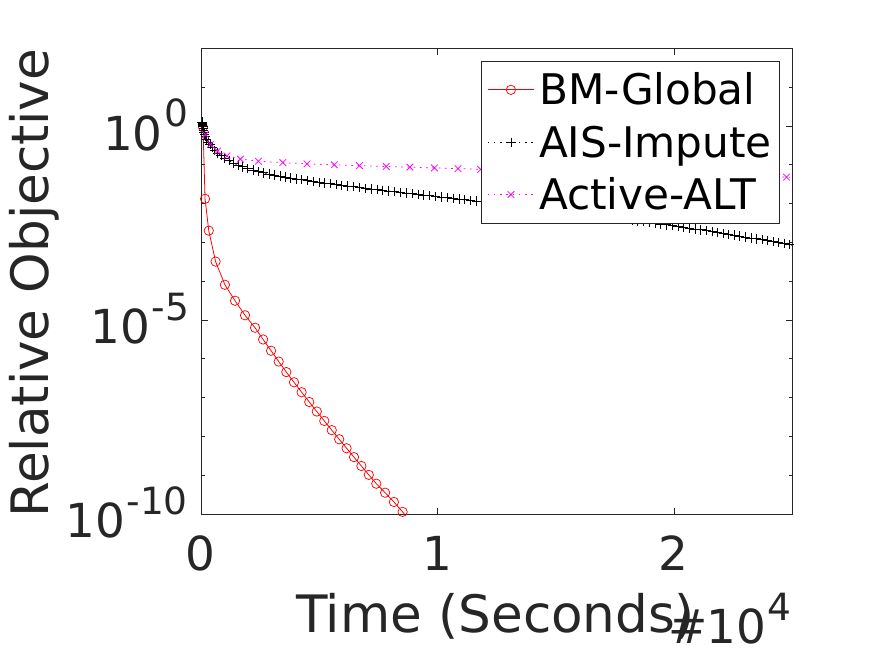}
			\caption{yahoo-music}
		\end{subfigure}\\
		\multicolumn{4}{c}{Relative RMSE}\\
		\begin{subfigure}[b]{.24\textwidth}
			\includegraphics[width=\textwidth]{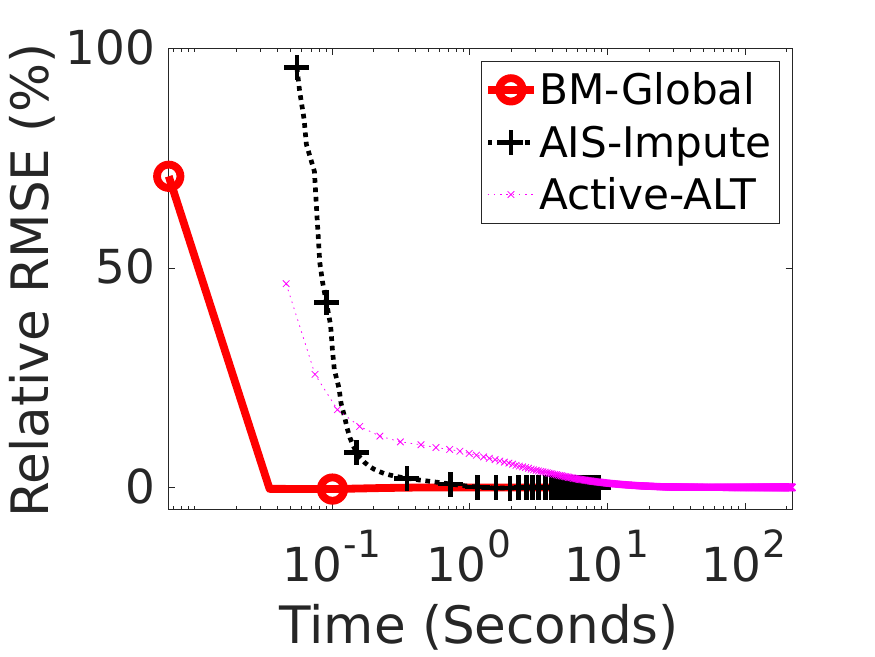}
			\caption{movielens100k}
		\end{subfigure} &
		\begin{subfigure}[b]{.24\textwidth}
			\includegraphics[width=\textwidth]{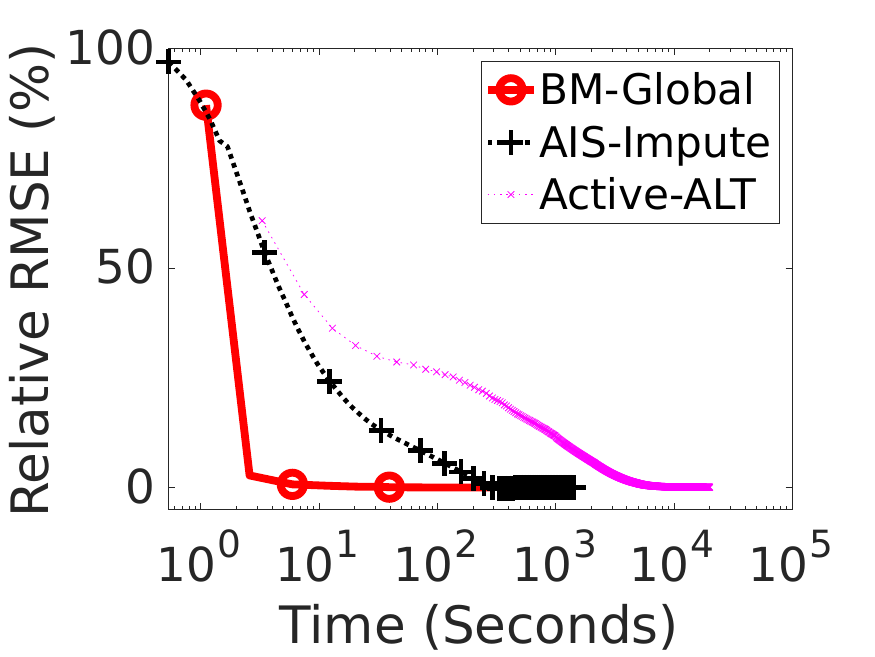}
			\caption{movielens10m}
		\end{subfigure} &
		\begin{subfigure}[b]{.24\textwidth}
			\includegraphics[width=\textwidth]{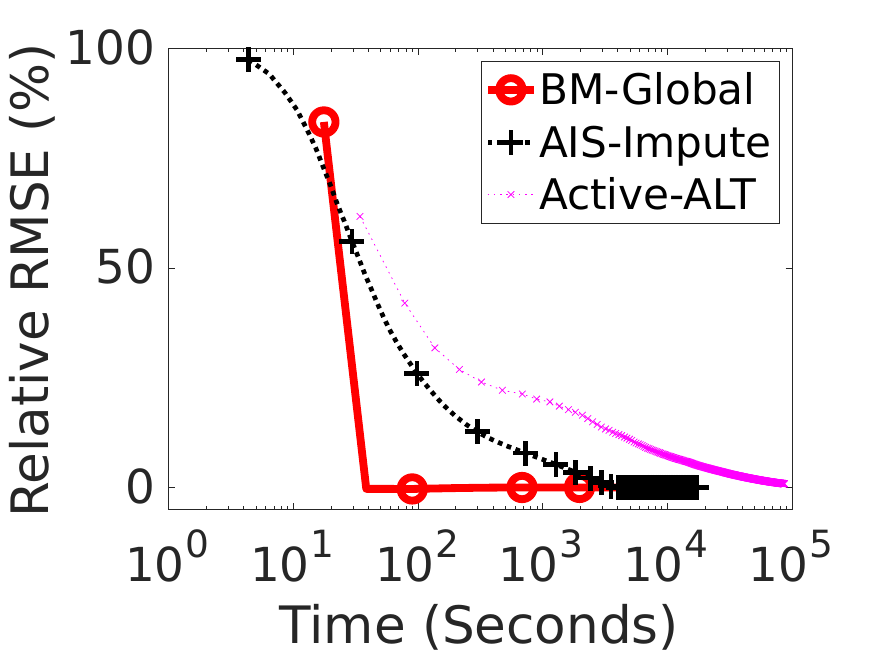}
			\caption{netflix}
		\end{subfigure} &
		\begin{subfigure}[b]{.24\textwidth}
			\includegraphics[width=\textwidth]{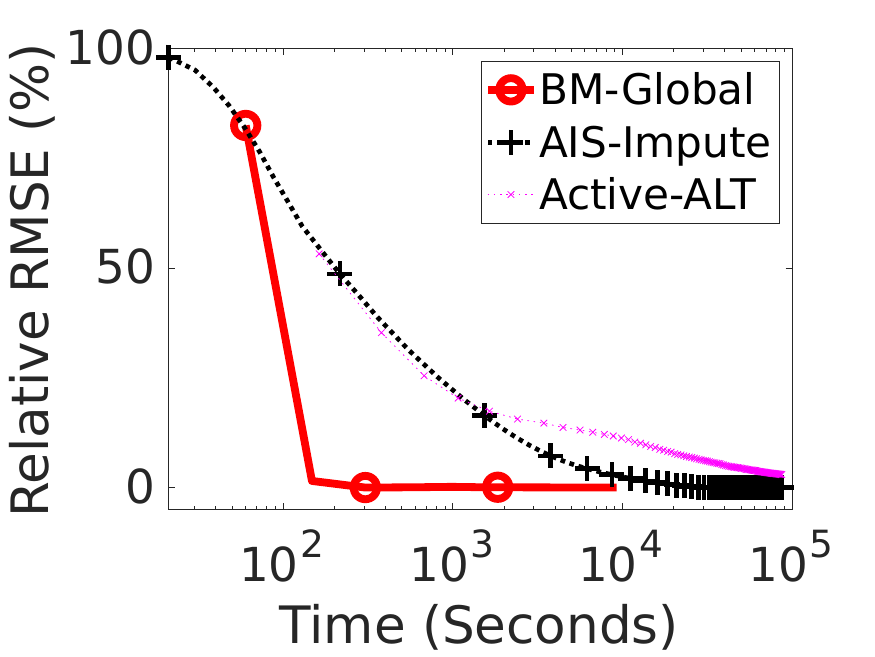}
			\caption{yahoo-music}
		\end{subfigure}
\end{tabular}
	\caption{Comparison between \ours and existing methods.
		Top row: running time v.s. relative objective.
	Bottom row: running time (log scale) v.s. relative RMSE.}
	\label{fig:mc}
\end{figure}

The results of relative objective and relative RMSE are shown in
\cref{fig:mc}.
Note that the running time for relative RMSE in \cref{fig:mc} is in
log scale to make the difference legible.
Clearly, \ours outperforms the state of the art for \cref{eq:mc}
significantly on both criteria.
Particularly, \cref{fig:mc} exemplifies even greater efficiency
difference in reaching satisfactory RMSE between \ours
and existing methods.
We can see that the proposed approach is actually magnitudes
faster than state of the art in this criterion.

\subsection{Convex QSDP} \label{sec:qsdpexp}
Next, we consider solving the two applications of QSDP described in \cref{sec:app-qsdp}. As mentioned before, PG and APG can be applied to solve the problem directly. However, based on our empirical experience, both methods require too many iterations and excessive runtime to reach a reasonably good solution, so their numerical results are excluded here. (Interested readers may refer to the supplementary material for the numerical results of the APG methods.) We hence only compare \ours with the efficient and robust QSDP solver, QSDPNAL \citep{li2018qsdpnal}.\footnote{Avaliable at \url{https://blog.nus.edu.sg/mattohkc/softwares/qsdpnal/}.}

Regarding the termination conditions, since QSDPNAL computes
both primal and dual iterates, its relative KKT residual is
computable (see \cite[Section 5.2]{li2018qsdpnal} for the
definition).
Thus, given a specific stopping tolerance \texttt{tol}, QSDPNAL is
terminated when the maximum relative KKT residual, denoted by $\eta_{\rm kkt}$, is less than \texttt{tol}.
Moreover, when $n$ is large, QSDPNAL may take too much computational
time (since it uses full eigendecompositions), so we also cap the
running time of QSDPNAL to four hours (initialization overhead
excluded) and its maximum number of iterations to $200$. For \ours,  we
terminate it when
\[
    \frac{\lvert f(W_tW_t^\top) - f(W_{t-1}W_{t-1}^\top)\rvert}{(1 + \lvert f(W_{t-1}W_{t-1}^\top)\rvert)} < \texttt{tol}.
\]
In our experiments, we set $\texttt{tol} = 10^{-6}$ for both methods.

Recall that the first-order optimality condition for problem \cref{eq-qsdp} is given by 
\[
X - P_{\mathcal{X}}(X - \nabla f(X)) = 0,\quad X\in \SS^n.
\]
Since we are testing problems with $n$ that can be handled by QSDPNAL
that uses full eigendecompositions, we are in fact able to check
whether an approximate solution $X\in \SS^n$ is optimal numerically,
even though this can be time-consuming. Therefore, to compare the
quality of the solutions returned by \ours and QSDPNAL,
we record the relative primal feasibility and  the relative
optimality, respectively defined as
\[
    \eta_{\rm prim}(X) \coloneqq \frac{\lvert \inprod{E}{X}\rvert}{1 +
		\norm{X}_F},\quad\text{ and }\quad
    \eta_{\rm opt}(X) \coloneqq \frac{\norm{X - P_{\mathcal{X}}(X - \nabla f(X))}_F}{1 + \norm{X}_F + \norm{\nabla f(X)}_F}.
\]

Experiments for this part are conducted on a Linux PC with an Intel Xeon E5-2650 processor and 96GB memory.

\subsubsection{Regularized kernel estimation} \label{sec:rke}
We consider problems with dissimilarity measures $d_{ij}$ for $1 \leq
i, j \leq n$ collected in \cite{duin2009datasets}.\footnote{Data
	available at
	\url{http://prtools.tudelft.nl/Guide/37Pages/distools.html}.} In
	our experiments, the data $d_{ij}$ are scaled to
	the interval $[0,1]$, and the elements of the index set $\Omega$
	are randomly selected such that $\lvert\Omega\rvert\approx n/20$.
	We set $w_{ij} = 1$ for all $(i,j)\in \Omega$ and $\lambda = \sqrt{n}/10$.

The results are presented in \cref{tab:my-rke}. Clearly, both methods
are able to compute nearly feasible and low-rank solutions.
In terms of the optimality
measure, \ours is able to solve all the problems with $\eta_{\rm opt}
< 10^{-3}$ while QSDPNAL fails to do so for one of the problems. In
terms of efficiency, we see that  \ours is faster in most cases, and
\ours can even be ten times faster than QSDPNAL in the case of the
largest instance.

	\begin{table}[tb]
		\centering
\begin{footnotesize}
\begin{tabular}{|l|r|r|r|r|r|r|r|r|r|r|}
\hline
&    & \multicolumn{5}{c|}{\texttt{QSDPNAL}} &
\multicolumn{4}{c|}{\ours} \\ \hline

%
\texttt{Name} & $n$ & $\eta_{\rm prim}$&$ \eta_{\rm opt} $&$ \eta_{\rm kkt} $&$ \texttt{rnk} $&$ \texttt{Time}$ & $\eta_{\rm prim}$&$ \eta_{\rm opt} $&$ \texttt{rnk} $&$ \texttt{Time}$ \\ \hline
 BrainMRI &  124 & 6e-11 & 1e-06 & 1e-06 &    5 &   0.7 & 2e-16 & 5e-07 &    5 &  0.5 \\
 protein &  213 & 4e-13 & 4e-06 & 5e-07 &   24 &   4.3 & 5e-11 & 7e-06 &   24 &  1.9 \\
 CoilDelftDiff &  288 & 9e-13 & 4e-04 & 9e-07 &   32 &   5.7 & 4e-15 & 5e-08 &   32 &  1.6 \\
 coildelftsame &  288 & 9e-14 & 4e-06 & 5e-07 &   32 &   6.7 & 2e-11 & 4e-06 &   32 &  2.3 \\
 CoilYork &  288 & 3e-13 & 2e-06 & 4e-07 &   22 &   4.8 & 2e-15 & 7e-06 &   22 &  2.6 \\
  Chickenpieces-5-45 &  446 & 4e-13 & 3e-06 & 7e-07 &   27 &  11.5 & 2e-11 & 8e-06 &   28 &  7.6 \\
 newgroups &  600 & 8e-13 & 3e-05 & 5e-07 &   81 &  39.6 & 2e-12 & 2e-04 &   83 & 29.6 \\
 flowcytodis &  612 & 2e-13 & 4e-06 & 9e-07 &   15 &  14.8 & 1e-12 & 1e-06 &   15 & 13.9 \\
 DelftPedestrians &  689 & 3e-12 & 1e-05 & 2e-07 &   69 &  43.7 & 3e-12 & 7e-06 &   69 & 32.8 \\
 WoodyPlants50 &  791 & 5e-13 & 1e-05 & 7e-07 &   48 &  44.3 & 9e-13 & 2e-07 &   48 & 78.2 \\
 delftgestures & 1500 & 1e-14 & 7e-06 & 5e-07 &   76 & 318.3 & 3e-13 & 2e-06 &   77 & 390.1 \\
 zongker & 2000 & 2e-11 & 1e-02 & 8e-07 &  264 & 1000.2 & 6e-12 & 1e-04 &  267 & 665.6 \\
 polydish57 & 4000 & 2e-12 & 1e-05 & 5e-07 &  100 & 3508.9 & 7e-15 & 3e-05 &  101 & 1765.2 \\
 polydism57 & 4000 & 1e-12 & 5e-04 & 8e-07 &   25 & 3286.4 & 1e-16 & 2e-07 &   26 & 305.0 \\ \hline 
 \end{tabular}
\caption{Computational results on regularized kernel estimation problems.}
\label{tab:my-rke}
\end{footnotesize}
\end{table}

\subsubsection{Molecular conformation} \label{sec:mcp}
In this experiment, we consider the molecules from the Protein Data
Bank (see \url{https://www.rcsb.org/}) with 
given noisy and sparse distance data 
to simulate distances measurable by nuclear
magnetic resonance (NMR) experiments.
For each molecule, if the distance between two 
compatible atoms is less than 
6\AA
($6\times 10^{-8}$ cm), then the distance can be measured by the NMR experiment; otherwise, we assume that no information is known for this pair. 
To simulate the sparse set of distances measurable by the NMR experiment, 
among all the pairwise distances less than 6\AA,
we select 25\% of them to generate our index set $\Omega$.
We then add in additional noise to the observed data as follows.
Let $\tau $ be a given noise level and $\hat{d}_{ij}$ be the exact
distance between atom $i$ and atom $j$ for $(i,j) \in \Omega$, we
sample two independent random variables $\underline{\epsilon}_{ij},
\overline{\epsilon}_{ij}$ from the normal distribution $\mathcal{N}(0,
\pi\tau^2/2)$ and define
\[
\underline{d}_{ij}\coloneqq \max\{1, (1-
	\lvert\underline{\epsilon}_{ij}\rvert)\hat{d}_{ij}\},\quad
	\overline{d}_{ij} \coloneqq (1 + \lvert\overline{\epsilon}_{i,j}\rvert)\hat{d}_{ij}.
\]
Then, the input distances are set as $d_{ij}\coloneqq
(\underline{d}_{ij} + \overline{d}_{ij})/2$. Given $d_{ij}$, we set
$w_{ij} = 1/d_{ij}^2$. Moreover, we let $\lambda = -10\sqrt{n}/\sum_{(i,j)\in\Omega}d_{ij}^2$. In our tests, we set $\tau = 0.1$.
To measure the accuracy of the estimated positions, we record the root mean square deviation (\texttt{RMSD}):
\[
    \texttt{RMSD}\coloneqq \frac{1}{\sqrt{n}}\left(\sum_{i = 1}^n \norm{x_i - \hat{x}_i}^2\right)^2
\]
where $x_i$ is the estimated position and $\hat{x}_i$ is the actual position. Note that a smaller \texttt{RMSD} means a better estimation, 
and an RMSD of less than 2\AA\, is considered to be good in molecular
conformation.

The computational results are presented in \cref{tab:my-mcp}.
It is clear that both methods return nearly feasible solutions.
However, we can see that \ours outperforms QSDPNAL in all other
measures.
In particular, QSDPNAL often returns solutions with a large suboptimality
measure, and those solutions tend to be of a higher rank and give
a larger \texttt{RMSD}. On the other hand, the solutions returned by
\ours are always of low rank with very small \texttt{RMSD}. Moreover,
by the presented computational time, we see that \ours is much more
efficient than QSDPNAL, and its generated solutions are also often much more
accurate.

The results in this and the previous subsections also suggest that the
numerical performance of QSDPNAL may depend on the sign of $\lambda$
while \ours is robust with respect to parameter selection of $\lambda$.

\begin{table}[tb]
		\centering
\begin{footnotesize}
\begin{tabular}{|l|r|r|r|r|r|r|r|r|r|r|r|r|}
\hline 
&    & \multicolumn{6}{c|}{\texttt{QSDPNAL}} &
\multicolumn{5}{c|}{\ours} \\ \hline  
%
%
\texttt{Name} & $n$ & $\eta_{\rm prim}$&$ \eta_{\rm opt} $&$ \eta_{\rm kkt} $&$
\texttt{rnk} $&$ \texttt{RMSD} $&$ \texttt{Time}$ & $\eta_{\rm
prim}$&$ \eta_{\rm opt} $&$ \texttt{rnk} $&$ \texttt{RMSD} $&$
\texttt{Time}$ \\ \hline 
1PBM &  126 & 1e-12 & 9e-09  & 7e-07 &   10 & 2.7 &  25.0 & 1e-13  & 3e-08  &   10 & 2.7 &   2.0 \\ 
 1AU6 &  161 & 3e-12 & 3e-08  & 8e-07 &   12 & 1.1 &  34.1 & 1e-13  & 2e-08  &   12 & 1.1 &   4.0 \\ 
 1PTQ &  402 & 2e-12 & 4e-08  & 8e-07 &   14 & 0.7 & 358.6 & 2e-16  & 3e-08  &   15 & 0.7 &  13.9 \\ 
 1CTF &  487 & 3e-12 & 8e-09  & 9e-07 &   13 & 0.7 & 570.3 & 1e-15  & 3e-08  &   15 & 0.8 &  18.0 \\ 
 1HOE &  558 & 1e-12 & 2e-08  & 1e-06 &   15 & 0.8 & 899.6 & 7e-16  & 1e-08  &   16 & 0.7 &  32.7 \\ 
 1LFB &  641 & 5e-13 & 1e-08  & 3e-06 &   15 & 1.2 & 2739.5 & 4e-16  & 3e-08  &   17 & 1.5 &  25.5 \\ 
 1PHT &  666 & 2e-13 & 8e-08  & 1e-06 &   15 & 1.1 & 2091.9 & 8e-16  & 4e-08  &   18 & 1.4 &  36.3 \\ 
 1F39 &  767 & 4e-12 & 2e-08  & 3e-06 &   17 & 1.3 & 3230.1 & 8e-16  & 2e-08  &   20 & 0.8 &  44.0 \\ 
 1DCH &  806 & 7e-13 & 2e-08  & 4e-06 &   18 & 1.6 & 3897.0 & 3e-16  & 4e-09  &   18 & 1.1 &  41.8 \\ 
 1HQQ &  891 & 6e-13 & 2e-08  & 4e-06 &   17 & 2.7 & 5089.2 & 7e-16  & 1e-08  &   21 & 1.0 &  43.5 \\ 
 1POA &  914 & 1e-12 & 2e-08  & 4e-06 &   16 & 2.1 & 4734.1 & 1e-15  & 4e-09  &   19 & 1.1 &  65.4 \\ 
 1AX8 & 1003 & 6e-13 & 2e-08  & 4e-06 &   17 & 3.0 & 6397.0 & 2e-16  & 4e-09  &   18 & 1.7 &  61.1 \\ 
 1TJO & 1394 & 5e-13 & 2e-08  & 5e-06 &   21 & 12.3 & - & 2e-15  & 9e-10  &   29 & 2.3 &  77.4 \\ 
 1RGS & 2015 & 3e-12 & 8e-08  & 6e-06 &   39 & 16.1 & - & 1e-15  & 4e-09  &   31 & 2.3 &  168.8 \\ 
 1TOA & 2138 & 9e-12 & 2e-07  & 8e-06 &   49 & 16.8 & - & 2e-15  & 2e-09  &   31 & 1.0 &  142.4 \\ 
 1KDH & 2846 & 3e-11 & 2e-06  & 2e-05 &  150 & 21.9 & - & 5e-16  & 2e-09  &   40 & 2.1 &  199.1 \\ 
 1NFG & 3501 & 2e-12 & 3e-05  & 1e-04 &  325 & 21.2 & - & 3e-15  & 3e-09  &   43 & 1.0 &  275.8 \\ 
 1BPM & 3672 & 1e-12 & 5e-05  & 2e-04 &  396 & 23.8 & - & 5e-17  & 8e-10  &   40 & 1.4 &  438.6 \\ 
 1MQQ & 5510 & 1e-14 & 4e-04  & 1e-03 &  911 & 26.0 & - & 2e-15  & 1e-09  &   61 & 1.3 &  947.1 \\
 \hline 
 \end{tabular}
\caption{Computational results on regularized molecular conformation
problems. ``-'' indicates that the solver is terminated because the
maximum running time of four hours is reached.}
\label{tab:my-mcp}
\end{footnotesize}
\end{table}

\section{Conclusions}
\label{sec:conclusions}
In this work, we proposed an efficient algorithm \ours for solving the
low-rank matrix optimization problem.
We utilized both the efficiency from a smooth objective of
the Burer-Monteiro decomposition approach and the convexity and
partial smoothness of the nuclear-norm-regularized convex form to
obtain a highly efficient algorithm with sound theoretical guarantees.
Extensive numerical experiments showed that our proposed algorithm
outperforms the state of the art for low-rank matrix optimization.
Based on this research, we have released an open-source package
of the proposed \ours at
\url{https://www.github.com/leepei/BM-Global/.}

\section*{Acknowledgement}
We thank Po-Wei Wang for providing us the source code of PolyMF-SS,
and Ting Kei Pong for pointing to us the work \cite{HouZSL13a}.
This work was partially done when Lee was visiting the Department of
Mathematics at the National University of Singapore.
Lee's research was supported in part by the JSPS Grant-in-Aid for
Research Activity Start-up 23K19981 and Grant-in-Aid for Early-Career
Scientists 24K20845.
Toh's research is supported by the Ministry of Education, Singapore,
under its Academic Research Fund Tier 3 grant call (MOE-2019-T3-1-010).
\putbib[bm]
\end{bibunit}

\begin{bibunit}[plainnat]
\appendix
\part{Appendices} 
\label{part:2}
\parttoc 

\input{supplementary_text.tex}

\putbib[bm]
\end{bibunit}

\end{document}

%% file: header.tex
\documentclass[twoside,11pt]{article}
\usepackage{blindtext,ntheorem}
\usepackage{jmlr2e,xr}
\usepackage{enumitem}

\usepackage{changes}
\usepackage{longtable}
\usepackage{mathtools,booktabs,multirow}
\usepackage{amsmath,amssymb}
\usepackage{graphicx,subcaption,xcolor}
\usepackage{xspace}
\graphicspath{{fig/}}
\usepackage{amsfonts}
\usepackage[ruled,linesnumbered,noend]{algorithm2e}
\usepackage[capitalize,nameinlink]{cleveref}
\crefformat{equation}{\textup{#2(#1)#3}}
\crefrangeformat{equation}{\textup{#3(#1)#4--#5(#2)#6}}
\crefmultiformat{equation}{\textup{#2(#1)#3}}{ and \textup{#2(#1)#3}}
{, \textup{#2(#1)#3}}{, and \textup{#2(#1)#3}}
\crefrangemultiformat{equation}{\textup{#3(#1)#4--#5(#2)#6}}%
{ and \textup{#3(#1)#4--#5(#2)#6}}{, \textup{#3(#1)#4--#5(#2)#6}}{, and \textup{#3(#1)#4--#5(#2)#6}}

\Crefformat{equation}{#2Equation~\textup{(#1)}#3}
\Crefrangeformat{equation}{Equations~\textup{#3(#1)#4--#5(#2)#6}}
\Crefmultiformat{equation}{Equations~\textup{#2(#1)#3}}{ and \textup{#2(#1)#3}}
{, \textup{#2(#1)#3}}{, and \textup{#2(#1)#3}}
\Crefrangemultiformat{equation}{Equations~\textup{#3(#1)#4--#5(#2)#6}}%
{ and \textup{#3(#1)#4--#5(#2)#6}}{, \textup{#3(#1)#4--#5(#2)#6}}{, and \textup{#3(#1)#4--#5(#2)#6}}

\crefdefaultlabelformat{#2\textup{#1}#3}

\newcommand{\dist}{\mbox{\rm dist}}
\newcommand{\prox}{\mbox{\rm prox}}

\newcommand{\norm}[1]{{\left\|{#1}\right\|}}
\newcommand{\R}{\mathbb{R}}
\renewcommand{\SS}{\mathbb{S}}
\renewcommand{\S}{\mathcal{S}}
\newcommand{\U}{P}

\newcommand{\N}{\mathbb{N}}

\newcommand{\inprod}[2]{\langle#1,\,#2\rangle}
\DeclareMathOperator*{\argmin}{\arg\,min}

\newcommand{\rank}{\mbox{\rm rank}}
\newcommand{\ran}{\mbox{\rm ran}}
\renewcommand{\ker}{\mbox{\rm ker}}
\newcommand{\diag}{\mbox{\rm diag}}
\newcommand{\tr}{\mbox{\rm tr}}
\newcommand{\relint}{\mbox{\rm relint}}
\newcommand{\M}{\mathcal{M}}

\newcommand{\dom}{\text{dom}}
\newcommand{\Epsilon}{\mathcal{E}}
\newcommand{\alert}[1]{\textcolor{red}{#1}}
\def\ours{{\sf BM-Global}\xspace}

\newcommand{\blue}[1]{\textcolor{blue}{#1}}

\def\cX{\mathcal{X}}

\newcommand{\TheTitle}{Accelerating Nuclear-norm Regularized Low-rank
	Matrix Optimization Through Burer-Monteiro Decomposition}
	\newcommand{\ShortTitle}{Nuclear-norm Regularized Low-rank
	Optimization Through BM Decomposition}

\ifpdf
\hypersetup{
  pdftitle={\TheTitle}
}
\fi

\author{\name Ching-pei Lee \email chingpei@ism.ac.jp\\
\addr Institute of Statistical Mathematics\\
10-3 Midori-cho, Tachikawa\\
Tokyo 190-8562, Japan
	\AND
	\name Ling Liang \email liang.ling@u.nus.edu\\
	\addr Department of Mathematics\\ University of Maryland at College Park\\
	4176 Campus Drive, College Park, MD, USA 20742
	\AND
	\name Tianyun Tang \email ttang@u.nus.edu\\
	\addr Institute of Operations
	Research and Analytics\\ National University of
	Singapore\\
	10 Lower Kent Ridge Road, Singapore 119076
	\AND
	\name Kim-Chuan Toh \email mattohkc@nus.edu.sg\\
	\addr Department of Mathematics and Institute of Operations
	Research and Analytics\\ National University of Singapore\\
	10 Lower Kent Ridge Road, Singapore 119076
}

\firstpageno{1}

%% file: supplementary_text.tex
\section{Implementation Details}
\subsection{Matrix Completion}
\label{sup:mc-implement}
We now describe our implementation details of \ours that are tailored
for the matrix completion problem.
In particular, we will discuss
the mechanism for deciding $\alpha_t$ in \cref{eq:inexact},
the algorithm for obtaining the approximate eigendecomposition using
only matrix-vector products,
details of the safeguard to ensure $\epsilon_t \downarrow 0$ in
\cref{eq:inexact}, initialization strategy for $X_0$, the solver
for \cref{eq:bm1,eq:bm2}, and the degree of parallelism of our algorithm.

To avoid redundancy, we focus on the case of \cref{eq:bm1} in this
section, and keep in our mind that it can be easily adapted to the
case of \cref{eq:bm2} by straightforward changes from SVDs to
eigendecompositions.


\subsubsection{Approximate SVD}
\label{sec:svd}
Let us denote
\begin{equation}
	Z_t \coloneqq \tilde X_t - \alpha_t\nabla f(\tilde X_t).
	\label{eq:zt}
\end{equation}
For the proximal operation \cref{eq:proxnuc}, since $m \le n$ by our
assumption, it is cheaper to compute an approximate eigendecomposition
of $Z_t Z_t^{\top}$ (instead of $Z_t^{\top} Z_t$) to get
\begin{equation}
	\label{eq:ZZ}
	Z_t Z_t^{\top} \approx \hat U_t \diag\left(
	\left(\bar \Sigma_t\right)^2 \right) \hat U_t^{\top}
\end{equation}
for some $\bar \Sigma_t \in \R^{k_t}$ with $\bar \Sigma_t > 0$ and
some orthogonal $\hat U_t \in \R^{m \times k_t}$ for a given rank $k_t$.
The notation $(\bar \Sigma_t)^2$ denotes the element-wise square.
Note that here we have removed the eigenvectors corresponding to the eigenvalue
$0$ as it does not affect the product matrix.
We then conduct an exact SVD on $\hat U_t^{\top} Z_t \in \R^{k_t \times n}$
(whose calculation can be done without forming $Z_t$ or $\tilde X_t$ explicitly)
to obtain
\begin{equation}
	\hat U_t^{\top} Z_t = \tilde U_t \diag\left(\hat \Sigma_t\right)
	V_t^{\top},
	\label{eq:exactSVD}
\end{equation}
with cost $O(k_t^3 + n k_t^2)$, which is much cheaper than SVD for $Z_t$
when $k_t \ll m$.
The approximate SVD of $Z_t$ is then obtained through
\begin{equation}
	Z_t \approx U_t \diag\left( \hat \Sigma_t \right) V_t^{\top}
	\eqqcolon \tilde Z_t, \quad U_t
	\coloneqq \hat U_t \tilde U_t,
	\label{eq:approxSVD}
\end{equation}
where $\tilde Z_t$ coincides with the one we used in \cref{eq:inexactpg,eq:errordef}.
Clearly, $U_t$ and $V_t$ are both orthonormal, so this is indeed a
valid SVD for the matrix 
\[
	\tilde Z_t = \hat U_t \hat U_t^{\top} Z_t =
P_{\text{ran}(\hat U_t)}(Z_t).
\]
The inexact proximal gradient step is then finished as
\begin{equation}
	X^+_{t}(\alpha_t) = 
		U_t \diag\left( \Sigma_t \right)
		V_t^{\top}, \quad
		\Sigma_t \coloneqq \left[ \hat \Sigma_t -
		\lambda\alpha_t e \right]_+.
	\label{eq:truncate}
\end{equation}
Note that we only store $U_t, \Sigma_t, V_t$ but do not explicitly
form $X_{t+1}$.
Here we slightly abuse the notation to let $\Sigma_t$ denote only the
coordinates of the thresholded vector that have a nonzero value,
and let $U_t$ and $V_t$ contain only the columns corresponding to
these values to save spatial and computational cost.

\subsubsection{An Efficient Algorithm for Approximate Eigendecompositions
Using Only Matrix-Matrix Products}
As mentioned before, forming $X_t$ and $\tilde X_t$ is prohibitively
expensive.
Therefore, for computing \cref{eq:ZZ}, we need to rely on iterative
methods that only require evaluating matrix multiplications involving
$Z_t$ and $Z_t^{\top}$, so that we can utilize the decomposition of
$\tilde X_t = W_t H_t^{\top}$ as well as the structured assumption of
$\nabla f(X)$ made in \cref{sec:intro}.
A highly efficient and robust approach is the limited
memory block Krylov subspace method (LmSVD) proposed by
\cite{LiuWZ13a}; see also the references therein for other popular
algorithms.
LmSVD is an extension of the classic simple subspace iteration (SSI)
method for computing the extremal eigenvalues and eigenvectors that
extends from the renowned power method (i.e., $k_t = 1$).
For any initial guess for right-singular vectors $U^{t,0}\in
\mathbb{R}^{m\times k_t}$, the SSI method computes the new iterates
$U^{t,i}$ via
\begin{equation}
    \label{eq-ssi}
    U^{t,i} \leftarrow \mathrm{orth}\left(Z_tZ_t^\top U^{t,i-1}\right),\quad \forall \;i\geq 1,
\end{equation}
where $\mathrm{orth}(M)$ extracts an orthonormal basis for the range
space of the given matrix $M$, and $i$ is the iteration counter for SSI.
For the ease of description, we abstract the operation $Z_t Z_t^{\top} (U)$
for an input $U$ as a self-adjoint semidefinite operator $L_t(\cdot)$.
Note that in each iteration of SSI, one needs to perform two matrix
multiplications (one for $Z_t^{\top}$ and the other for $Z_t$) and one
orthonormalization that cost $O(mnk_t)$ and $O(mk_t^2)$ flops,
respectively. In our case, suppose that $k_t \ll m \leq n$, then the
main computational bottleneck is the matrix products. Therefore, LmSVD tries to
accelerate the practical convergence via cutting down the total number
of iterations to reduce such matrix products without incurring additional heavy computation.
To achieve this goal, LmSVD finds the next iterate via replacing
$U^{t,i}$ in \cref{eq-ssi} with an improved candidate $\hat U^{t,i}$ via solving the following constrained optimization problem:
\begin{equation}
    \label{eq-hat-U}
	\hat U^{t,i} = \mathrm{argmax}\left\{ \inprod{U}{Z_tZ_t^\top U}\;|\; U^TU
	= I_{k_t},\; \ran(U)\subseteq \S_{t,i}\right\},
\end{equation}
where the subspace $\S_{t,i} \subseteq \R^m$ is selected as
\[
	\S_{t,i} = \ran\left(U^{t,i}, U^{t,i-1}, \dots,
	U^{t,i-p_t}\right),
	\quad U^{t,j}
	\coloneqq L_t(\hat U^{t,j-1}),\quad j=i,i-1,\dotsc,i-p_t
\]
for some pre-specified $p_t \geq 0$.
Let $q_t \coloneqq k_t(p_t+1)$ and 
\[\U^{t,i} \coloneqq [ U^{t,i},
	U^{t,i-1},
\dots,  U^{t,i-p_t}] \in \R^{m \times q_t},\]
then $U\in
\S_{t,i}$
if and only if there exists $V\in \mathbb{R}^{q_t\times k_t}$ such that 
\begin{equation}
	\label{eq-V}
	U = \U^{t,i} V.
\end{equation}
Direct computation then shows that \cref{eq-hat-U} is equivalent to
\begin{equation}
	\label{eq-blk-subspace-opt}
	\max_{V\in \mathbb{R}^{q_t\times k_t}} \;
	\inprod{V}{((\U^{t,i})^\top L_t(
	\U^{t,i} V)}\quad \mathrm{s.t.}\quad V^\top ((\U^{t,i})^\top \U^{t,i}) V =
	I_{k_t}. 
\end{equation}
However, the matrix $\U^{t,i}$ may be rank deficient to cause numerical
issues in solving \cref{eq-blk-subspace-opt}.
To resolve this issue, LmSVD replaces $\U^{t,i}$ with an orthonormal basis
of $\ran(\U^{t,i})$.
To extract such a basis,
since $U^{t,i}$ (i.e., the first block in matrix $\U^{t,i}$) always has a full
column rank, LmSVD first projects the remaining blocks in $\U^{t,i}$ to
$\ker((U^{t,i})^{\top})$ to form
\begin{equation}
    \label{eq-PUi}
	P_{t,i} \coloneqq \left(I_m - U^{t,i}(U^{t,i})^\top\right)
	\left[U^{t,i-1}, \dots, U^{t,i-p}\right]
    \in \mathbb{R}^{m\times p_t k_t},
\end{equation}
and then consider its orthonormalization.
We denote the eigendecomposition of $P_{t,i}^\top P_{t,i}
\in \mathbb{R}^{p_t k_t \times p_t k_t}$ by
\[
	P_{t,i}^\top P_{t,i} = \tilde U_{P_{t,i}} \tilde \Lambda_{P_{t,i}}
	\tilde U_{P_{t,i}}^T,
\]  
for matrices $\tilde U_{P_{t,i}}, \tilde \Lambda_{P_{t,i}} \in \R^{p_t
k_t \times p_t k_t}$ with
$\tilde U_{P_{t,i}}$ orthonormal and $\tilde \Lambda_{P_{t,i}}$ diagonal.
Clearly, if $\tilde \Lambda_{P_{t,i}}$ is nonsingular,
\begin{equation}
    \label{eq-orth-W}
	\hat \U^{t,i}\coloneqq
	\left[U^{t,i} , P_{t,i} \tilde U_{P_{t,i}}\tilde \Lambda_{P_{t,i}}^{-1/2}\right]
\end{equation}
is an orthonormal basis of $\ran(\U^{t,i})$. In practice, we can drop those
columns of $\hat \U^{t,i}$ that correspond to nearly zero eigenvalues
of  $P_{t,i}^\top P_{t,i}$.
Moreover, there may exist columns of $\hat \U^{t,i}$ whose norms are nearly
zero. To stabilize the numerical computation, one might also want to drop these columns.
From here on, we always assume that $\hat{\U}^{t,i} \in
\mathbb{R}^{m\times q_{t,i}}$, for some $q_{t,i} > 0$, forms an orthonormal
basis of $\ran(\U^{t,i})$ and it is obtained via performing the above two
trimming procedures to the matrix in \cref{eq-orth-W}.

After knowing an orthonormal basis $\hat \U^{t,i}$i of $\U^{t,i}$, we
then express any $U \in \S_{t,i}$ as 
\[
	U = \hat \U^{t,i} V
\]
for some $V \in \R^{q_{t,i} \times k_t}$.
The above expression then yields the following optimization problem to
be solved at each iteration of LmSVD.
\begin{equation}
    \label{eq-lmsvd-opt}
	\max_{V\in \mathbb{R}^{q_{t,i}\times k_t}} \; \inprod{V}{L^{t,i}
V}\quad \mathrm{s.t.}\quad V^\top V = I_{k_t}, 
\end{equation}
where $L^{t,i} \coloneqq (\hat \U^{t,i})^\top L_t ( \hat \U^{t,i}) \in
\mathbb{R}^{q_{t,i}\times q_{t,i}}$. The solution $V^{t,i}_*$ for
problem \cref{eq-lmsvd-opt} is nothing but the $k_t$ leading
eigenvectors of the matrix $L^{t,i}$. Therefore, we can compute the
full spectral decomposition of $L^{t,i}$ to get $V^{t,i}_*$ and the
computational cost is acceptable provided that $p_t$ and $k_t$ are small.
The overall algorithm of LmSVD is summarized in \cref{alg:lmsvd}.

We emphasize that LmSVD has an efficient and robust official implementation
by \cite{LiuWZ13a}.\footnote{Available at \url{https://www.mathworks.com/matlabcentral/fileexchange/46875-lmsvd-m}.}
In the present work, we borrow most parts of the implementation of
\cite{LiuWZ13a} but impose some minor modifications to adapt for our
purposes.
First, as we shall see in the following subsection, instead of using a
randomly generated initial point $U^{t,0}$, we use a more sophisticated initialization scheme.
Second, the implementation of \cite{LiuWZ13a} terminates by following a two-level strategy,
but in our implementation, we simply terminate the algorithm as long
as the difference between the eigenvalues of $L^{t,i}$ and $L^{t,i-1}$ is
small.

\begin{algorithm}[tb]
\DontPrintSemicolon
\SetKwInOut{Input}{input}\SetKwInOut{Output}{output}
\SetKwComment{Comment}{*}{}
\caption{$\textrm{LmSVD}(L_t, U^{t,0}, p_t)$}
\label{alg:lmsvd}
\Input{A self-adjoint semidefinite operator $L_t$ over $\mathbb{R}^m$,
an initial guess $U^{t,0}\in \mathbb{R}^{m\times k_t}$, a threshold
$\epsilon > 0$ for ruling out small eigenvalues, and an integer
$p_t > 0$} 

$U^{t,0}\leftarrow \mathrm{orth}(U^{t,0})$

\For{$i = 0,\dots, $}{

    $p \leftarrow \min\{p_t, i\}$
    
	$\S_{t,i}\leftarrow \ran\left(U^{t,i}, U^{t,i-1}, \dots,
	U^{t,i-p}\right)$ \Comment*{Block Krylov subspace selection}
    
	$\hat U^{t,i} \leftarrow \mathrm{argmax}\left\{
		\inprod{U}{L_t(U)}\;|\; U^\top  U = I_{k_t}, U\in
		\S_{t,i}\right\}$ \Comment*{Block subspace optimization}
    
		$U^{t,i+1} \leftarrow \mathrm{orth}\left(L_t(\hat U^{t,i})\right)$ \Comment*{Orthonormalization}
    
}

$ s \leftarrow \left(\mathrm{diag}\left((U^{t,i+1})^\top
L(U^{t,i+1})\right)\right)^{1/2}$

$J_s \leftarrow \mathrm{find}(s > \epsilon) $ \Comment*{Rule out small eigenvalues and associated eigenvectors}

$\hat U_t \leftarrow (U^{t,i+1})_{:, J_s}$


\Output{$\hat U_t$}
\end{algorithm}

%
%
%
%
%
%
%

\subsubsection{Ensuring Sufficient Precision in the Proximal Operation}
We notice that \cref{eq:truncate} suggests that all entries in $\bar
\Sigma_t$ smaller than the threshold as well as their corresponding columns of
$V_t$ and $U_t$ do not contribute to the calculation of $X^+_t(\alpha_t)$,
so we just need to compute the entries not truncated by the proximal
operation.
Therefore, if $\rank(X_{t+1}) =k_{t+1}$, ideally we just need to
compute the first $k_{t+1}$ eigenvalues in our approximate
eigendecomposition at the $t$th iteration of proximal gradient.
On the other hand, to ensure that we are recovering a global solution
$X^*$ of \cref{eq:matrixform}, it is necessary to check that the ranks
of the iterates are large enough so that we do not get stuck at an
approximation of $X^*$ with an insufficient rank.
To safeguard that our algorithm converges to a global optimum, or
more explicitly, to make $\epsilon_t$ in \cref{eq:inexact}
decrease to $0$ fast enough (see \cref{sec:analysis}), we want
to ensure that eventually the smallest eigenvalue we obtain will be
truncated out, so that we can be certain that all
eigenvalues/eigenvectors that contribute to the computation of
$X^+_t(\alpha_t)$ have already been obtained.
Therefore, we will need a mechanism to adaptively adjust the rank of
$X_t$.
As the decrease of the rank is achieved by the truncation in the
proximal operation,
following our usage of LmSVD described in the previous subsection,
what we need is a way to make the initial guess $U^{t,0}$ input to
LmSVD have a rank sufficiently higher than that of $X_t$ and $\tilde X_t$.

As noted in \cref{lemma:equivalent}, we know that the output of
approximately solving \cref{eq:bm1} should be close to the singular
vectors of $\tilde X_t$ (up to column-wise scaling).
Moreover, when $X_t$ and $\tilde X_t$ are close to a global optimum
$X^*$, we expect that $X_{t+1}$ will be close to $X^*$ and thus
also to $X_t$ and $\tilde X_t$, so the SVD of $X_{t+1}$ is also expected to be
close to that of $X_t$ and $\tilde X_t$.
We therefore use $U_t$ from \cref{eq:approxSVD} and $W_t$ from
the output of the BM phase to form
\begin{equation}
	\hat U^{t,0} \coloneqq \text{orth}\left( \left[ U_t, W_t \right] \right)
	\label{eq:rt0}
\end{equation}
as the base of our the warmstart input to the approximate eigendecomposition in
obtaining $X_{t+1}$ from $Z_t$.
If the BM phase is not entered or does not produce any change in the iterate,
we use
\begin{equation}
	\hat U^{t,0} \coloneqq \text{orth}\left( \left[ U_t, U_{t-1}\right] \right)
	\label{eq:rt0-1}
\end{equation}
instead.
To further guarantee that $\epsilon_t \rightarrow 0$ in
\cref{eq:inexact}, we need to ensure that the rank of $ U^{t,0}$ is
sufficiently large, and that $\hat U^{t,0}$ will approach the singular
vectors corresponding to the singular values not truncated out.
Ideally, we hope that the output of our approximate eigendecomposition
will be exactly all the eigenvalues or singular values that are
retained nonzero, plus the largest one that is truncated out in
\cref{eq:proxnuc}.
Therefore, we add in one column with randomness to $R_t$ whenever
the rank of $X_{t}$ and $X_{t-1}$ are the same (namely, the rank has
stopped increasing) and there is at most
one eigenvalue truncated out in the inexact proximal gradient step at
the $(t-1)$th iteration.
The idea is that the case of truncating only one eigenvalue is the
ideal scenario we want and we want the next iteration to still have
one eigenvalue to truncate as the safeguard, while if no truncation
happened, then we should continue increasing the rank.
Utilizing this idea, we retain the singular vector $u$ that
corresponds to the largest truncated singular value in the latest
iteration where such a truncation took place,
and compute its projection $u_t$ to $\ker( (\hat U^{t,0})^{\top})$.
(It is possible and acceptable that $u_t = 0$.)
The warmstart input is finally formed by
\begin{equation}
	U^{t,0} \coloneqq \text{orth}\left( \left[\hat U^{t,0}, u_t + \xi_t\right] \right),
	\quad \xi_t \in \text{ker}\left( [\hat U^{t,0}, u_t]^{\top}
\right), \quad \norm{\xi_t} \leq \psi_t,
\label{eq:rt}
\end{equation}
where $\xi_t$ is a random vector and $\{\psi_t\}$ is a sequence such
that $\psi_t \downarrow 0$.

When the rank of $X_{t}$ and $U^{t-1,0}$ are the same, it means no
truncation took place in the proximal operation, and we view this as
that the maintained $u$ has been added to $U^{t,0}$ as a column in the next
iteration when we call LmSVD to compute an approximate eigendecomposition.
In this situation, we then seek the eigenvector that
corresponds to the next eigenvalue truncated as our new $u$.
When there is no more such vectors available, we simply add in a unit
random vector that is orthogonal to the columns of $\hat U^{t,0}$.

Here we provide further explanations to our design above.
Assume that the eigenvalues in the exact eigendecomposition of
$Z_tZ_t^{\top}$ are $\sigma_1 \geq \sigma_2 \geq \dotsc \geq
\sigma_m$,
$S_1 \coloneqq \{\sigma_1,\dotsc,\sigma_k\}$ is the set of those
eigenvalues that will not become zero after the truncation in
\cref{eq:truncate}, and $S_2 \coloneqq
\{\sigma_{k+1},\dotsc,\sigma_m\}$ are those that will become zero
after the truncation in \cref{eq:truncate}.
To cope with pathological cases in which some eigenvectors corresponding to
some $S_3 \subsetneq S_1$ and some corresponding to $S_4 \subset S_2$
are obtained, we inject noise to $u$ so that during the procedure of
\cref{alg:lmsvd}, it will approach an eigenvector that corresponds to
some value in $S_1 \setminus S_3$ instead of getting stuck at an
eigenvector that will be truncated out.
(Analysis of the classical SSI suggests that $U^{t,i}$ approaches the
leading eigenvectors as long as no column is exactly a multiple of an
eigenvector that corresponds to an eigenvalue with a smaller absolute
value.)
On the other hand, when we are close to an optimal solution $X^*$ of
\cref{eq:matrixform},
and the eigenvectors corresponding to $S_1$ are all identified or
well-approximated, it is natural that we do not want to add in much
noise in the initialization of \cref{alg:lmsvd}, as such noise will
decelerate the convergence of \cref{alg:lmsvd}.
Therefore, in our design, we only add $u$ with noise to $U^{t,0}$ when
$\rank(X_{t}) = \rank(X_{t-1})$ or $\rank(U^{t-1,0}) \le
\rank(X_{t})+1$, namely when no truncation happened or when only one
vector is truncated.
In the latter case, adding $u$ to $U^{t,0}$ is for the purpose of making
$U^{t,0}$ contain one eigenvector that is likely to be truncated, so that
we can still have the safeguard for ensuring that we have found the correct
rank in the approximate eigendecomposition.
We then decrease the level of noise by a certain factor whenever
$\rank(X_t) = \rank(R_t) - 1$.  That is, when exactly only one
eigenvalue is truncated.
This corresponds to \cref{eq:rt}.
For the noise level $\psi_t$, in our implementation, we start with
$\psi_0 = 0.1$ so that the noise will not dominate $u_t + \xi_t$, and
whenever we need to decrease the significance of the noise, we just
let $\psi_t = \psi_{t-1}/2$, and otherwise we assign $\psi_t =
\psi_{t-1}$.

\subsubsection{Initialization}
It can be clearly seen that the point of origin is a saddle point of
\cref{eq:bm1}, and this is also the case for many other popular
problems that has the form of \cref{eq:bm}.
Therefore, it is essential to have an effective way to initialize
$X_0$, or equivalently $(\tilde W_0, \tilde H_0)$, in
\cref{alg:mfstride}.
Existing methods for \cref{eq:bm} usually take a random
initialization, but such approaches often lead to an unideal initial
objective even worse than using $X_0 = 0$.
Moreover, it is hard to decide what is an appropriate value for the
initial rank -- too large the rank takes longer running time, but too
small the rank might lead to slow convergence at the early stage.
The most straightforward idea would be to conduct one (inexact)
proximal gradient step from the origin, but the difficulty is that we
will be in lack of a warmstart matrix for the approximate
eigendecomposition in \cref{alg:lmsvd}, and we still need to decide
the rank of this matrix.

To get a good initialization for $X_0$, we follow the recent
developments in randomized numerical linear algebra by
\cite{HalMT11a,Mar19a} to combine the HMT method \citep{HalMT11a}
with the Nystr\"om method \citep{Nys30a}.
We imagine that our starting point is actually $X_{-1} = 0$, and then
conduct one step of inexact proximal gradient from there with the fixed
stepsize $\alpha_{-1} = L^{-1}$ to initialize $X_{-1}$.
Regarding the approximate SVD for
\[
	Z_{-1} \coloneqq X_{-1} -
	\alpha_{-1} \nabla f(X_{-1}) = -L^{-1} \nabla f(0),
\]
we describe how to obtain the eigendecomposition of $A_{-1} \coloneqq
Z_{-1} Z_{-1}^{\top}$.
Given the initial rank $k_{-1}$, we start with a random matrix
$\tilde U^{-1,0} \in \R^{m \times k_{-1}}$ whose entries are independently
and identically distributed as the standard normal distribution.
Then we conduct
\[
	Q_{-1} = \text{orth}\left(A_{-1} \tilde U^{-1,0}\right)
\]
as in the HMT method, and use this $Q_{-1}$ as the sketching matrix in
the Nystr\"om method to consider the approximation
\begin{equation}
	A_{-1} \approx A_{-1} Q_{-1}
	\left( Q_{-1}^{\top} A_{-1} Q_{-1} \right)^\dag \left(A_{-1}
	Q_{-1}\right)^{\top} \eqqcolon \hat
	A_{-1},
	\label{eq:nys}
\end{equation}
where for any matrix $B$, $B^\dag$ is its pseudo inverse.
The approximation matrix $\hat A_{-1}$ here is not really explicitly
computed, but only serves as an intermediate variable for our further
process.
Clearly, $\rank(\hat A_{-1}) \leq k_{-1}$, and thanks to the randomness
from $\tilde U_{-1,0}$ and therefore $Q_{-1}$, with high probability we
have $\rank(\hat A_{-1}) = k_{-1}$.
We can then compute the exact eigendecomposition for $\hat A_{-1}$ by
separately considering $A_{-1} Q_{-1}$ and $(Q_{-1}^{\top} A_{-1}
Q_{-1} )^\dag$.
As long as $k_{-1}$ is small, the computation of both $A_{-1} Q_{-1}$
and $Q_{-1}^{\top} A_{-1} Q_{-1} \in \R^{k_{-1} \times k_{-1}}$ is
affordable under our assumption of efficient matrix-matrix products
involving $\nabla f$, and so is the calculation of the pseudo inverse
that costs $O(k_{-1}^3)$.
For obtaining the exact eigendecomposition of $\hat A_{-1}$, we first
compute a QR decomposition of $A_{-1} Q_{-1}$
\[
	\hat Q_{-1} \hat R_{-1} = A_{-1} Q_{-1},
\]
where $\hat Q_{-1}$ is an orthonormal matrix and
$\hat R_{-1}$ is upper triangular.
We then obtain the eigendecomposition of
\[
\hat R_{-1} (Q_{-1}^{\top} A_{-1} Q_{-1} )^\dag \hat R_{-1}^{\top} =
\tilde U_{-1} \diag\left( (\hat \Sigma_{-1})^2 \right)
\tilde U_{-1}^{\top}
\]
which costs only $O(k_{-1}^3)$ in both forming the matrix on the
left-hand side and calculating its eigendecomposition, as the matrices
involved are all $k_{-1}$ by $k_{-1}$.
When $k_{-1} \ll m$, this cost is negligible in comparison to other steps.
The eigendecomposition of $\hat A_{-1}$ is then obtained by
setting $\hat U_0 = \hat Q_{-1} \tilde U_{-1}$ to get $\hat U_0$ in the
right-hand side of \cref{eq:ZZ}.
The next steps for initializing $X_0$ then directly follow the
procedure of \cref{eq:exactSVD,eq:approxSVD,eq:truncate}.

The remaining issue is to decide the rank $k_{-1}$ for $\tilde R_{-1}$.
The most naive idea is to set $k_{-1} = 1$ so that all computation in the
initialization is of the lowest possible overhead.
However, in this case, we will not be able to fully exploit the
multicore advantage of modern computing devices.
We therefore set the initial $k_{-1}$ to be the number of cores we
can use, so that all the matrix-matrix computation of this step can be
fully parallelized without increasing the overhead for initialization.

\subsubsection{Solver for \cref{eq:bm1}}
The step of optimizing \cref{eq:bm1} relies on an off-the-shore
solver,
and to obtain the best efficiency, the choice should be application-dependent.
As we aim for large-scale problems and wish to fully exploit multicore
parallelization ubiquitous in modern computers, we will use methods
that efficiently utilize multiple computational cores.
For \cref{eq:bm1}, as the objective is smooth, many algorithms are
available for this choice, ranging from asynchronous, low-order ones to
synchronous, high-order approaches.
For the low-order ones, we note that since each row of $W$ affects
different entries in $WH^{\top}$, one can update multiple rows
simultaneously if $f(X)$ is separable, and the same argument applies
to the update of $H$ as well.
Therefore, as long as $k$ is no fewer than the number of cores, we
should be able to enjoy full parallelism.
For problems like \cref{eq:mf}, many efficient algorithms such as
those by \cite{YuHSD14a,ZhuCJL13a,WanLK17a} are readily available.
On the other hand, for \cref{eq:bm2}, the property of the constraint
set $C$ will limit our choices for the solver.
In most applications, however, the constraint $WW^{\top} \in C$ can be
formulated as a smooth manifold $\mathcal{M}$ for $W$, and one can
utilize efficient manifold optimization approaches for such problems;
see, for example, \cite{AbsMS09a,manopt}.
The major computation in such manifold optimization approaches is
usually the computation of the Riemannian gradient and the
Hessian-vector products, which are by nature parallelizable.


\subsubsection{Parallelism}
A key to cope with large-scale problems in modern computing
environments is to utilize multicore  parallelization.
Although inherently the bottleneck of \cref{alg:lmsvd}, \cref{eq-ssi},
can be highly parallelized, in practice usually the parallelism is low
as this operation is data-heavy but not computation-heavy, so higher
parallelism is hindered by the memory bandwidth.
On the other hand, solvers for \cref{eq:bm1} tend to be more
computationally intensive, so they can often achieve better
parallelism, by utilizing all cores available, than solvers for
\cref{eq:matrixform}.
By switching to the BM phase, our algorithm hence exploits better
parallelism than state of the art for \cref{eq:matrixform}.
The initial rank $k$ in the previous subsection is also selected to
fully exploit the advantage of multiple cores from the beginning on.

\subsection{Quadratic SDP}

\subsubsection{The BM formulations for the RKE problem and the molecular conformation problem.}

For the RKE problem and the molecular conformation problem, the corresponding factorized problem is given as  
\begin{equation}
\min_{W\in \R^{n \times k}}
\Big(g\left( W \right) \coloneqq \frac{1}{2} \sum_{(i,j)\in\Omega} w_{ij} \left(\inprod{E_{ij}}{WW^\top }-d_{ij}^2\right)^2 + \lambda\norm{W}_F^2
\Big), \quad \text{s.t.} \quad W^\top e = 0.
\label{eq:EDM}
\end{equation}
The gradient of this function $g:\R^{n\times k}\rightarrow \R$ is
\[
    \nabla g(W) = 2\left(\sum_{(i,j)\in\Omega} w_{ij} \left(\inprod{E_{ij}}{WW^\top }-d_{ij}^2\right)E_{ij} +\lambda I\right)W.
\]
For any given $D\in \R^{n\times k}$, the Hessian operator of $g$ performed on $H$ can be computed as
\[
   \begin{aligned}
         \nabla^2g(W)[D] = &\; 2\sum_{(i,j)\in \Omega}w_{ij}\inprod{E_{ij}}{WD^\top + D W^\top} E_{ij}W \\
         &\; + 2\sum_{(i,j)\in \Omega}w_{ij}\left( \inprod{E_{ij}}{WW^\top} - d_{ij}^2\right)E_{ij}D + 2\lambda D.
   \end{aligned}
\]

\subsubsection{Implementation Details}
Recall that for a given problem size $n$, \cref{alg:mfstride}
iterates on a low-rank matrix $W_t\in \mathbb{R}^{n\times k_t}$ with
$k_t < n$ dynamically adjusted.
In our experiment, we choose the initial $k_0$ as
\[
	k_0 = \min\{100, \lfloor 0.15 n\rfloor \}.
\]
Given $k_0$, we then run APG for $100$ iterations with this rank
constraint to generate an initial point $W_0$.
When applying the APG method, we never form the matrix $X \coloneqq
WW^\top$ and use only eigendecompositions with rank $k_0$.
In particular, we use MATLAB's built-in \texttt{eigs} subroutine to
compute the largest $k$ eigenpairs.
After the initialization stage,
at the $t$-th iteration of \cref{alg:mfstride}, we first apply
the Manopt solver to solve the factorized problem \cref{eq:EDM} with
the initial point $W_t$ to compute a matrix $\tilde{W}_t$.
Using $\tilde{W}_t$, we then perform one inexact PG step with an
approximate eigendecomposition with fixed step size $\alpha_t = 1/L$.
If all the computed eigenvalues are positive, then it is highly
possible that the current rank is too small. In this case, we set
$k_{t+1}\leftarrow k_t + \lfloor \frac{n}{20}\rfloor$. Otherwise, we set $k_{t+1}$ as the number of positive
eigenvalues returned by \texttt{eigs}.

\subsubsection{The accelerated proximal gradient method}
Since we utilize the APG method for the initialization of \ours for
\cref{eq-qsdp},
for completeness, we provide a concise description of the APG method
in \cref{alg:apg}.
To perform the projection $P_\mathcal{X}$ for \cref{eq-qsdp} through  \cref{lemma-PiX},
we use the \texttt{eig} subroutine in MATLAB.
(Note that the PG method is equivalent to setting $\beta_t \equiv 1$
in \cref{alg:apg}.)

\begin{algorithm}[tb]
\DontPrintSemicolon
\SetKwInOut{Input}{input}\SetKwInOut{Output}{output}
\SetKwComment{Comment}{*}{}
\caption{The APG method for solving \cref{eq-qsdp}}
\label{alg:apg}
\Input{Initial point: $Y_0 = X_0 \in \SS^n$, Lipschitz constant $L>0$ and $\beta_0 = 1$ } 

\For{$t = 0,\dots, $}{

    $X_{t+1} \leftarrow P_{\mathcal{X}}(Y_t - L^{-1} \nabla f(Y_t))$

    $\beta_{t+1} \leftarrow \frac{1 + \sqrt{1 + 4\beta_t^2}}{2}$

    $Y_{t+1} \leftarrow X_{t+1} + \frac{\beta_{t}-1}{\beta_{t+1}}(X_{t+1} - X_t)$
    
}

\Output{$X_{t+1}$}
\end{algorithm}

\section{Additional Experimental Details and More Experiments}
\label{supp:exp}

\subsection{The SpaRSA variant for Matrix Completion}
It is known that the convergence speed of standard PG could
be slow, thus we also considered following \cite{WriNF09a} to use the
Barzilai-Borwein (BB) initialization \citep{barzilai1988two} with
linesearch to decide the step size to accelerate the practical
convergence of the convex lifting step.
Given $\alpha_{\max} \geq \alpha_{\min} > 0$, we first compute
\begin{equation}
	\alpha_t^{\text{BB}} \coloneqq \max\left\{\alpha_{\min},
	\min\left\{\alpha_{\max},\frac{ \inprod{X_t-X_{t-1}}{\nabla
	f(X_t)-\nabla f(X_{t-1})}}{\norm{X_t-X_{t-1}}_F^2}\right\}\right\}.
\label{eq:proxBB}
\end{equation}
But different from \cite{WriNF09a}, our backtracking
linesearch does not search for an $\alpha_t$ that gives sufficient
descent.
Instead, given $\beta,\delta \in (0,1)$, we find the smallest nonnegative
integer $i$ such that $\alpha_t = \alpha_t^{\text{BB}} \beta^i$
satisfies \cref{eq:back} or \cref{eq:decrease}.

In our scenario, we only have the factorized form $X_t=W_t H_t^\top$
for each $t$ but not the iterate $X_t$ itself, so we use the following
formula to calculate the numerator of \cref{eq:proxBB}, which is also
used as the last term in \cref{eq:back}.
\begin{align}
\nonumber
&~\norm{X_t - X_{t-1}}_F^2 \\
\nonumber
=&~ \inprod{X_t-X_{t-1}}{X_t-X_{t-1}} \\
\nonumber
=&~ \inprod{X_t}{X_t} + \inprod{X_{t-1}}{X_{t-1}} - 2
\inprod{X_t}{X_{t-1}}\\
\label{eq:trace}
=&~ \tr\left( (W_t^{\top} W_t) (H_t^{\top} H_t)
\right) + \tr\left( (W_{t-1}^{\top} W_{t-1})
(H_{t-1}^{\top} H_{t-1}) \right) + \tr\left(
(W_{t-1}^{\top} W_t) (H_{t-1}^{\top} H_{t})
\right).
\end{align}
Similar calculation is also applied to compute the denominator of
\cref{eq:proxBB}.
If $\rank(X_t) = k_t$ and $\rank(X_{t-1}) = k_{t-1}$,
the cost of computing \cref{eq:trace} is $O((m+n)(k_t^2 + k_{t-1}^2 +
k_t k_{t-1})$.

\subsection{Experimental Details of  \cref{subsec:mc-parameter}}
We report the required running time for making the relative objective
smaller than $\epsilon \in \{10^{-4}, 10^{-8}, 10^{-12}\}$ for both
the fixed-step variant and the SpaRSA variant of our algorithm.
The algorithms are terminated either when they have reached $\epsilon
= 10^{ -12}$ or when they have conducted $100$ inexact
proximal gradient steps.
The results are shown in \cref{tbl:toy,tbl:ml100k}.

Aside from the selection of $x=1$ and $y=3$ made in the main paper for
the fixed-step variant,
for SpaRSA, we also see that $x=1$ and $y=3$ is still robust and we
thus use this setting for SpaRSA from now on.

\begin{table}
\caption{Time (seconds) required for solving \cref{eq:mc} on
toy-example using \cref{alg:mfstride} to make \cref{eq:obj} no
larger than $\epsilon$. We alternate between $x$ consecutive inexact
proximal gradient steps and $y$ consecutive iterations of the BM phase
solver.  The fastest one for each case is labeled in red. ``-''
indicates that the designated $\epsilon$ is not reached within $100$
inexact proximal gradient steps.}
\label{tbl:toy}
	\centering
	\begin{tabular}{@{}ll|cccccccc@{}}
\multicolumn{9}{c}{$\epsilon = 10^{-4}$}\\
\toprule
& $y$ & $1$& $2$& $3$& $ 4$& $ 5$& $6$ & $7$ & $8$\\
\midrule
\multirow{5}{*}{SpaRSA} & $x = 1$ & 5.4e$+$0& 1.1e$+$0& 1.2e$+$0& 1.4e$+$0& 1.5e$+$0& 1.6e$+$0& 1.7e$+$0& 1.8e$+$0\\
& $x = 2$ & 1.4e$+$0& 1.3e$+$0& 1.8e$+$0& 1.7e$+$0& 1.9e$+$0& 2.1e$+$0& 2.4e$+$0& 2.6e$+$0\\
& $x = 3$ & 1.1e$+$0& 1.1e$+$0& 1.1e$+$0& 1.2e$+$0& 1.2e$+$0& 1.2e$+$0& 1.3e$+$0& 1.3e$+$0\\
& $x = 4$ & 1.3e$+$0& 1.0e$+$0& 1.1e$+$0& 1.2e$+$0& 1.6e$+$0& 1.7e$+$0& 1.8e$+$0& 1.8e$+$0\\
& $x = 5$ & \alert{1.0e$+$0}& 1.1e$+$0& 1.3e$+$0& 1.3e$+$0& 1.4e$+$0& 1.5e$+$0& 1.6e$+$0& 1.7e$+$0\\
\midrule
\multirow{5}{*}{$\alpha_t \equiv 1$} & $x = 1$ & 1.2e$+$0& \alert{1.0e$+$0}& 1.2e$+$0& 1.3e$+$0& 1.4e$+$0& 1.6e$+$0& 1.7e$+$0& 2.0e$+$0\\
& $x = 2$ & 1.2e$+$0& 1.5e$+$0& 1.5e$+$0& 1.7e$+$0& 1.9e$+$0& 2.1e$+$0& 2.3e$+$0& 2.5e$+$0\\
& $x = 3$ & 1.5e$+$0& 1.2e$+$0& 1.1e$+$0& 1.2e$+$0& 1.2e$+$0& 1.3e$+$0& 1.3e$+$0& 1.4e$+$0\\
& $x = 4$ & 1.2e$+$0& 1.3e$+$0& 1.3e$+$0& 1.4e$+$0& 1.5e$+$0& 1.6e$+$0& 1.7e$+$0& 1.7e$+$0\\
& $x = 5$ & 1.3e$+$0& 1.2e$+$0& 1.3e$+$0& 1.4e$+$0& 1.5e$+$0& 1.6e$+$0& 1.8e$+$0& 1.9e$+$0\\
\bottomrule
\multicolumn{9}{c}{}\\
\multicolumn{9}{c}{$\epsilon = 10^{-8}$}\\
\toprule
\multirow{5}{*}{SpaRSA} & $x = 1$ & 1.2e$+$1& 3.7e$+$0& 4.9e$+$0& 5.0e$+$0& 5.5e$+$0& 5.8e$+$0& 7.0e$+$0& 7.1e$+$0\\
& $x = 2$ & 3.6e$+$0& 6.3e$+$0& 4.3e$+$0& 4.9e$+$0& 4.7e$+$0& 5.8e$+$0& 5.3e$+$0& 5.6e$+$0\\
& $x = 3$ & 3.2e$+$0& 3.6e$+$0& 4.2e$+$0& 4.1e$+$0& 4.5e$+$0& 4.8e$+$0& 5.2e$+$0& 5.7e$+$0\\
& $x = 4$ & 3.3e$+$0& 3.2e$+$0& \alert{3.2e$+$0}& 3.4e$+$0& 3.6e$+$0& 3.8e$+$0& 4.1e$+$0& 4.3e$+$0\\
& $x = 5$ & 3.3e$+$0& 3.7e$+$0& 3.5e$+$0& 4.4e$+$0& 4.5e$+$0& 4.8e$+$0& 5.1e$+$0& 5.4e$+$0\\
\midrule
\multirow{5}{*}{$\alpha_t \equiv 1$} & $x = 1$ & 4.4e$+$0& 3.7e$+$0& 4.9e$+$0& 5.4e$+$0& 6.4e$+$0& 7.1e$+$0& 7.7e$+$0& 8.4e$+$0\\
& $x = 2$ & 4.4e$+$0& 3.8e$+$0& 4.2e$+$0& 4.6e$+$0& 5.1e$+$0& 5.2e$+$0& 5.5e$+$0& 5.7e$+$0\\
& $x = 3$ & 4.4e$+$0& 3.8e$+$0& 4.1e$+$0& 4.6e$+$0& 4.9e$+$0& 5.2e$+$0& 5.6e$+$0& 6.0e$+$0\\
& $x = 4$ & 5.0e$+$0& \alert{3.6e$+$0}& 3.6e$+$0& 3.9e$+$0& 4.0e$+$0& 4.3e$+$0& 4.5e$+$0& 4.7e$+$0\\
& $x = 5$ & 4.9e$+$0& 4.1e$+$0& 4.2e$+$0& 4.5e$+$0& 4.5e$+$0& 4.8e$+$0& 5.3e$+$0& 5.3e$+$0\\
\bottomrule
\multicolumn{9}{c}{}\\
\multicolumn{9}{c}{$\epsilon = 10^{-12}$}\\
\toprule
\multirow{5}{*}{SpaRSA} & $x = 1$ & 1.6e$+$1& 7.4e$+$0& 8.1e$+$0& 7.9e$+$0& 9.6e$+$0& 9.1e$+$0& 1.0e$+$1& 1.2e$+$1\\
& $x = 2$ & 1.1e$+$1& 8.9e$+$0& 7.7e$+$0& 8.2e$+$0& 8.4e$+$0& 9.4e$+$0& 9.2e$+$0& 9.0e$+$0\\
& $x = 3$ & 1.0e$+$1& 7.0e$+$0& 7.2e$+$0& 7.2e$+$0& 7.1e$+$0& 7.3e$+$0& 7.8e$+$0& 8.6e$+$0\\
& $x = 4$ & 9.0e$+$0& 7.6e$+$0& 7.1e$+$0& 7.0e$+$0& 7.4e$+$0& 7.9e$+$0& 7.6e$+$0& 7.5e$+$0\\
& $x = 5$ & 1.1e$+$1& 7.8e$+$0& \alert{6.2e$+$0}& 6.8e$+$0& 6.9e$+$0& 7.3e$+$0& 6.9e$+$0& 7.9e$+$0\\
\midrule
\multirow{5}{*}{$\alpha_t \equiv 1$} & $x = 1$ & 1.2e$+$1& 6.8e$+$0& 8.0e$+$0& 7.7e$+$0& 8.8e$+$0& 9.3e$+$0& 1.0e$+$1& 1.0e$+$1\\
& $x = 2$ & 1.7e$+$1& 6.9e$+$0& 7.1e$+$0& 7.1e$+$0& 7.8e$+$0& 7.8e$+$0& 8.4e$+$0& 8.7e$+$0\\
& $x = 3$ & 1.4e$+$1& 7.2e$+$0& 6.9e$+$0& 7.9e$+$0& 7.8e$+$0& 8.0e$+$0& 8.6e$+$0& 8.5e$+$0\\
& $x = 4$ & 1.3e$+$1& 7.7e$+$0& 6.9e$+$0& 7.4e$+$0& 7.7e$+$0& 8.0e$+$0& 8.6e$+$0& 8.8e$+$0\\
& $x = 5$ & 1.7e$+$1& 7.6e$+$0& \alert{6.5e$+$0}& 7.1e$+$0& 7.4e$+$0& 7.9e$+$0& 8.4e$+$0& 8.3e$+$0\\
\bottomrule
\end{tabular}
\end{table}

\begin{table}
\caption{Time (seconds) required for solving \cref{eq:mc} on
movielens100k using \cref{alg:mfstride} to make \cref{eq:obj} no
larger than $\epsilon$. We alternate between $x$ consecutive inexact
proximal gradient steps and $y$ consecutive iterations of the BM phase
solver.  The fastest one for each case is labeled in red. ``-''
indicates that the designated $\epsilon$ is not reached within $100$
inexact proximal gradient steps.}
\label{tbl:ml100k}
\centering
\begin{tabular}{@{}ll|cccccccc@{}}
\multicolumn{10}{c}{$\epsilon = 10^{-4}$}\\
\toprule
 & $y$ & $1$& $2$& $3$& $ 4$& $ 5$& $6$ & $7$ & $8$\\
\midrule
\multirow{5}{*}{SpaRSA} & $x = 1$ & 5.7e$-$1& 5.2e$-$1& \alert{4.1e$-$1}& 4.6e$-$1& 5.1e$-$1& 5.4e$-$1& 5.7e$-$1& 6.0e$-$1\\
&$x = 2$ & 2.0e$+$0& 1.4e$+$0& 7.0e$-$1& 6.8e$-$1& 5.7e$-$1& 5.6e$-$1& 4.9e$-$1& 9.5e$-$1\\
&$x = 3$ & 3.8e$+$0& 2.4e$+$0& 1.6e$+$0& 1.6e$+$0& 9.1e$-$1& 9.3e$-$1& 8.0e$-$1& 5.8e$-$1\\
&$x = 4$ & 4.1e$+$0& 2.9e$+$0& 2.3e$+$0& 1.7e$+$0& 1.3e$+$0& 1.3e$+$0& 7.7e$-$1& 7.4e$-$1\\
&$x = 5$ & 3.9e$+$0& 2.6e$+$0& 2.1e$+$0& 1.5e$+$0& 1.5e$+$0& 1.4e$+$0& 1.1e$+$0& 9.7e$-$1\\

\midrule
\multirow{5}{*}{$\alpha_t \equiv 1$} &$x = 1$ & 7.8e$-$1& 4.5e$-$1& 4.4e$-$1& \alert{4.3e$-$1}& 5.3e$-$1& 5.7e$-$1& 6.0e$-$1& 6.3e$-$1\\
&$x = 2$ & 1.6e$+$0& 8.0e$-$1& 5.7e$-$1& 5.2e$-$1& 4.5e$-$1& 4.9e$-$1& 5.2e$-$1& 5.5e$-$1\\
&$x = 3$ & 2.2e$+$0& 1.4e$+$0& 1.0e$+$0& 1.1e$+$0& 8.1e$-$1& 7.7e$-$1& 5.3e$-$1& 5.6e$-$1\\
&$x = 4$ & 2.5e$+$0& 1.6e$+$0& 1.3e$+$0& 1.3e$+$0& 1.0e$+$0& 9.4e$-$1& 6.6e$-$1& 6.3e$-$1\\
&$x = 5$ & 2.8e$+$0& 1.9e$+$0& 1.6e$+$0& 1.4e$+$0& 1.1e$+$0& 1.1e$+$0& 8.2e$-$1& 8.1e$-$1\\
\bottomrule
\multicolumn{10}{c}{}\\
\multicolumn{10}{c}{$\epsilon = 10^{-8}$}\\
\toprule
\multirow{5}{*}{SpaRSA} &$x = 1$ & 2.4e$+$0& 2.2e$+$0& 1.9e$+$0& 1.7e$+$0& 1.6e$+$0& \alert{1.6e$+$0}& 1.7e$+$0& 1.7e$+$0\\
&$x = 2$ & 4.6e$+$0& 4.2e$+$0& 3.2e$+$0& 2.4e$+$0& 2.1e$+$0& 1.8e$+$0& 1.7e$+$0& 1.6e$+$0\\
&$x = 3$ & 7.0e$+$0& 5.2e$+$0& 4.4e$+$0& 3.5e$+$0& 2.9e$+$0& 2.7e$+$0& 2.7e$+$0& 2.4e$+$0\\
&$x = 4$ & 7.6e$+$0& 5.9e$+$0& 4.8e$+$0& 4.0e$+$0& 3.5e$+$0& 3.1e$+$0& 3.0e$+$0& 2.6e$+$0\\
&$x = 5$ & 7.2e$+$0& 5.8e$+$0& 4.6e$+$0& 4.2e$+$0& 3.6e$+$0& 3.4e$+$0& 3.3e$+$0& 2.9e$+$0\\
\midrule
\multirow{5}{*}{$\alpha_t \equiv 1$} &$x = 1$ & 2.4e$+$0& 2.0e$+$0& 1.7e$+$0& 1.6e$+$0& \alert{1.6e$+$0}& 1.8e$+$0& 2.0e$+$0& 2.2e$+$0\\
&$x = 2$ & 3.4e$+$0& 2.7e$+$0& 2.1e$+$0& 1.9e$+$0& 1.8e$+$0& 1.7e$+$0& 1.7e$+$0& 1.7e$+$0\\
&$x = 3$ & 4.3e$+$0& 3.4e$+$0& 2.8e$+$0& 2.7e$+$0& 2.3e$+$0& 2.1e$+$0& 2.1e$+$0& 1.9e$+$0\\
&$x = 4$ & 4.9e$+$0& 3.8e$+$0& 3.2e$+$0& 3.0e$+$0& 2.6e$+$0& 2.4e$+$0& 2.2e$+$0& 2.1e$+$0\\
&$x = 5$ & - 	& 4.2e$+$0& 3.5e$+$0& 3.2e$+$0& 2.8e$+$0& 2.6e$+$0& 2.5e$+$0& 2.3e$+$0\\
\bottomrule
\multicolumn{10}{c}{}\\
\multicolumn{10}{c}{$\epsilon = 10^{-12}$}\\
\toprule
\multirow{5}{*}{SpaRSA} &$x = 1$ & 4.6e$+$0& 3.8e$+$0& 3.5e$+$0& 3.3e$+$0& 3.4e$+$0& 3.2e$+$0& 2.9e$+$0& \alert{2.9e$+$0}\\
&$x = 2$ & 7.0e$+$0& 6.6e$+$0& 5.3e$+$0& 4.2e$+$0& 3.8e$+$0& 3.4e$+$0& 3.2e$+$0& 3.0e$+$0\\
&$x = 3$ & 9.9e$+$0& 7.5e$+$0& 6.8e$+$0& 5.4e$+$0& 4.6e$+$0& 4.5e$+$0& 4.3e$+$0& 4.0e$+$0\\
&$x = 4$ & - 	& 8.8e$+$0& 7.3e$+$0& 6.2e$+$0& 5.4e$+$0& 5.1e$+$0& 4.7e$+$0& 4.2e$+$0\\
&$x = 5$ & - 	& 8.8e$+$0& 7.3e$+$0& 6.2e$+$0& 5.6e$+$0& 5.5e$+$0& 5.0e$+$0& 4.6e$+$0\\
\midrule
\multirow{5}{*}{$\alpha_t \equiv 1$} &$x = 1$ & 5.5e$+$0& 3.7e$+$0& 3.3e$+$0& 3.1e$+$0& 3.0e$+$0& 3.0e$+$0& 3.1e$+$0& 3.4e$+$0\\
&$x = 2$ & 5.7e$+$0& 4.5e$+$0& 3.6e$+$0& 3.4e$+$0& 3.2e$+$0& 3.2e$+$0& 3.1e$+$0& \alert{3.0e$+$0}\\
&$x = 3$ & - 	& 5.2e$+$0& 4.3e$+$0& 4.3e$+$0& 3.7e$+$0& 3.6e$+$0& 3.4e$+$0& 3.3e$+$0\\
&$x = 4$ & - 	& 5.8e$+$0& 4.8e$+$0& 4.5e$+$0& 4.1e$+$0& 3.9e$+$0& 3.7e$+$0& 3.5e$+$0\\
&$x = 5$ & - 	& - 	& 5.3e$+$0& 4.9e$+$0& 4.4e$+$0& 4.2e$+$0& 3.9e$+$0& 3.7e$+$0\\
\bottomrule
\end{tabular}
\end{table}

\subsection{Comparison with the SpaRSA variant for Matrix Completion}
\begin{figure}[tb]
	\centering
	\begin{tabular}{@{}c@{\hspace{5pt}}c@{\hspace{5pt}}c@{\hspace{5pt}}c@{}}
		\multicolumn{4}{c}{Iterations}\\
		\begin{subfigure}[b]{.24\textwidth}
			\includegraphics[width=\textwidth]{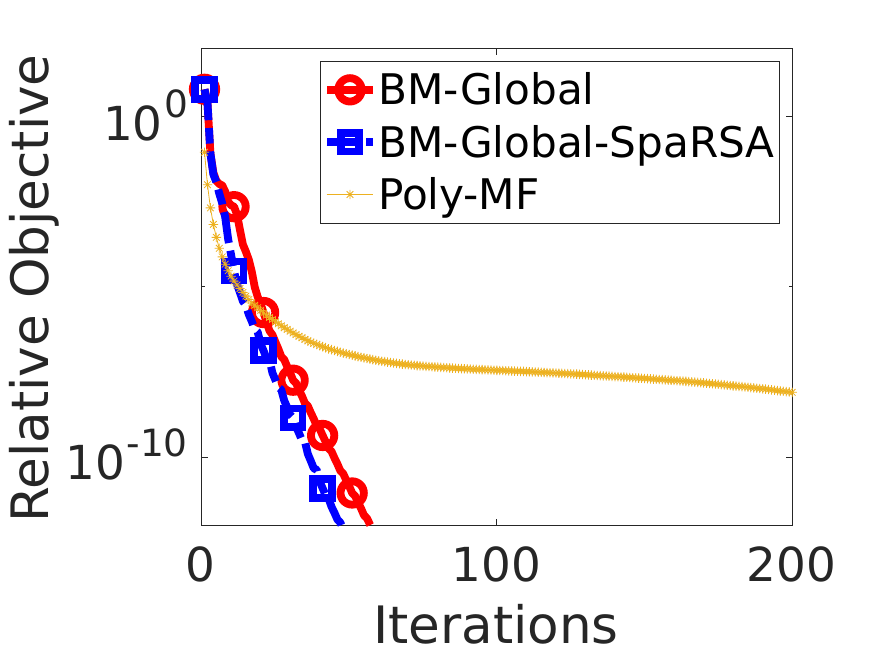}
			\caption{movielens100k}
		\end{subfigure} &
		\begin{subfigure}[b]{.24\textwidth}
			\includegraphics[width=\textwidth]{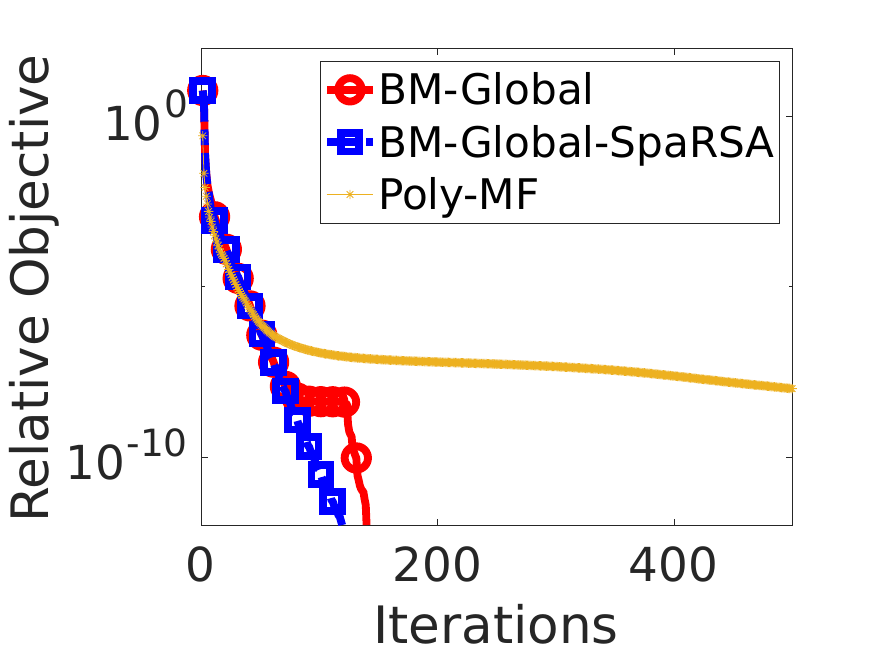}
			\caption{movielens10m}
		\end{subfigure} &
		\begin{subfigure}[b]{.24\textwidth}
			\includegraphics[width=\textwidth]{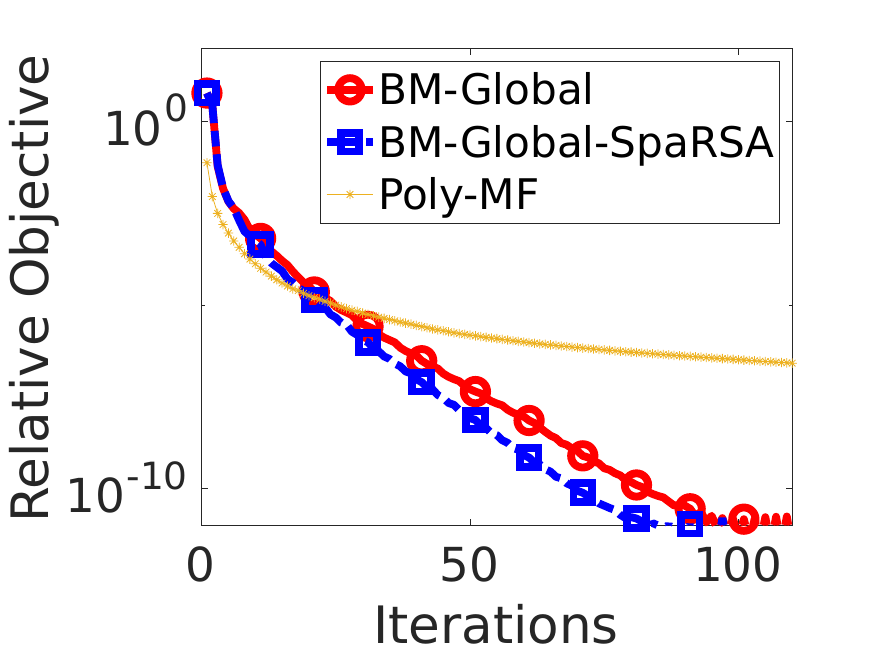}
			\caption{netflix}
		\end{subfigure} &
		\begin{subfigure}[b]{.24\textwidth}
			\includegraphics[width=\textwidth]{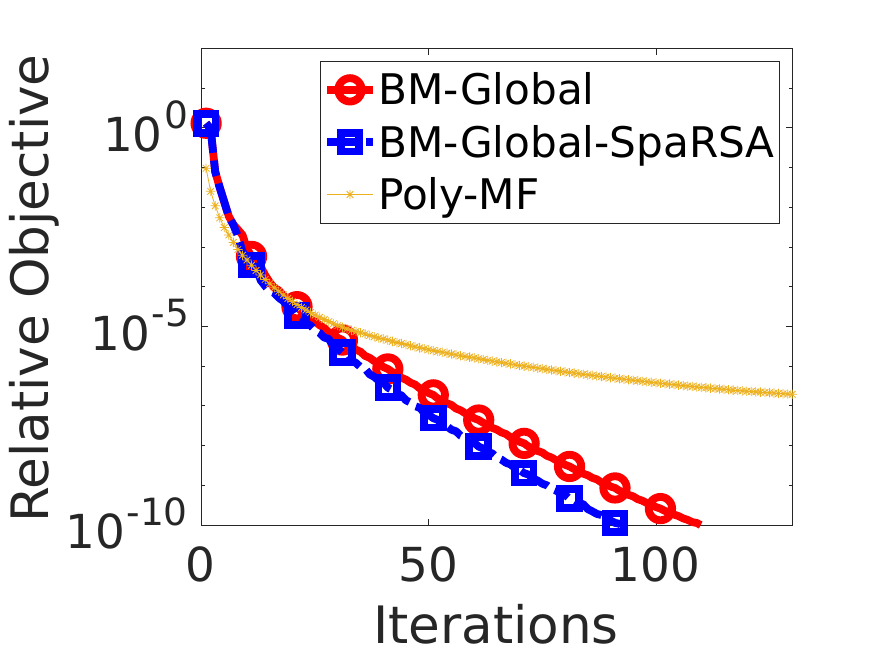}
			\caption{yahoo-music}
		\end{subfigure}\\
		\multicolumn{4}{c}{Running time}\\
		\begin{subfigure}[b]{.24\textwidth}
			\includegraphics[width=\textwidth]{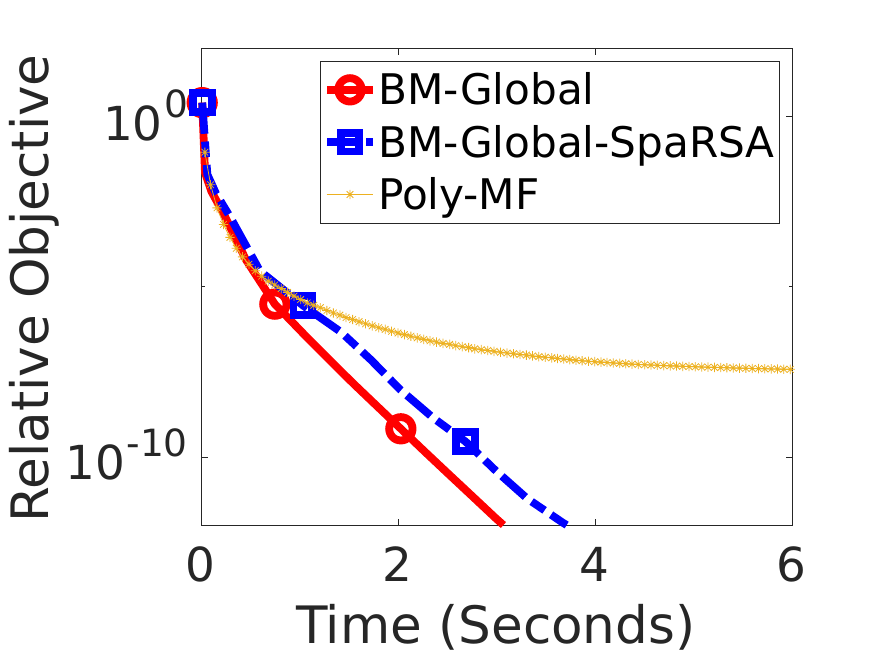}
			\caption{movielens100k}
		\end{subfigure} &
		\begin{subfigure}[b]{.24\textwidth}
			\includegraphics[width=\textwidth]{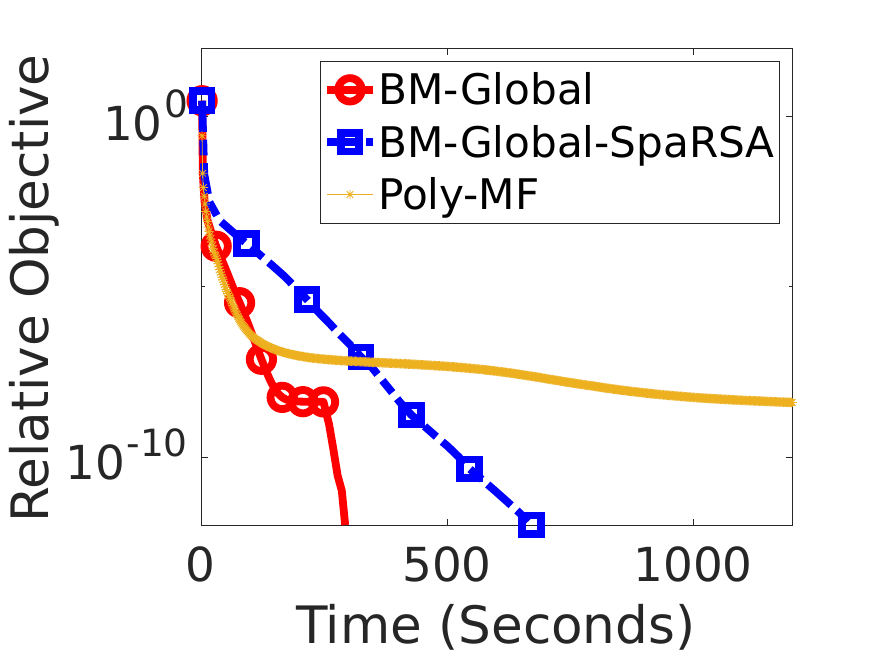}
			\caption{movielens10m}
		\end{subfigure} &
		\begin{subfigure}[b]{.24\textwidth}
			\includegraphics[width=\textwidth]{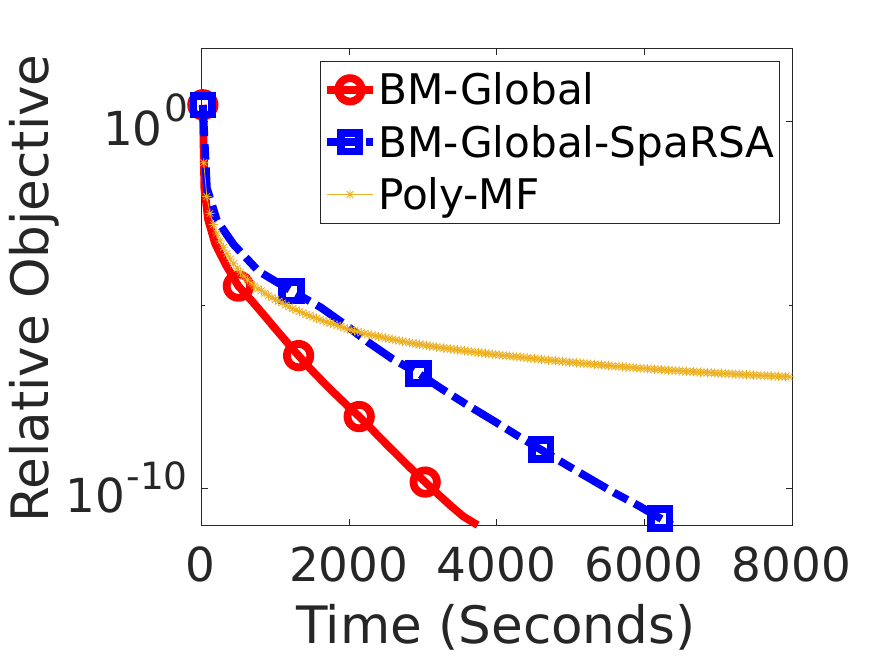}
			\caption{netflix}
		\end{subfigure} &
		\begin{subfigure}[b]{.24\textwidth}
			\includegraphics[width=\textwidth]{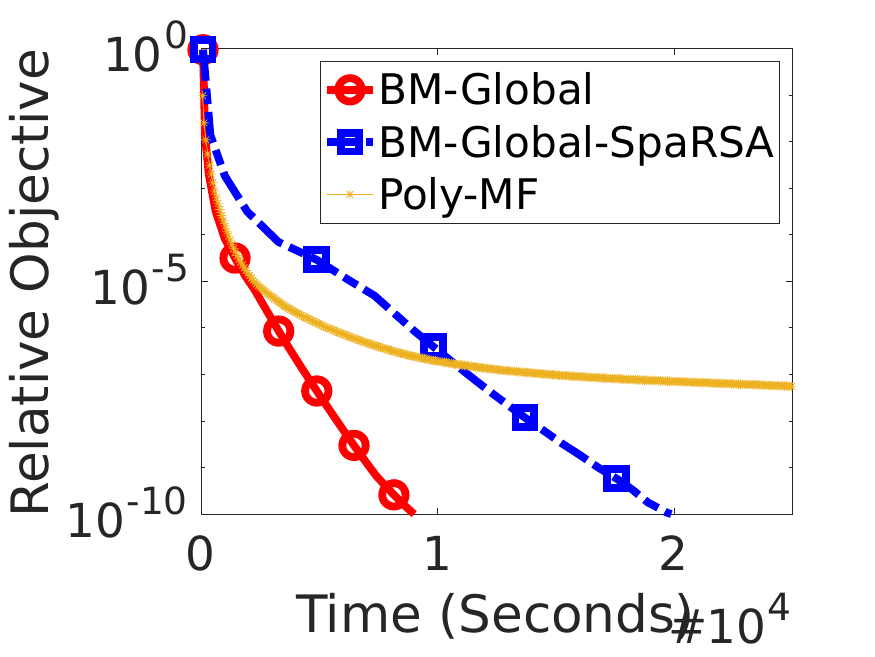}
			\caption{yahoo-music}
		\end{subfigure}
\end{tabular}
	\caption{Performance of the SpaRSA variant of \ours.
		Top row: iterations v.s. relative objective.
		Bottom row:
		running time (seconds) v.s. relative objective.}
	\label{fig:sparsa}
\end{figure}

In this subsection, we present experimental results with the SpaRSA variant.
The results are shown in \cref{fig:sparsa}.
As expected, this variant is faster than the fixed-step variant in
terms of iterations, although the difference is minor.
On the other hand, it is significantly slower in terms of the real
running time.
The likely reason is the possible additional approximate SVDs executed
in the backtracking procedure, while the inexact PG step in our
algorithm is not the major contributor to the objective decrease, so
the improvement in the convergence speed of this part cannot
counterbalance the corresponding additional cost.

\subsection{Parallelism}
\begin{figure}[tb]
	\centering
	\begin{tabular}{@{}c@{\hspace{5pt}}c@{\hspace{5pt}}c@{\hspace{5pt}}c@{}}
		\begin{subfigure}[b]{.24\textwidth}
			\includegraphics[width=\textwidth]{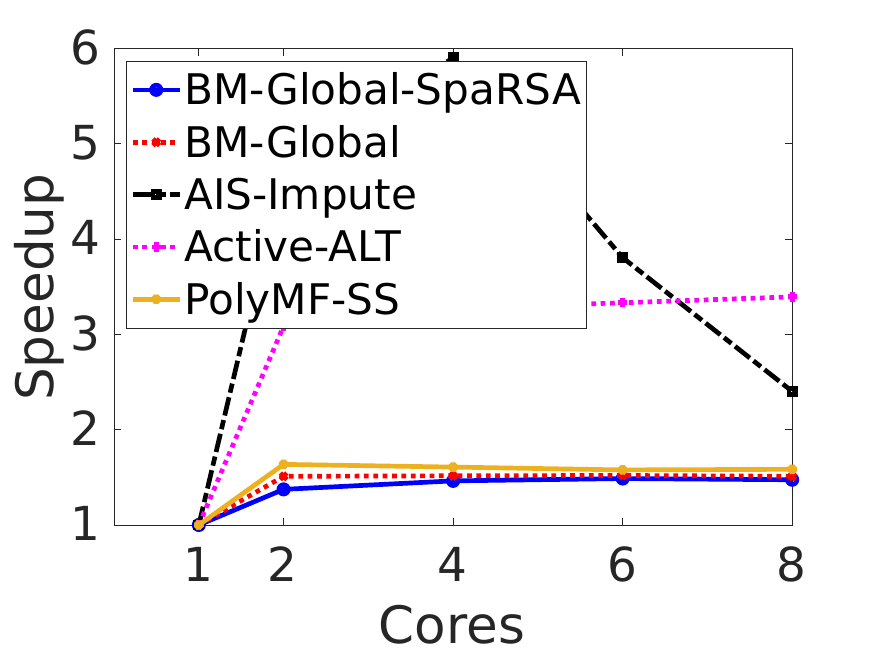}
			\caption{movielens100k}
		\end{subfigure} &
		\begin{subfigure}[b]{.24\textwidth}
			\includegraphics[width=\textwidth]{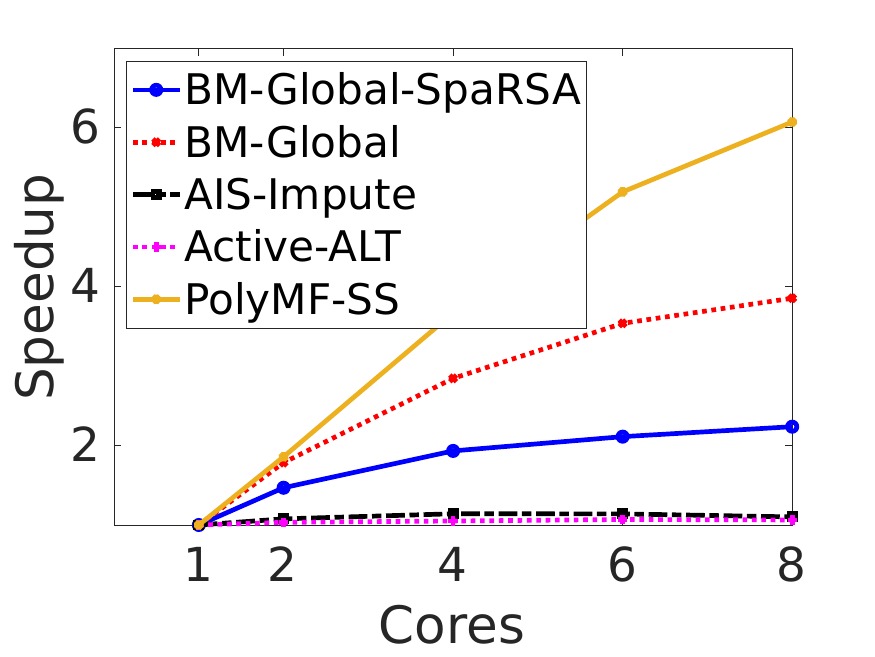}
			\caption{movielens10m}
		\end{subfigure} &
		\begin{subfigure}[b]{.24\textwidth}
			\includegraphics[width=\textwidth]{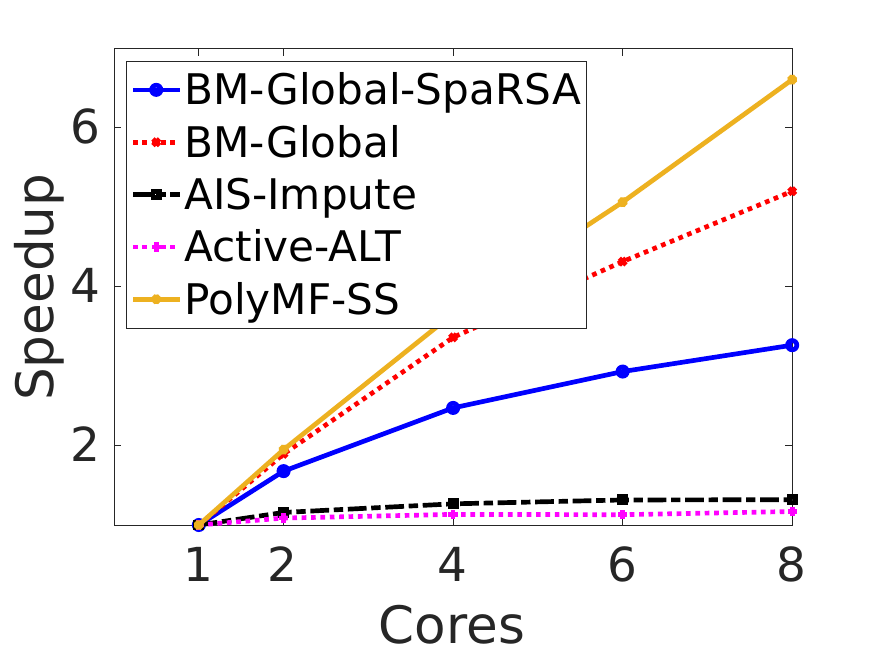}
			\caption{netflix}
		\end{subfigure} &
		\begin{subfigure}[b]{.24\textwidth}
			\includegraphics[width=\textwidth]{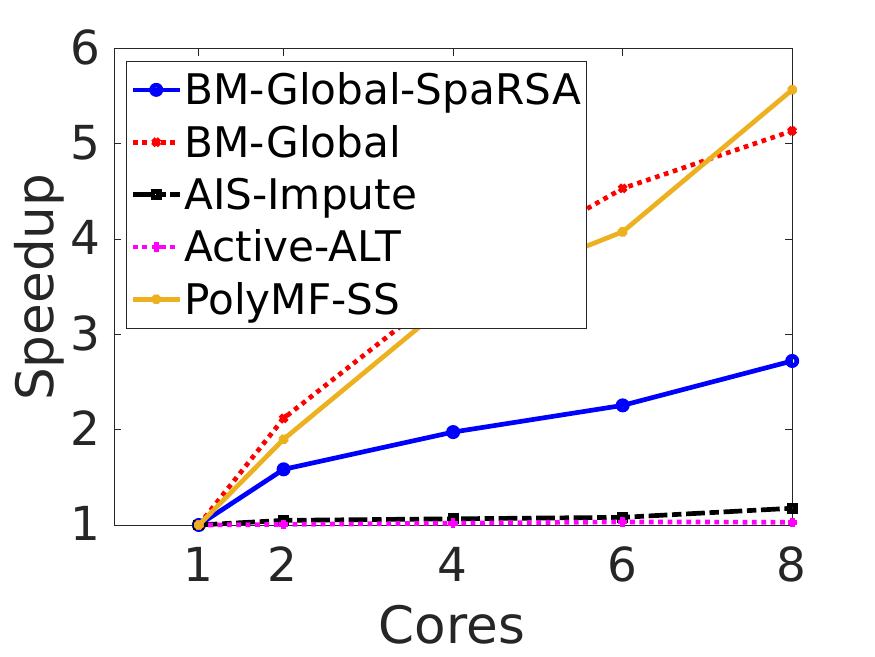}
			\caption{yahoo-music}
		\end{subfigure}
\end{tabular}
\caption{Comparison of different algorithms in terms of speedup with
respect to different number of cores.}
	\label{fig:speedup}
\end{figure}

We also examine the parallelism of our method.
In particular, we run our method with $1,2,4,8$ cores and check how much
time they respectively take to make \cref{eq:obj} no larger than
$10^{-6}$, and compute the speedup as follows.
\begin{equation*}
	\text{Speedup}(x) = \frac{\text{Running time of $x$
	cores}}{\text{Running time of $1$ core}}.
\end{equation*}
We can observe from \cref{fig:speedup} that except for the small
dataset movielens100k, PolyMF-SS always achieves the highest speedup
because most of its operations are inherently parallel and
computationally heavy, and \ours comes the next, while AIS-Impute
and Active-ALT barely exhibit any parallelism at all.

\subsection{Numerical results for the APG method}
This subsection presents numerical results of the APG method \cref{alg:apg} on
testing problems considered in \cref{sec:qsdpexp}.

In our experiments, the APG method is executed with $Y_0 = X_0 = 0$
and a fixed step size $\alpha = 1/L$, where $L\coloneqq
\norm{\mathcal{A}\mathcal{A}}_2$ is estimated by the \texttt{eigs} subroutine in MATLAB.
To perform the projection $P_\mathcal{X}$ through  \cref{lemma-PiX}, we use the \texttt{eig} subroutine in MATLAB.
Our stopping condition for the APG method is
\[
    \frac{\lvert f(X_{t+1}) - f(X_t)\rvert}{1 + \lvert f(X_t)\rvert } \leq \texttt{tol},
\]
where $\texttt{tol}$ is the tolerance (chosen as $10^{-6}$ in our
experiments).
Moreover, we set the maximal number of iterations to be $10000$ and
the maximal computational time to be four hours.
The computational results for the RKE problems and the molecular
conformation problems are presented in \cref{tab:rke-apg,tab:mcp-apg}, respectively.
We can observe from the presented results that the APG method usually
performs worse than \ours and QSDPNAL. Indeed,
it returns solutions with a lower accuracy and takes more
computational time. Also, the numerical results suggest that the APG
method could be sensitive to the sign of the parameter $\lambda$,
since it performs much better in the cases of $\lambda > 0$ then in the cases of $\lambda < 0$.

\begin{table}[htb!]
    \centering
    \begin{tabular}{|l|r|r|r|r|r|r|} 
        \hline 
         \texttt{Name} & $n$ & $\eta_{\rm prim}$ & $\eta_{\rm opt}$ & \texttt{rnk} & \texttt{Time} & \texttt{Iter} \\
         \hline 
 BrainMRI &  124 & 2e-16 & 1e-03 &    5 &   0.7 &   86 \\
 protein &  213 & 3e-15 & 1e-03 &   27 &   6.7 &  285 \\
 CoilDelftDiff &  288 & 2e-15 & 1e-03 &   32 &   9.0 &  234 \\
 coildelftsame &  288 & 3e-15 & 2e-03 &   32 &   7.7 &  189 \\
 CoilYork &  288 & 1e-15 & 1e-03 &   21 &   9.2 &  231 \\
 Chickenpieces-5-45 &  446 & 4e-16 & 2e-03 &   28 &  32.6 &  288 \\
 newgroups &  600 & 3e-17 & 2e-03 &   84 &  69.7 &  341 \\
 flowcytodis &  612 & 8e-16 & 1e-03 &   16 &  84.6 &  353 \\
 DelftPedestrians &  689 & 2e-15 & 2e-03 &   70 & 105.0 &  377 \\
 WoodyPlants50 &  791 & 8e-16 & 2e-03 &   49 & 185.7 &  478 \\
 delftgestures & 1500 & 1e-15 & 1e-02 &  106 & 283.3 &  191 \\
 zongker & 2000 & 1e-15 & 4e-03 &  303 & 1236.5 &  451 \\
 polydish57 & 4000 & 2e-15 & 1e-02 &  139 & 8530.7 &  357 \\
 polydism57 & 4000 & 3e-15 & 1e-02 &   83 & 8023.2 &  337 \\ \hline 
    \end{tabular}
    \caption{Computational results for the APG method on RKE problems.}
    \label{tab:rke-apg}
\end{table}

\begin{table}[htb!]
    \centering
    \begin{tabular}{|l|r|r|r|r|r|r|r|} 
        \hline 
         \texttt{Name} & $n$ & $\eta_{\rm prim}$ & $\eta_{\rm opt}$ & \texttt{rnk} & \texttt{RMSD} & \texttt{Time}  & \texttt{Iter}  \\
         \hline 
 1PBM &  126 & 4e-16 & 6e-06 &   14 & 7.4 &  30.1 & 5073 \\
 1AU6 &  161 & 2e-16 & 6e-06 &   15 & 6.8 &  55.9 & 5250 \\
 1PTQ &  402 & 2e-15 & 2e-06 &   20 & 10.2 & 727.8 & 10000 \\
 1CTF &  487 & 5e-16 & 2e-06 &   23 & 11.2 & 1250.5 & 10000 \\
 1HOE &  558 & 1e-15 & 2e-06 &   24 & 11.6 & 1831.3 & 10000 \\
 1LFB &  641 & 1e-15 & 2e-06 &   28 & 13.4 & 2379.5 & 10000 \\
 1PHT &  666 & 5e-16 & 2e-06 &   28 & 12.2 & 2598.5 & 10000 \\
 1F39 &  767 & 1e-15 & 2e-06 &   33 & 13.6 & 3432.6 & 10000 \\
 1DCH &  806 & 3e-17 & 2e-06 &   32 & 13.4 & 3776.6 & 10000 \\
 1HQQ &  891 & 1e-15 & 2e-06 &   41 & 15.0 & 4539.2 & 10000 \\
 1POA &  914 & 9e-16 & 2e-06 &   37 & 14.2 & 4844.0 & 10000 \\
 1AX8 & 1003 & 8e-16 & 2e-06 &   40 & 14.3 & 5969.8 & 10000 \\
 1TJO & 1394 & 2e-15 & 2e-06 &   62 & 19.0 & 12512.3 & 10000 \\
 1RGS & 2015 & 6e-16 & 4e-06 &  126 & 20.2 & - & 4952 \\
 1TOA & 2138 & 2e-15 & 5e-06 &  136 & 19.2 & - & 4303 \\
 1KDH & 2846 & 5e-16 & 1e-05 &  254 & 21.9 & - & 1901 \\
 1NFG & 3501 & 3e-15 & 5e-05 &  391 & 21.2 & - &  903 \\
 1BPM & 3672 & 2e-16 & 6e-05 &  436 & 23.8 & - &  788 \\
 1MQQ & 5510 & 7e-16 & 3e-04 &  888 & 26.0 & - &  247 \\
 \hline 
    \end{tabular}
    \caption{Computational results for the APG method on molecular conformation problems.
``-'' indicates that the solver is terminated because the maximum
running time of four hours is reached.}
    \label{tab:mcp-apg}
\end{table}